\pdfoutput=1 

\documentclass[10pt,journal,compsoc]{IEEEtran}
\usepackage{amsmath}
\usepackage{amsfonts}
\usepackage{bbding}
\usepackage{amsmath}
\usepackage{amsfonts}
\usepackage[utf8]{inputenc} 
\usepackage[T1]{fontenc}    
\usepackage{hyperref}       
\usepackage{url}            
\usepackage{booktabs}       
\usepackage{amsfonts}       
\usepackage{nicefrac}       
\usepackage{microtype}      
\usepackage{amsthm,amsmath,amssymb}
\usepackage{mathrsfs}
\usepackage{graphicx}
\usepackage{subfigure}
\usepackage{wrapfig}
\usepackage{bbding}
\usepackage{float}

\ifCLASSOPTIONcompsoc
  \usepackage[nocompress]{cite}
\else
  \usepackage{cite}
\fi
\ifCLASSINFOpdf
\else
\fi
\begin{document}
%
\title{Class-aware Sounding Objects Localization \\via Audiovisual Correspondence}

%
%
%
%

\author{
        Di Hu,
        Yake Wei,
        Rui Qian,
        Weiyao Lin,~\IEEEmembership{Senior~Member,~IEEE,}
         Ruihua Song,~\IEEEmembership{Senior~Member,~IEEE,}
        Ji-Rong Wen,~\IEEEmembership{Senior~Member,~IEEE}
\IEEEcompsocitemizethanks{

\IEEEcompsocthanksitem D. Hu (corresponding author), Y. Wei, R. Song and J-R. Wen are with the Gaoling School of Artificial Intelligence, and Beijing Key Laboratory of Big Data Management and Analysis Methods, Renmin University of China, Beijing 100872, China.\protect\\
E-mail: \{dihu, yakewei, rsong, jrwen\}@ruc.edu.cn

\IEEEcompsocthanksitem R. Qian is with the Department of Information Engineering, the Chinese University of Hong Kong, Hong Kong, China.\protect\\
E-mail: qr021@ie.cuhk.edu.hk

\IEEEcompsocthanksitem W. Lin is with the School of Electronic Information and Electrical Engineering, Shanghai Jiao Tong University, Shanghai 200240, China.\protect\\
E-mail: wylin@sjtu.edu.cn
}
}

\IEEEtitleabstractindextext{%
\begin{abstract}
Audiovisual scenes are pervasive in our daily life. It is commonplace for humans to discriminatively localize different sounding objects but quite challenging for machines to achieve class-aware sounding objects localization without category annotations, i.e., localizing the sounding object and recognizing its category. To address this problem, we propose a two-stage step-by-step learning framework to localize and recognize sounding objects in complex audiovisual scenarios using only the correspondence between audio and vision. First, we propose to determine the sounding area via coarse-grained audiovisual correspondence in the single source cases. Then visual features in the sounding area are leveraged as candidate object representations to establish a category-representation object dictionary for expressive visual character extraction. We generate class-aware object localization maps in cocktail-party scenarios and use audiovisual correspondence to suppress silent areas by referring to this dictionary. Finally, we employ category-level audiovisual consistency as the supervision to achieve fine-grained audio and sounding object distribution alignment. Experiments on both realistic and synthesized videos show that our model is superior in localizing and recognizing objects as well as filtering out silent ones. We also transfer the learned audiovisual network into the unsupervised object detection task, obtaining reasonable performance.
\end{abstract}

\begin{IEEEkeywords}
Class-aware sounding object localization, audiovisual correspondence, distribution alignment
\end{IEEEkeywords}}

\maketitle

\IEEEdisplaynontitleabstractindextext

%
\IEEEpeerreviewmaketitle

\section{Introduction}
\IEEEPARstart{O}{ur} daily life consists of multi-modal scenes, where audio and visual messages are the most pervasive ones. We can easily infer rich semantic information from their natural correspondence to achieve effective multi-modal perception~\cite{stein1993merging,proulx2014multisensory,Hu2020CrossTaskTF}, e.g., when hearing barking, we instinctively associate it with the dog, while when hearing meow, we know it is from the cat nearby.
Given this, we wonder that can machine intelligence also benefit from audiovisual correspondence?

\begin{figure}
    \centering
    \includegraphics[width=\linewidth]{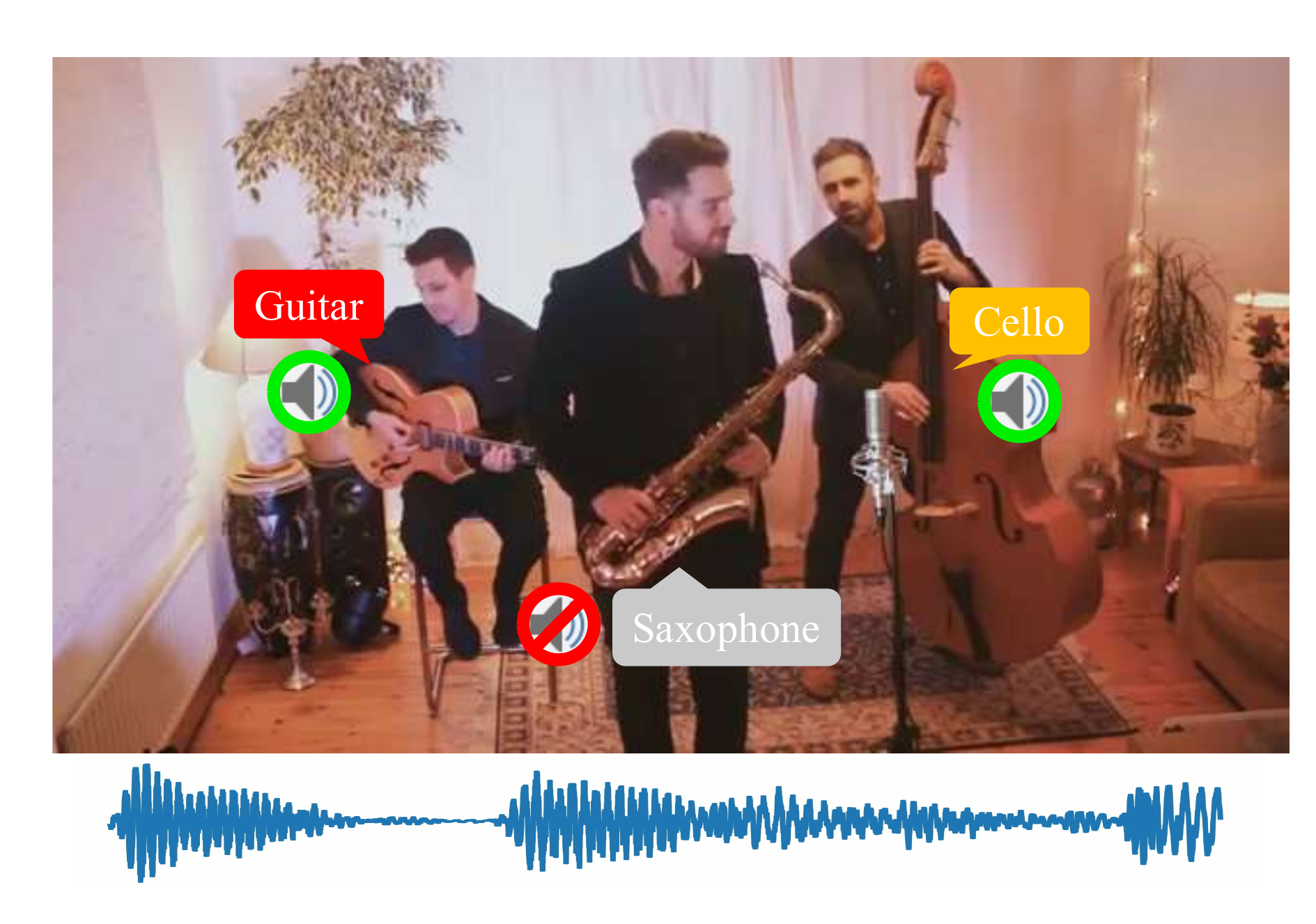}
    \vspace{-3mm}
    \caption{\textbf{An example of cocktail-party scenario.} It contains sounding guitar, sounding cello, and silent saxophone. We aim to discriminatively localize the sounding instruments and filter out the silent ones. Video URL: https://www.youtube.com/watch?v=ebugBtNiDMI.}
    \label{fig:teaser}
\end{figure}

To help machines pursue human-like multi-modal perception, it is significant to solve the core challenge of visual sound localization, which targets to semantically correlate sounds with relevant visual regions without resorting to category annotations~\cite{soundloc99,izadinia2012multimodal,arandjelovic2017objects,senocak2018learning,vehicle}, i.e., not only localize the region of the sounding object but also recognize its category. One intuitive method is to maximize the similarity between sound embeddings and the visual features of the corresponding source object, which has exhibited promising performance in simple scenarios consisting of single sound source~\cite{oquab2015object,avscene,senocak2018learning}. However, in our daily scenario, it is commonplace that there simultaneously exist multiple sounding objects and silent ones (the silent objects are considered capable of producing sound), e.g., cocktail-party scenario. It is also challenging to discriminatively localize sounding objects, i.e., localizing while identifying the category of sounding objects like humans. Under such circumstances, the aforementioned simple strategy mostly fails to discriminatively localize all sound sources that produce the mixed sound~\cite{hu2019deep}. Audiovisual content modelling has been proposed to excavate concrete audiovisual components in the scene to address this problem. However, due to the lack of semantic annotations, existing works have to require extra scene prior knowledge~\cite{hu2019deep,curriculum,qian2020multiple} or construct complex pretext task~\cite{zhao2018sop,som} to achieve this goal.
Despite this, these methods fail to achieve discriminative sounding objects localization, i.e., answer not only \emph{where the sounding area is} but also infer \emph{what the sounding area is}.

In this work, we aim to achieve class-aware sounding objects localization given mixed sound of a complex cocktail-party scenario, where multiple visual objects are making sound or remaining silent, as shown in Fig.~\ref{fig:teaser}. The challenge of this interesting problem majorly lies in two aspects: 1) How to discriminatively localize and recognize visual objects that belong to different categories without extra semantic annotations of objects; 2) How to determine which visual objects are producing sound and filter out silent ones by referring to the corresponding mixed sound. When facing the above challenges, humans could transform these seemingly unlearnable tasks into learnable by starting from a simpler initial state and then gradually building more complicated structural representations on it~\cite{elman1993learning}. Motivated by this, we propose a two-stage step-by-step learning framework to pursue class-aware sounding objects localization, starting from single sound scenarios and then expanding to cocktail-party cases. The core idea is that different objects usually make different sounds, and it is trivial to align the object in visual and its sound in the simple single source scenes, which can provide prior alignment and categories knowledge for complex multiple sound scenarios. Particularly, we first extract visual object features from class-agnostic sounding area localization in single source scenario and establish a category-representation object dictionary to generate the correspondence between category and object visual information.
Based on this dictionary, we figure out class-aware candidate sound sources in multi-source scenarios. Then considering the consistency between co-occurring audio and visual messages, we further reduce class-aware sounding objects localization into audiovisual matching by minimizing the category-level distribution difference. We manage to achieve class-aware sounding objects localization and filter out silent ones in complex cocktail-party scenarios with this evolved curriculum. 

Our previous conference paper~\cite{hu2020discriminative} has achieved considerable performance in music scenes. However, it requires the number of instrument categories as prior and has difficulty adapting to harder daily cases for general application, which cover a variety of sounding objects, e.g., humans, animals, vehicles. In this paper, we get rid of this prior and generalize our model to diverse daily scenarios. We evaluate our model on natural and synthesized videos in music scenes and more general daily cases. A wide range of ablation studies is also conducted to validate our framework thoroughly. Also, such an object localization framework can be leveraged for the typical vision task of object detection. 

To sum up, our main contributions are listed as follow:
\begin{itemize}
  \item [1)] 
  Inspired by the human perception of the multimodal environment, we introduce a challenging problem, class-aware sounding objects localization, i.e., to localize and recognize objects via their produced sound without category annotations. 
  \item [2)]
  We propose a novel two-stage learning framework to gain the correspondence between object visual representations and categories knowledge from single source localization using only the alignment between audio and vision as the supervision, then expand to complex cocktail-party scenarios for class-aware sounding objects localization. The experiment results show that our framework can well solve the class-aware sounding objects localization task in complex multiple source scenarios. 
  \item [3)]
  We collect and synthesize a new cocktail-party video dataset and annotate bounding boxes of sounding objects to evaluate class-aware sounding objects localization performance. Our method exhibits promising performance on both synthetic and realistic data. Further, the class-aware object localization framework learned from audiovisual consistency can be applied to the typical vision task of object detection.

\end{itemize}
\section{Related Work}

\subsection{Object Localization}

Weakly- and self-supervised methods for object localization arise since annotating bounding boxes for images is a huge burden. These methods are committed to achieving comparable model performance while reducing the dependence on bounding box annotations. Existing weakly-supervised methods aim to detect objects by using image-level annotations instead of bounding box-level annotations~\cite{oquab2014learning,oquab2015object,bazzani2016self,zhou2016learning,grad-cam,grad-cam++}. Oquab et al.~\cite{oquab2014learning} proposed to localize objects by evaluating classification scores on multiple overlapping patches. In~\cite{oquab2015object}, authors employed global max pooling to localize the most salient point on objects. Bazzani et al.~\cite{bazzani2016self} proposed a self-taught mechanism to identify the image regions that exert the most impact on recognition scores. Later, \emph{Class Activation Map} (CAM) method \cite{zhßou2016learning} was proposed to use global average pooling to identify all discriminative regions of an object, and two gradient-based methods~\cite{grad-cam,grad-cam++} further generalized CAM to arbitrary CNN architectures. Differently, semantic information of objects is agnostic in self-supervised learning. Baek et al.~\cite{psynet} proposed to utilize point symmetric transformation as the self-supervision signal to promote class-agnostic heatmap extraction in the object localization task. 

The aforementioned methods just utilize visual information to mine effective supervision. However, in this work, we propose to adopt the correspondence between audio and vision as the supervision for class-ware object localization.

\subsection{Audiovisual Learning}

In daily life, we perceive our world through multiple senses, such as seeing with eyes and hearing with ears at the same time, where visual information is usually accompanied by sound information. Most existing audiovisual learning methods leverage such natural correspondence between audio and vision to establish the joint embedding space~\cite{L3,arandjelovic2017objects,avscene,soundnet,soundsupv}. Aytar et al.~\cite{soundnet} exploited the temporal co-occurrence between audio and vision to learn good audio representations. A teacher-student learning framework was proposed to transfer knowledge from the visual modality to the audio modality. Owens et al.~\cite{soundsupv} utilized sound as the supervision to learn visual representations. These methods are proposed to establish representation of one modality under the supervision of another modality. 

Unlike that, Arandjelovi\'{c} et al.~\cite{L3} employed clip-level correspondence guiding audiovisual content modeling. Owens et al.~\cite{avscene} caught the temporal synchronization within a clip for the same purpose. Besides, cluster-related methods are explored in many scenarios~\cite{hu2019deep,curriculum,hu2019deep,curriculum,alwassel2020self,asano2020labelling,tian2021unsupervised}. Hu et al.~\cite{hu2019deep,curriculum} performed multi-modal clustering to associate latent sound-object pairs, but the predefined number of clusters has a great impact on the performance. Alwassel et al.~\cite{alwassel2020self} expanded deep clustering to audiovisual data to establish more robust multi-modal representation. Asano et al.~\cite{asano2020labelling} leveraged the correspondence between audio and vision and proposed a clustering method aiming to self-label a video dataset without any human annotations. In addition, Alayrac et al.~\cite{alayrac2020self} introduced a network that learns representation from multiple modalities, which can be leveraged in multiple downstream tasks.

The previous representation learning methods simply used scene-level audiovisual consistency, which did not sufficiently mine the semantics of the objects in the scene. Unlike this, our method focuses on fine-grained category-level audiovisual consistency and learning a category-representation object dictionary to achieve the goal of discriminative sounding objects localization. 

\subsection{Sounding Object Localization in Visual Scenes}

Sounding object localization in the visual scene refers to associating the sound signal with the specific visual location. This task takes full advantage of the audiovisual consistency and enables the machine model to learn human-like audiovisual perception. Before the booming of deep learning, some methods for sound source localization were born. These methods relied on spatial sparsity prior of audio-visual events~\cite{kidron2005pixels} or hand-crafted motion cues~\cite{barzelay2007harmony}. The semantic information of the object is known in these methods. In addition, there are some acoustic hardware-based approaches~\cite{van2004optimum,zunino2015seeing} with the aid of devices, e.g., microphone arrays, to grad phase the difference of sound arrival. Conversely, our proposed method neither requires semantic information of objects nor other devices, overcoming the limitations of the above methods.

Recent methods for visual sound localization majorly focus on correlating audio and visual information~\cite{arandjelovic2017objects,avscene,senocak2018learning,ave,hu2019deep,som,zhao2018sop,NEURIPS2018_01161aaa,gao20192}. Arandjelovi\'{c} et al.~\cite{arandjelovic2017objects} and Owens et al.~\cite{avscene} adopted CAM~\cite{zhou2016learning} or similar methods to calculate the correspondence score between global audio feature and spatial visual features to figure out sounding areas. Senocak et al.~\cite{senocak2018learning} developed a cross-modal attention mechanism to capture the primary visual areas in a semi-supervised or unsupervised manner. The cross-modal attention mechanism can reflect the localization information of the sound source for interaction between the audio and visual modalities. Tian et al.~\cite{ave} proposed audio-driven visual attention as well as temporal alignment to figure out the semantically correlated visual regions. These methods mostly perform well in simple scenes with single sound source, but are comparatively poor to localize multiple objects with mixed sounds.

To address this challenge, Zhao et al.~\cite{zhao2018sop,som} developed a self-supervised mix-then-separate sound-separation framework to build the channel-wise association between audio and visual feature maps, where the sounding object is determined through the sound energy of each pixel.
Hu et al.~\cite{hu2019deep} leveraged clustering to represent different audiovisual components to correlate each sound center with its visual source, but it requires the number of sound sources as prior. Qian et al.~\cite{qian2020multiple} developed a two-stage learning framework that disentangles visual and audio representation of different categories, then performs cross-modal feature alignment. However, this work still requires extra scene prior due to the lack of one-to-one annotations. Oya et al.~\cite{oya2020we} proposed a step-wise training strategy that first gets potential sounding objects based on visual information and then identifies the proposal based on audio information. Nevertheless, the experimental scenario in the work is relatively simple (two objects, one of which makes a sound). Tian et al.~\cite{tian2021cyclic} performed sound separation with the help of visual object proposal but required a pre-trained object detector. Chen et al.~\cite{chen2021localizing} considered the background regions of the localization results and offered a more fine-grained localization framework.

Unlike the above methods, our method uses only the correspondence between audio and vision as the supervision and can deal with complex cocktail-party scenarios. We utilize the established category-representation object dictionary to produce class-aware object localization maps, which filter out the silent objects. Moreover, our method can be applied to the typical vision task of object detection.

\subsection{Sounding Object Localization in Psychophysics}

Some theories from cognitive science and psychophysics can inspire the object localization and recognition task. Many methods in this field attempt to explore the relationship between visual information and sounding object localization~\cite{jones1975eye,majdak20103,shelton1980influence}. In~\cite{jones1975eye}, Jones et al. found the sound-related visual information improves search efficiency in auditory space perception. Subsequently, Shelton et al.~\cite{shelton1980influence} demonstrated it can improve the accuracy of localization. These studies show that combining visual and audio modality for sounding object localization tasks is closely correlated to human psychophysics and cognition.

In addition, multi-source sounding object localization, e.g., the cocktail-party scenarios, is complicated. But people can talk smoothly despite the noise around them. This phenomenon is known as the cocktail party effect. Many researchers have tried to address this issue from different perspectives. Norman et al.~\cite{norman1976memory} pointed out the connection between short-term memory and listening and analyzed the cocktail-party effect from a cognitive perspective. Blauert et al.~\cite{blauert1997spatial} illustrated that the desired sound signal coming from one direction is less effectively masked by the noise that originates in a different direction when listening binaurally. 

In this work, motivated by the finding that humans can adopt the step-by-step method in dealing with complicated tasks: start by mastering simple tasks, and then apply what they have learned to complex tasks~\cite{elman1993learning}, a two-stage learning framework is proposed to solve the class-aware sounding object localization problem, modelled on human cognition.
\begin{figure*}[t]
    \centering
    \includegraphics[width=1\linewidth]{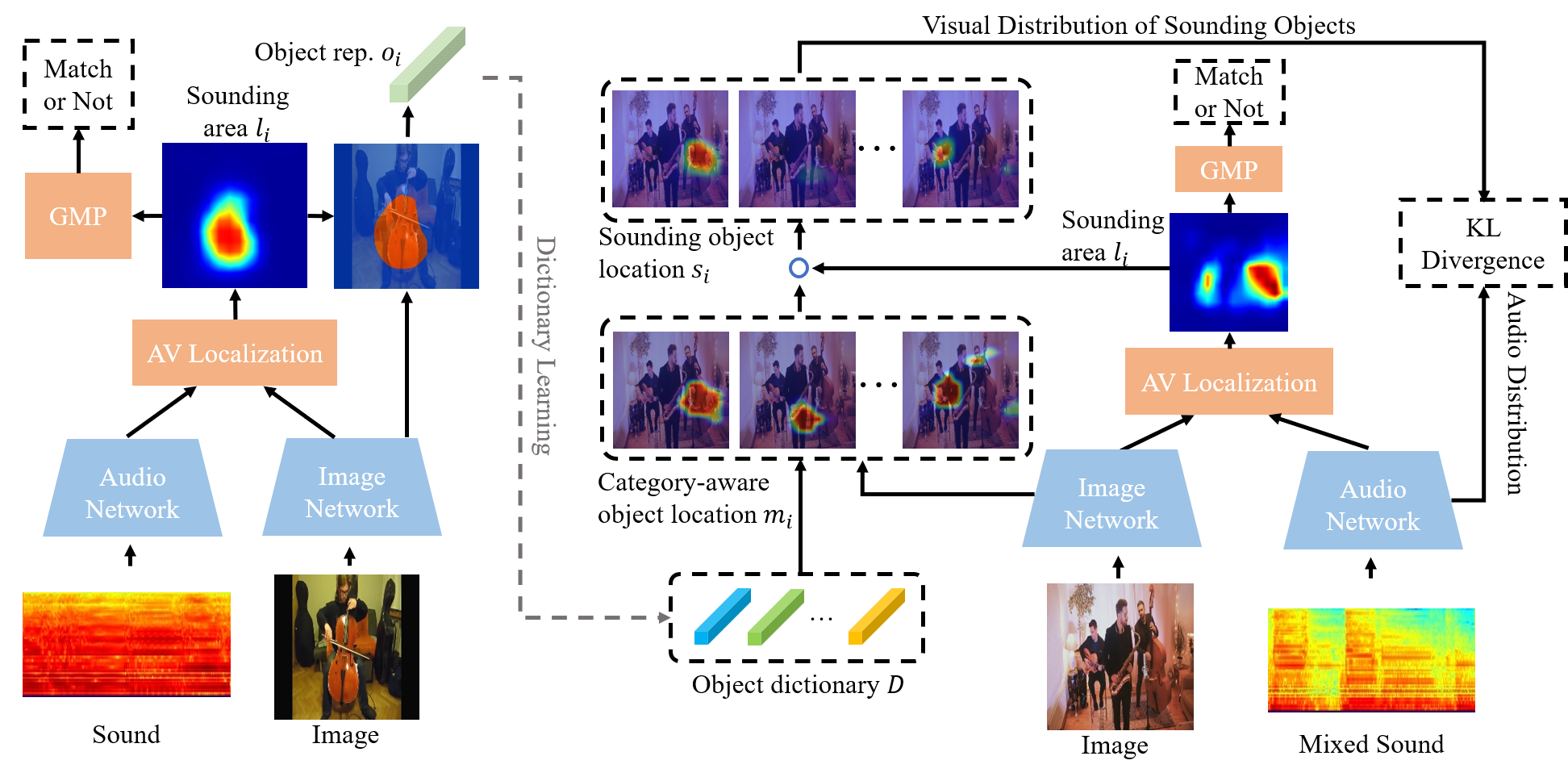}
    \vspace{-3mm}
    \caption{\textbf{An overview of the proposed framework.} First, we use audiovisual correspondence as the supervision to localize sounding area and learn object representation (left). Then, we employ a built object dictionary to generate class-aware localization maps and refer to the inferred sounding area to eliminate silent objects. In that way, we reduce the localization task into a distribution matching problem, where we use KL divergence to minimize the audiovisual distribution difference (right).}
    \label{fig:framework}
\end{figure*}

\section{Proposed Method}

In this section, we first introduce the problem of class-aware multiple sounding object localization. Next, we present our step-by-step learning framework to extract robust object representations and perform the cross-modal alignment. To simultaneously localize and recognize sounding objects, we propose a novel category distribution matching mechanism. The framework contains two core modules: category-representation object dictionary and distribution-level audiovisual alignment as illustrated in Fig.~\ref{fig:framework}.

\subsection{Problem Definition}

In this work, we take discriminative sounding objects localization as the core problem to be addressed, where we aim to visually localize and recognize sounding objects from the given mixed sound without resorting to category annotations. Particularly, the problem can be formulated as follows. Given a set of audiovisual pairs with arbitrary number of sounding objects, $\mathcal{X}=\left\{(a_i,v_i)|i=1,2,...,N\right\}$, we aim to establish category-representation object dictionary and achieve discriminative sounding objects localization through the self-supervision between co-occurring audio and visual message.

To facilitate this challenging problem, we propose a two-stage learning framework, evolving from the localization in simple scenario with a single sounding object to the complex one with multiple sounding objects, i.e., cocktail-party. Concretely, we divide $\mathcal{X}$ into one simple set consisting of single-sound videos, $\mathcal{X}^s=\left\{(a_i^s,v_i^s)\right|i=1,2,...,N^s\}$, and one complex set consisting of cocktail-party videos, $\mathcal{X}^c=\left\{(a_i^c,v_i^c)\right|i=1,2,...,N^c\}$, where $\mathcal{X}=\mathcal{X}^s \cup \mathcal{X}^c$ and $\mathcal{X}^s \cap \mathcal{X}^c=\emptyset$.
In the first stage, considering that it is comparatively easy to figure out the spatial area of the sounding object in simple scenes, we propose to extract a potential visual representation of the sounding object from audiovisual localization map in $\mathcal{X}^s$, based on which we build a category-representation object dictionary as the reference of category information. In the second stage, by referring to the learned object dictionary, we delve into discriminative sounding objects localization in the complex scenario $\mathcal{X}^c$, where the category distribution of localized sounding objects is required to match the distribution of their mixed sound according to the natural audiovisual consistency~\cite{hu2019deep}. In the following, we will detail the first and second learning stage for class-aware sounding objects localization as well as category-representation object dictionary generation. 

\subsection{Single Sounding Object Localization}

In this subsection, we focus on establishing the semantic relationship between audio and image pairs in simple scenes, $\mathcal{X}^s$. Based on the learned relationship, we aim to figure out the spatial area of the sounding object.

For an arbitrary audiovisual pair $(a_i^s,v_i^s)\in\mathcal{X}^s$, we need first to figure out the spatial area on the image that is highly correlated to the corresponding audio as the approximated object area. To achieve this goal, we employ convolution-based neural networks to extract audio and visual deep features. Note that we employ globally pooled sound embedding $g(a_i^s)\in\mathbb{R}^C$ as the global audio descriptor, and use visual feature map $f(v_i^s)\in\mathbb{R}^{C\times H\times W}$ with spatial resolution $H\times W$ to describe the local image regions, where $C$ denotes the number of feature channels. Then, we encourage the localization network to enhance the similarity between $g(a_i^s)$ and sounding area on $f(v_i^s)$, but suppress the similarity between mismatched audio and visual object, i.e., the pairs from different videos, $(a_i^s, v_j^s)$, where $ i\ne j$.
Particularly, we use cosine similarity between audio embedding and visual feature map to measure the relationship between the audio and visual pattern on each spatial grid. To match the scale of binary supervision, we employ a single-layer $1\times 1$ convolution followed with sigmoid activation to process the obtained cosine similarity and generate an audiovisual localization map $l\in\mathbb{R}^{H\times W}$. The specific calculation of $l$ can be formulated as
\begin{align}
    \label{eq:cosine}
    l(g(a),f(v)) = \sigma\left(conv\left(\frac{g(a)^Tf(v)}{||g(a)||_2||f(v)||_2}\right)\right),
\end{align}
where $\sigma$ indicates sigmoid activation and $conv$ is $1\times 1$ convolution.
In this way, our localization objective can be formulated as:
\begin{equation} 
\label{eq:simple_local}
\mathcal{L}_{loc}=\mathcal{L}_{bce}(y^{match}, {GMP}(l(g(a_i^s),f(v_j^s)))),
\end{equation}
where $l(g(a_i^s),f(v_j^s))$ is the audiovisual localization function, $y^{match}$ indicates whether the audio and image come from the same pair, i.e., $y^{match}=1$ when $i=j$, otherwise $y^{match}=0$, and $\mathcal{L}_{bce}$ is the binary cross-entropy loss.
Similar to~\cite{arandjelovic2017objects}, we employ \emph{Global Max Pooling} (GMP) to spatially aggregate the audiovisual localization map to fit the scene-level supervision. We optimize the localization model and audiovisual backbones via the correspondence between audio and visual information without resorting to extra semantic annotation.

\subsection{Visual Object Representation Learning}

Since the obtained audiovisual localization map could figure out the approximated region of the sounding object, we use it to filter out the disturbance of the visual background and facilitate the perception of the object appearance. To learn effective object representation via their produced sound, we employ the localization results to purify the visual object representation and establish a category-representation object dictionary $D$ that covers different object categories. To do this, we utilize the localization map $l_i\in [0,1]^{H \times W}$ of the $i$-th audiovisual pair as weight, which indicates the importance of each patch, to perform weighted pooling on the visual feature map $f(v_i^s)$.
In this way, we manage to extract the candidate object representation $o_i \in\mathbb{R}^{1\times C}$ of $i$-th visual scene $f(v_i^s)$ with global weighted pooling, i.e.,
\begin{equation} \label{eq:stage1_obj}
o_i = \frac{\sum_{m,n} f(v_i^s)(m,n) \circ l_i(m,n)}{\sum_{m,n} l_i(m,n)},
\end{equation}
where $m$ and $n$ represent map entries, $\circ$ is the Hadamard product.
Since the set of object representations $\mathcal{O} = \left\{o_1, o_2,...,o_{N^s}\right\}$ is extracted from the coarse audiovisual localization maps, it is not robust enough to express the visual object characters.
We propose a dictionary learning scheme to extract high-quality class-aware visual object indicators from these coarse candidate representations to address such a challenge.
Specifically, we target to jointly learn a dictionary $D \in \mathbb{R}^{K \times C}$ and pseudo category assignment $y_i$ of each object representation $o_i$. Note that each key of the learned dictionary $d^k \in \mathbb{R}^{1 \times C}$ denotes the representative visual character of $k-$th category.
Considering that K-means clustering can be leveraged as an efficient way to establish the representation dictionary~\cite{coates2012learning}, we learn $D$ and $y_i$ by minimizing the following equation,
\begin{align} \label{eq:stage1_cluster}
\mathcal{L}_{clu}(D, y_i) &= \sum_{i=1}^{N^s} \mathop{min}\limits_{y_i} ||o_i - D^{T} \cdot y_i||_2^2\\
s.t. &{\rm{~~}} y_i \in \left\{0,1 \right\}^K, \sum y_i = 1,
\end{align}
where $K$ corresponds to the number of cluster centers. In practice, the number of cluster centers is set to a large number, no less than the number of categories, to ensure that different categories can be separated. By solving this problem, we obtain the ideal dictionary $D^*$ with a set of pseudo category assignments $\left\{ y_i^*|i= 1,2,...N^s \right\}$. In this way, $D^*$ can be used to retrieve potential objects in cocktail-party scenarios, and $y_i^*$ can be regarded as pseudo labels that indicate different object categories. 

Considering that generalized categorization could benefit object localization~\cite{oquab2015object,zhou2016learning}, we, therefore, propose a pseudo-label guided classification task to optimize the audiovisual backbones with classification networks with Eq.~\ref{eq:classification}:
\begin{align}
    \mathcal{L}_{cls} = \mathcal{L}_{ce}(y_i^*,h_a(a_i^s)) + \mathcal{L}_{ce}(y_i^*,h_v(v_i^s)),
    \label{eq:classification}
\end{align}
where we use two multi-layer perceptions (MLP) to instantiate audio and visual classification head denoted as $h_a$ and $h_v$, $\mathcal{L}_{ce}$ is cross-entropy loss function. This classification training process could substantially facilitate audiovisual feature discrimination.

\begin{figure}
    \centering
    \includegraphics[width=\linewidth]{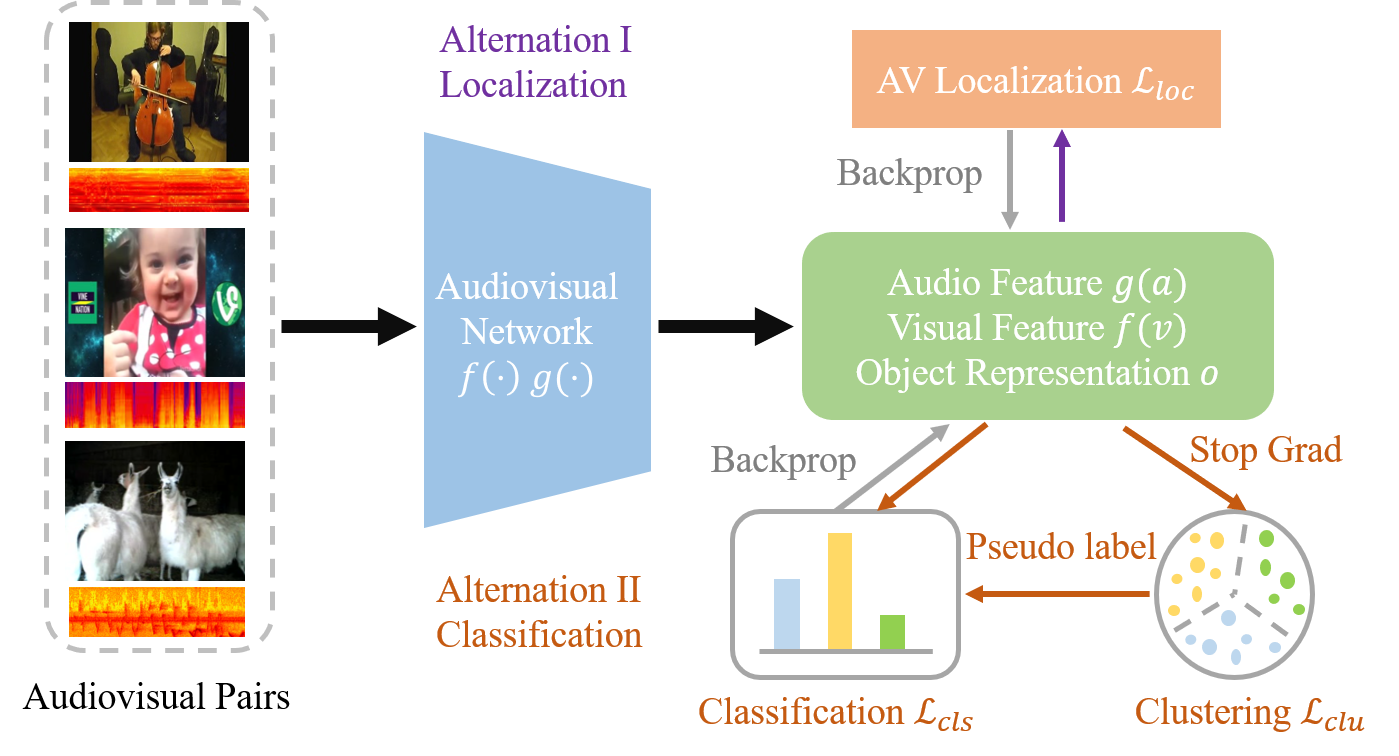}
    \caption{\textbf{Details of alternative localization-classification learning scheme.} We alternate between audiovisual localization and feature classification, where the localization loss $\mathcal{L}_{loc}$ and classification loss $\mathcal{L}_{cls}$ are back propagated, and there is no gradient w.r.t. the clustering objective $\mathcal{L}_{clu}$. }
    \label{fig:altoptim}
\end{figure}
In order to fully excavate the multi-modal knowledge contained in simple scenes $\mathcal{X}^s$, we propose an alternative localization-classification learning approach w.r.t. the localization objective in Eq.~\ref{eq:simple_local} and the classification objective in Eq.~\ref{eq:classification}. The optimization details are shown in Fig.~\ref{fig:altoptim}, where we alternate between scene-level audiovisual localization and pseudo-label guided classification. Note that the second alternation consists of two steps, first clustering, and second classification, we stop the gradient of clustering objective $L_{clu}$ in Eq.~\ref{eq:stage1_cluster} and only classification loss $L_{cls}$ is back propagated. This alternative learning strategy helps to progressively enhance audiovisual correspondence for object area refinement and improve feature discrimination of audiovisual backbones.

\subsection{Discriminative Sounding Objects Localization}

To achieve class-aware multiple sounding objects localization in cocktail-party scenarios, we propose a progressive approach to first perceive all objects in the visual scene, then based on whether they locate in the sounding area to determine which ones are sounding. Finally, we propose to leverage fine-grained audiovisual consistency on category-level to align sounding object distribution with the corresponding audio distribution. The overview is shown in the right part of Fig.~\ref{fig:framework}.

\begin{figure}
    \centering
    \includegraphics[width=\linewidth]{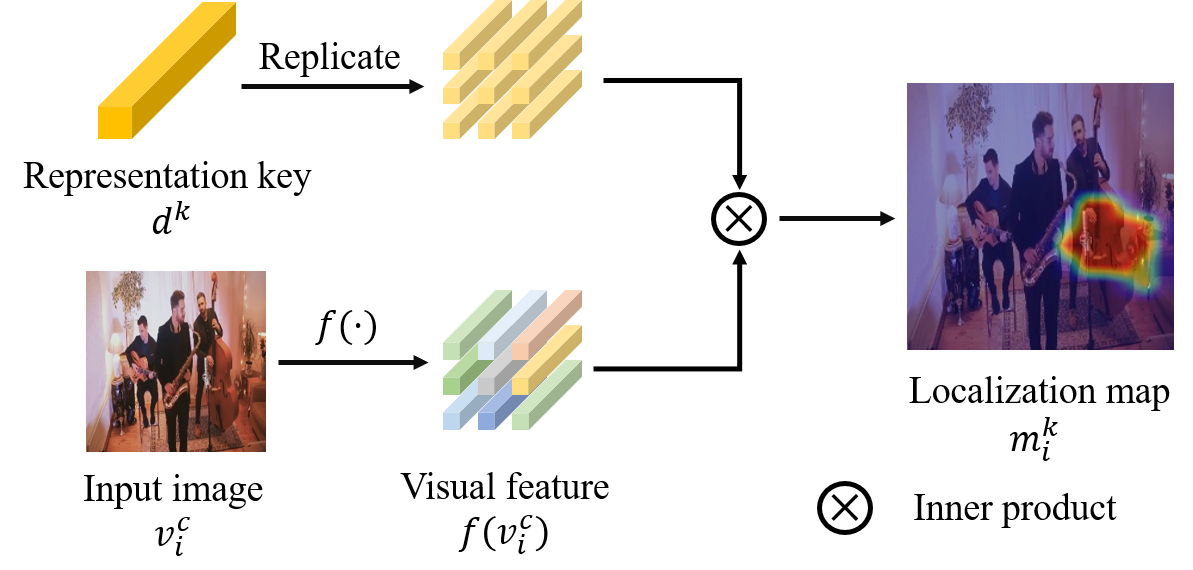}
    \caption{\textbf{Details of inner-product between object representation key and visual feature.} We take the cocktail-party music scene and object representation key of cello as an example. With image feature map $f(v_i^c)\in\mathbb{R}^{C\times H\times W}$ and the cello representation key $d^k\in\mathbb{R}^{C}$, we first spatially replicate the vector $d^k$ for $H\times W$ times, then perform inner-product between visual feature and replicated representation key on each spatial grid to obtain the cello localization map $m_i^k$.}
    \label{fig:prod}
\end{figure}

We denote the $i-$th audiovisual pair that contains multiple sounding objects, i.e., cocktail-party scenarios, as $(a_i^c, v_i^c) \in \mathcal{X}^c$. With reference to established category-representation object dictionary $D^*$, the objects in the visual scene can be indicated by the inner-product between the representation key $d^k \in \mathbb{R}^{1 \times C}$ in $D^*$ and each spatial grid of the visual feature map $f(v_i^c) \in \mathbb{R}^{C \times H \times W}$.
\begin{equation} \label{eq:stage2_loc}
m_i^k = d^k \cdot f(v_i^c),
\end{equation}
where $m_i^k$ is a probability map that indicates the potential object area of the $k-$th category in the $i-$th visual scene. The detailed operation is shown in Fig.~\ref{fig:prod}.
In this way, we could obtain $K$ different localization maps, each indicating object locations of a specific category. And if there does not exist any object belonging to the $k-$th category in the visual scenario, the corresponding localization map $m_i^k$ is likely to be in low response on each spatial grid.

As stated above, in cocktail-party scenarios, there could be multiple sounding objects as well as silent ones. To highlight the sounding objects while suppressing silent ones, we utilize the inferred global audiovisual localization map $l_i$ that indicates the coarse sounding area as a sounding object filter formulated as
\begin{equation} \label{eq:stage2_filter}
s^k_i = m^k_i \circ l_i.
\end{equation}
$s^k_i$ is regarded as the location of $k-$th category sounding objects. Intuitively speaking, when the $k-$th object exists in the visual scene but does not produce sound, the activation value on each spatial grid of $s^k_i$ will be low. In the way, we manage to suppress the silent visual objects and formulate the category-level sounding object distribution in $v_i^c$ as 
\begin{equation} \label{eq:stage2_so}
p^{so}_{v_i} = softmax([{GAP}(s^1_i), {GAP}(s^2_i),..., {GAP}(s^K_i)]),
\end{equation}
where GAP is the Global Average Pooling operation and the softmax regression is performed along the category dimension. 

As discussed in recent work~\cite{hu2019deep}, the natural synchronization between co-occurring audio and visual message not only provides coarse-grained audiovisual correspondence but also contributes to the fine-grained category-level distribution consistency in terms of mixed sound and sounding objects. That is, the sound property and the visual character of the sounding object correspond in taxonomy. 
Hence, similar to sounding object distribution $p^{so}_{v_i}$, based on the audio classification network trained in the first stage, we model the category-level distribution of sound $a_i^c$ as 
\begin{align}
    p^{so}_{a_i} = softmax(h_a(g(a_i^c))).
\end{align}
And the following learning objective is proposed to align the audiovisual distribution for discriminatively localizing sounding objects,
\begin{equation} \label{eq:stage2_kl}
\mathcal{L}_c = \mathcal{D}_{KL}(p^{so}_{v_i} || p^{so}_{a_i}),
\end{equation}
where $\mathcal{D}_{KL}$ is the Kullback–Leibler divergence. In this way, we manage to leverage category-level audiovisual consistency as the supervision to transform scene-level correspondence into fine-grained distribution matching.

To sum up, the second stage covers two learning objectives the one is class-agnostic audiovisual correspondence, and the other one is category distribution matching between mixed sound and detected sounding objects, i.e.,
\begin{equation} \label{eq:stage2_obj}
\mathcal{L}_2 =  \mathcal{L}_c + \lambda \cdot \mathcal{L}_{loc},
\end{equation}
where $\lambda$ is the hype parameter to balance the importance of two objectives.
By solving Eq.~\ref{eq:stage2_obj}, we manage to discriminatively reveal the location of different sounding objects in terms of a set of activation maps $\left\{s^1_i,s^2_i,...,s^K_i\right \}$. Since each spatial location only corresponds to one visual category, the final class-specific localization maps $S_i\in\mathbb{R}^{K\times H\times W}$ are obtained by performing softmax regression along category dimension on each spatial grid as
\begin{align}
    S_i(m,n) = softmax([s^1_i(m,n),s^2_i(m,n),...,s^K_i(m,n)]),
\end{align}
where $m$ and $n$ indicate spatial index. In this way, the activation will be high on sounding object area of corresponding categories, and remain low on background or silent area.

\section{Experiments}

\subsection{Dataset}
\textbf{MUSIC}
MUSIC dataset~\cite{zhao2018sop} contains 685 untrimmed videos, including 536 solo and 149 duet. 11 classes of musical instruments are covered. We select the first five/two videos of each instrument category in solo/duet for testing. The remaining videos are used for training. Furthermore, half of the individual training data is used for the first phase training, and the other half is used to generate synthetic data for the second phase training. Note that we finally obtain 489 solo videos and 141 duet videos because some videos are not available now on YouTube. 

\textbf{MUSIC 21}
MUSIC 21 dataset~\cite{som} is an extended version of MUSIC dataset, which extends the number of musical instruments to 21 categories. The number of solo videos increased to 1,164, while duet videos remained unchanged. We finally obtain 1,096 solo videos. We select the first five videos of each instrument category in solo videos for testing. The remaining videos are used for training.

\textbf{MUSIC-Synthetic}
The unbalanced instrument category distribution of duet videos of MUSIC dataset brings great difficulty to training and bias to testing, e.g., more than 80\% duet videos contain the sound of the guitar. Therefore, we artificially synthesize solo videos to build category-balanced multi-source videos for our second-stage learning and evaluation. Go in detail, we first randomly select four 1-second solo audiovisual pairs of different categories, then get the multi-source video frames by means of mixing random two audio clips with jittering, and multi-source video frame is synthesized through concatenating four frames of these clips. That means there are two sounding instruments and two silent instruments in the synthesized audiovisual pair. Thus, this synthesized dataset is quite appropriate for the evaluation of class-aware sounding object localization. Finally, we generate 24,978 audiovisual pairs for training and 456 pairs for testing.

\textbf{AudioSet-instrument}
Because the MUSIC dataset is relatively clean, we chose the noisier AudioSet-instrument dataset to validate the model performance further. AudioSet-instrument dataset is a subset of AudioSet~\cite{2017audioset}, containing 63,989 10-second video clips. 15 categories of instruments are covered. Following~\cite{gao2019coseparation}, videos from the "unbalanced" split are used for training, and those from the "balanced" split are used for testing. The solo videos with a single sound source are utilized for the first-stage training and testing, and videos with multiple sound sources are utilized for the second-stage training and testing. 

\textbf{VGGSound dataset \& VGGSound-Synthetic}
Besides music scenes, we use the VGGSound dataset~\cite{vedaldi2020vggsound} to train and evaluate our method in general cases. VGGSound dataset contains over 210k single source videos, covering 310 audio categories in daily life scenes. To better evaluate sounding object localization ability, we filter out a subset of 98 categories whose sound source can be visually localized in the VGGSound validation split for quantitative evaluation. This subset is quite challenging and provides diverse audiovisual scenes for class-aware sounding object localization. Like the MUSIC dataset, we use half of them as simple scene data and the other half to synthesize VGGSound-Synthetic. For the VGGSound, we finally take 14,349 videos for training and 2,787 for testing. The VGGSound-Synthetic dataset contains 40,000 train samples and 500 test samples.

\textbf{Realistic MUSIC \& DailyLife dataset}
To further evaluate the audiovisual model in real-world scenes, we collect 70 real multi-source instrument videos and 100 real multi-source daily life scenes on YouTube. Each instrument video includes 3 to 4 instruments and each daily life scene includes 2 events. 19 classes of musical instruments and 44 categories in the VGGSound dataset are covered. Because of the high noise of the real data, we perform manual labeling for bounding box annotation. We select one second from each video and manually label the corresponding frame, marking out the bounding boxes and whether each object is sounding or not.

\textbf{Bounding box annotation}
For the annotation of instrument dataset, including MUSIC, MUSIC 21, and AudioSet-instrument, a Faster RCNN detector w.r.t 15 instruments~\cite{gao2019coseparation}, which is trained with Open Images dataset~\cite{krasin2017openimages}, is utilized to generate bounding boxes on the test set for quantitatively evaluating the sound localization performance. Note that for the MUSIC/MUSIC 21 dataset, only the categories contained in the Faster RCNN detector are used for quantitative evaluation (7 categories in MUSIC and 10 categories in MUSIC 21). The categories of the AudioSet-instrument dataset and Faster RCNN detector overlap exactly. 7 categories in the MUSIC dataset that Faster RCNN can identify are used to generate the MUSIC-Synthetic dataset. Besides generating bounding boxes by Faster RCNN detector on the MUSIC-Synthetic dataset, we refine the detection results and manually annotate whether each object is sounding or not. We put annotations in the released code to facilitate reproducibility.

The annotation of the test set of the VGGSound dataset is also done with the help of a Faster RCNN detector, which is trained on the COCO dataset~\cite{lin2014microsoft}. Note that the sounding objects of the 98 audio event categories we selected from the original VGGSound dataset are all included in COCO. 

\subsection{Experimental Settings}

\textbf{Implementation details}
We divide each video in the above dataset into one-second clips equally, with no intersection. A randomly selected image from the clip is used as visual information. The image size is adjusted to $256 \times 256$, and then it is cropped randomly to  $224 \times 224$. The audio messages are processed as follows. We firstly re-sample the audio into 16K Hz and then translate it into spectrogram employing Short Time Fourier Transform. The Hann window length is 160, and the hop length is 80. To represent sound characteristics better, we perform Log-Mel projection over the spectrogram, similarly with ~\cite{zhao2018sop,hu2019deep}. After the above processing, the audio messages become a $201 \times 64$ matrix. A matched pair is defined as visual and audio message from the same video clip, otherwise mismatched. Variants of ResNet-18~\cite{resnet} are used as visual and audio feature extractors. The ImageNet-pretrain model is used as visual feature extractor and the audio feature extractor is trained from scratch. We train our model using Adam optimizer with learning rate of $10^{-4}$. During the training phase, a threshold of 0.05 is used to binarize the localization maps aiming to get the object mask. The object representations can be extracted over feature maps with the object mask. In the experiment, the number of cluster centers is set to a large number, no less than the number of categories, aiming to separate different categories. We accordingly assign each cluster center in the object dictionary to one object category.

Specifically, we use simple enumeration to construct cluster-to-category correspondence, which satisfies that each object category corresponds to at least one cluster center, and each cluster center corresponds to only one object category. For each cluster-to-category correspondence, we calculate the proportion of samples of assigned categories in each cluster center. The summation of the proportion values is the purity value of this cluster-to-category correspondence. The cluster-to-category correspondence with the largest purity value is used to generate a category-representation object dictionary. During multiple sounding objects localization, in the case of multiple cluster centers corresponding to the same category, the activation summation of these cluster centers is the final activation of this category. Note that we train and evaluate the proposed model on the same dataset. The code and dataset are available at \url{https://gewu-lab.github.io/CSOL_TPAMI2021/}.

\textbf{Evaluation metric}
\emph{Intersection over Union} (IoU) and \emph{Area Under Curve} (AUC), which are calculated based on the annotated bounding box and the predicted sounding area, are used as evaluation metrics for single source sounding object localization. 

\begin{figure*}
    \centering
    \includegraphics[width=0.32\linewidth]{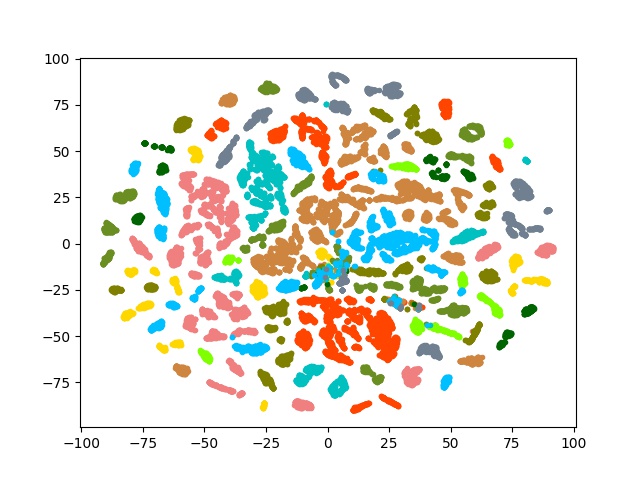}
    \includegraphics[width=0.32\linewidth]{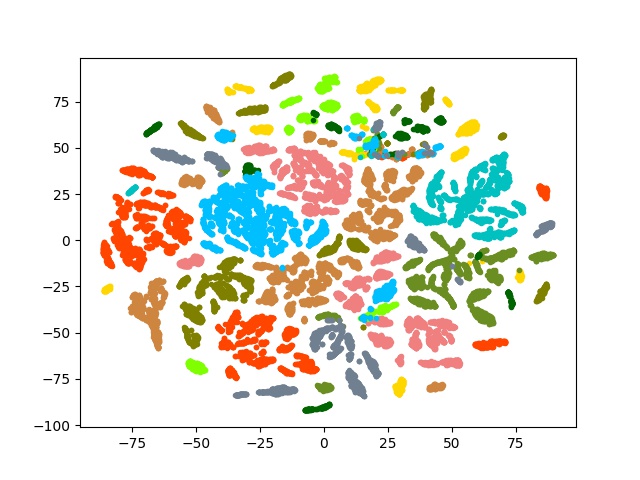}
    \includegraphics[width=0.32\linewidth]{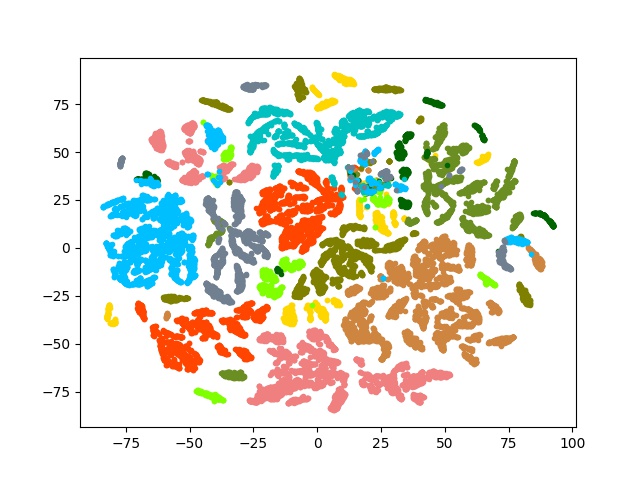}
    \vspace{-3mm}
    \caption{\textbf{Visual feature distribution visualized by t-SNE on MUSIC-solo.} These figures from left to right are global image features without alternative learning, image features with alternative learning, and masked object features with alternative learning. The categories are indicated in different colours.}
    \vspace{-1em}
    \label{fig:tsne}
\end{figure*}

\begin{table*}
  \caption{\textbf{Localization results on MUSIC-synthetic, MUSIC-duet, Realistic MUSIC and AudioSet-instrument-multi.} Note that the CIoU reported in this table is CIoU@0.3, and NSA of DMC is not evaluated since it relies on given knowledge to determine whether the object is sounding or silent.}
  \label{tbl:syn}
  \centering{
    \begin{tabular}{l|ccc|ccc|ccc|ccc}
      \hline
      Data              & \multicolumn{3}{c|}{MUSIC-Synthetic}          & \multicolumn{3}{c|}{MUSIC-Duet}               & \multicolumn{3}{c|}{Realistic MUSIC}         & \multicolumn{3}{c}{AudioSet-multi}            \\ \hline
      Methods           & CIoU@0.3      & AUC           & NSA           & CIoU@0.3      & AUC           & NSA           & CIoU@0.3     & AUC          & NSA           & CIoU@0.3      & AUC           & NSA           \\ \hline
      Sound-of-pixel~\cite{zhao2018sop}    & 8.1           & 11.8          & 97.2          & 16.8          & 16.8          & \textbf{92.0} & 3.2          & 6.1          & 94.3          & 39.8          & 27.3          & \textbf{88.8} \\
      Object-that-sound~\cite{arandjelovic2017objects} & 3.7           & 10.2          & 19.8          & 13.2          & 18.3          & 15.7          & 5.0          & 8.0          & 31.0          & 27.1          & 21.9          & 16.5          \\
      Attention~\cite{senocak2018learning}         & 6.4           & 12.3          & 77.9          & 21.5          & 19.4          & 54.6          & 3.4          & 6.2          & 51.2          & 29.9          & 23.5          & 4.5           \\
      DMC~\cite{hu2019deep}               & 7.0           & 16.3          & -             & 17.3          & 21.1          & -             & 3.7          & 7.1          & -             & 32.0          & 25.2          & -             \\
      Ours              & \textbf{32.3} & \textbf{23.5} & \textbf{98.5} & \textbf{30.2} & \textbf{22.1} & 83.1          & \textbf{7.1} & \textbf{8.2} & \textbf{98.3} & \textbf{48.7} & \textbf{29.7} & 56.8          \\ \hline
      \end{tabular}
  }
\end{table*}

We introduce two new metrics for the quantitative evaluation of discriminative sounding objects localization, \emph{Class-aware IoU} (CIoU) and \emph{No-Sounding-Area} (NSA). The average over class-specific IoU score is deemed as CIoU, and NSA is defined as the average activation area on localization maps of silent categories where the activation is below threshold $\tau$.
\begin{align}
  CIoU = \frac{\sum_{k=1}^K\delta_k IoU_k}{\sum_{k=1}^K\delta_k},\quad
\end{align}

\begin{align}
  NSA = \frac{\sum_{k=1}^K(1-\delta_k)\sum s^k<\tau}{\sum_{k=1}^K(1-\delta_k)A},\quad
\end{align}
where $IoU_k$ is calculated with the annotated bounding box and predicted sounding object area for the $k-$th class. $A$ is the total area of the localization map. $s^k$ is the localization map of $k$-th class. When class $k$ is making sound, $\delta_k=1$, otherwise $\delta_k=0$. These two evaluation metrics respectively measure the ability of the model to discriminatively localizing sounding objects and filtering out silent objects.

Besides, we use \emph{Normalized Mutual Information} (NMI) between cluster assignments and ground-truth category labels to measure the discrimination of the extracted feature. And for object detection task, we leverage \emph{Average Precision} (AP) in~\cite{lin2014microsoft} as standard metric for quantitative evaluation.

\subsection{Single Sounding Object Localization}
The simple single sounding object localization is mainly discussed in this subsection. Table~\ref{tab:solo} shows the experimental results on MUSIC-solo, MUSIC21 and AudioSet-instrument-solo videos. We compared our method with SOTA methods~\cite{arandjelovic2017objects,hu2019deep,senocak2018learning}. Note that the public source code from \cite{zhao2018sop,hu2019deep} is used. And there are two points to note in light of the results shown:

First, the training strategy of the compared methods~\cite{arandjelovic2017objects,hu2019deep,senocak2018learning} utilizes the contrastive~\cite{hu2019deep,senocak2018learning} or classification~\cite{arandjelovic2017objects} objective to match the correct audio-visual pair, whose training strategy is similar to ours. However, our performance is substantially superior to these methods. This phenomenon indicates that the feature representations learned through alternative localization-classification strategies are effective for semantic discrimination, which is conducive to the subsequent class-aware object localization based on the learned object dictionary. To explain this more clearly, we utilize the t-SNE method~\cite{maaten2008visualizing} to plot the distribution of extracted visual features from the well-trained vision network. Fig.~\ref{fig:tsne} illustrates that when training the model in an alternative manner, the extracted features on MUSIC-solo videos are more discriminative in terms of object categories, and the clustering with masked object features reveals high discrimination with NMI of 0.74.

Second, our method is comparable to Sound-of-pixel~\cite{zhao2018sop} on MUSIC-solo and AudioSet-instrument-solo dataset. Sound-of-pixel uses an audio-based mix-then-separate learning strategy to build audiovisual channel correlations. Therefore, when the training set contains fewer instrument categories, the algorithm can effectively associate a specific visual region with audio embedding and achieve good results. However, when the diverse of objects increases, the performance drops significantly on MUSIC 21. On the contrary, our method can tackle this challenge without constructing complex learning objective.

\begin{table}
\vspace{-1em}
  \caption{Localization results on MUSIC-solo, MUSIC 21 and AudioSet-instrument-solo.}  
  \centering  
  \subtable[Results on MUSIC-solo.]{  
    \begin{tabular}{l|cc}
      \hline
      Methods           & IoU@0.5       & AUC           \\ \hline
      Sound-of-pixel~\cite{zhao2018sop}    & 40.5          & 43.3          \\
      Object-that-sound~\cite{arandjelovic2017objects} & 26.1          & 35.8          \\
      Attention~\cite{senocak2018learning}         & 37.2          & 38.7          \\
      DMC~\cite{hu2019deep}               & 29.1          & 38.0          \\
      Ours              & \textbf{51.4} & \textbf{43.6} \\ \hline
      \end{tabular}
         \label{tab:music-solo}  
  }  
  \qquad
  \subtable[Results on MUSIC 21.]{  
    \begin{tabular}{l|cc}
      \hline
      Methods           & IoU@0.5       & AUC           \\ \hline
      Sound-of-pixel~\cite{zhao2018sop}    & 39.9          & 41.2          \\
      Object-that-sound~\cite{arandjelovic2017objects} & 30.5          & 42.9          \\
      Attention~\cite{senocak2018learning}         & 60.2          & 50.2          \\
      DMC~\cite{hu2019deep}               & 46.1          & 41.6          \\
      Ours              & \textbf{66.1} & \textbf{56.4} \\ \hline
      \end{tabular}
         \label{tab:music21-solo}  
  }  
  \qquad  
  \subtable[Results on AudioSet-instrument-solo.]{
    \begin{tabular}{l|cc}
      \hline
      Methods           & IoU@0.5       & AUC           \\ \hline
      Sound-of-pixel~\cite{zhao2018sop}    & 38.2          & 40.6          \\
      Object-that-sound~\cite{arandjelovic2017objects} & 32.7          & 39.5          \\
      Attention~\cite{senocak2018learning}         & 36.5          & 39.5          \\
      DMC~\cite{hu2019deep}               & 32.8          & 38.2          \\
      Ours              & \textbf{38.9} & \textbf{40.9} \\ \hline
      \end{tabular}
         \label{tab:audioset-solo}  
  }  
   \label{tab:solo}
  \end{table}
  
  \begin{figure}
    \centering
    \includegraphics[width=0.3\linewidth]{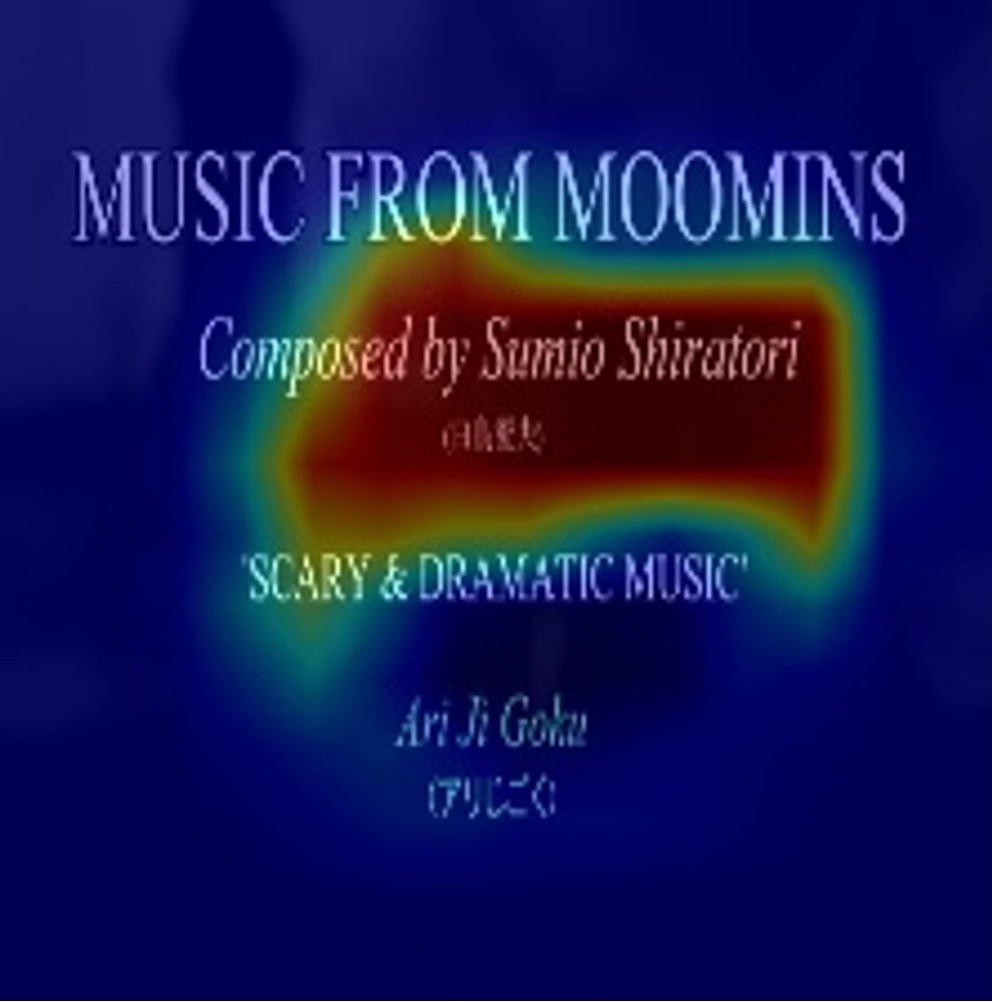}
    \includegraphics[width=0.3\linewidth]{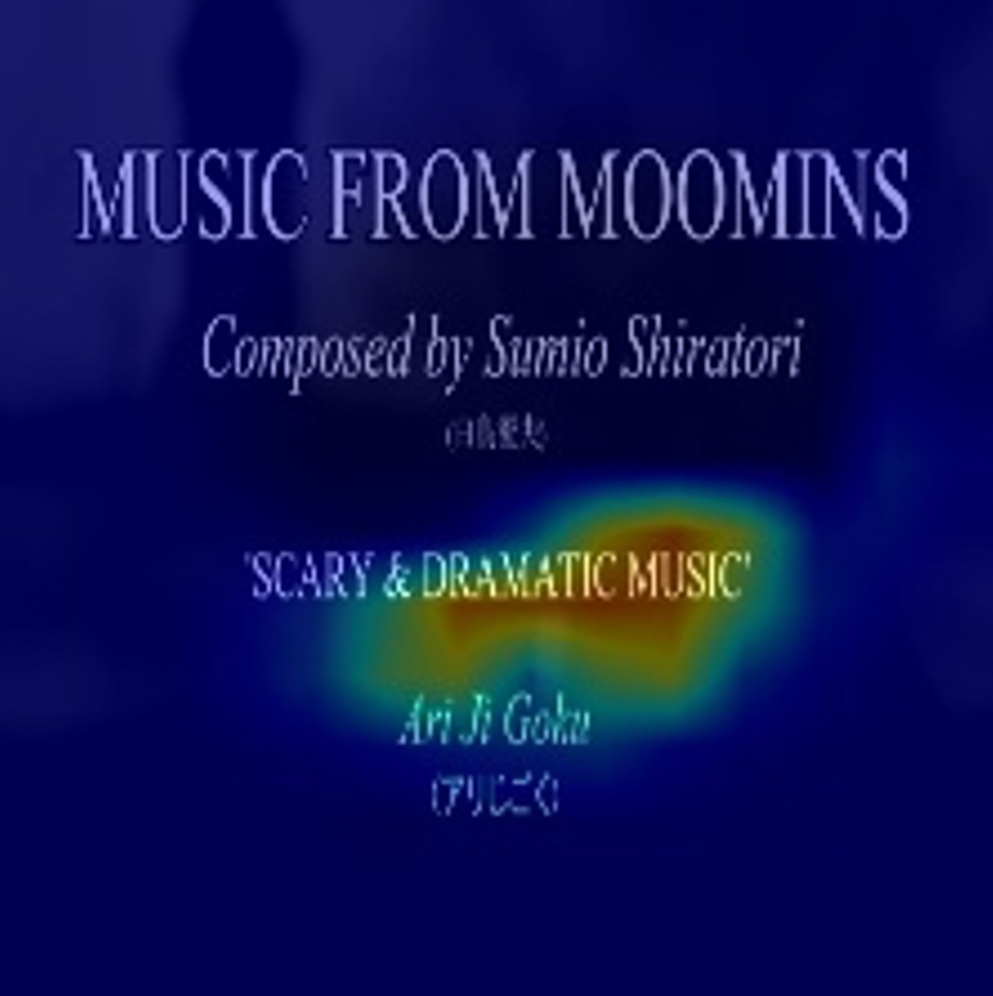}
    \includegraphics[width=0.3\linewidth]{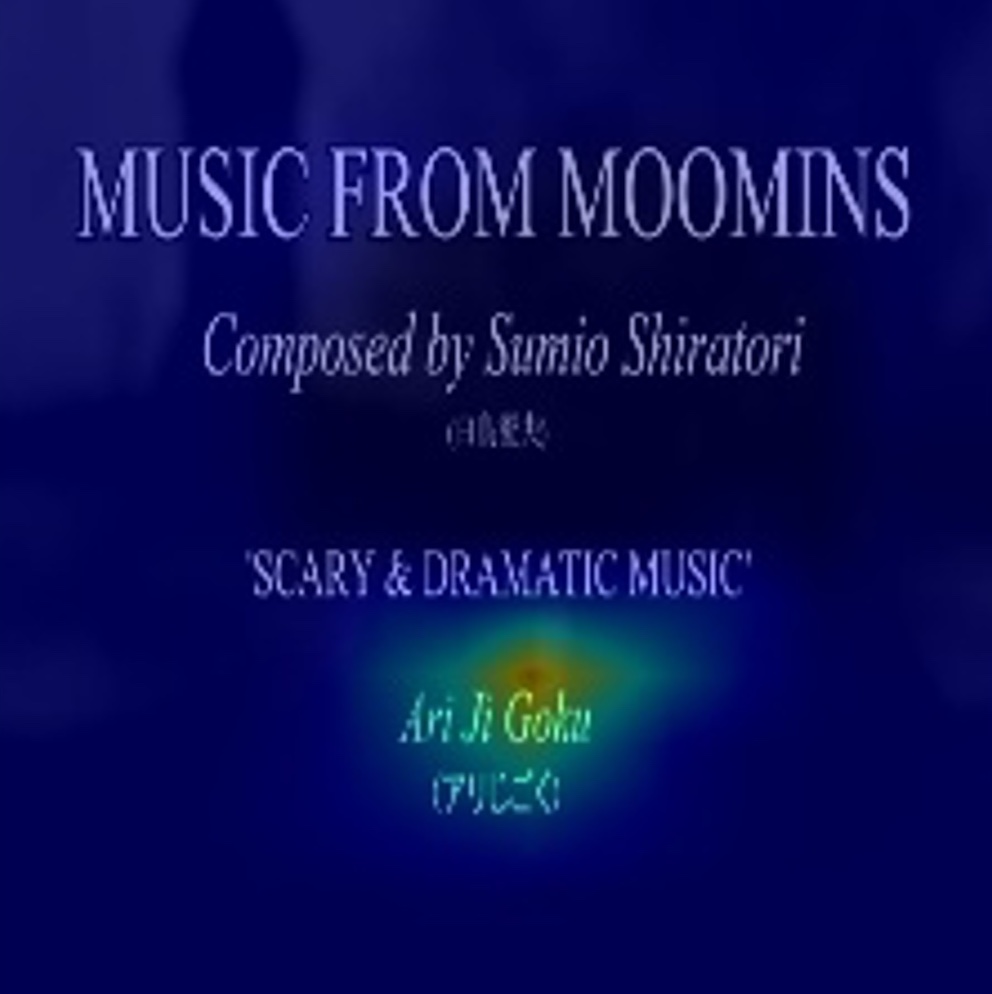}
    \caption{Failure case of out-of-screen sounding objects. From left to right are the activation of three categories in object dictionary (cello, trombone, and violin) to a picture that does not contain the visual information of the object (but with corresponding audio information).}
    \vspace{-1em}
    \label{fig:ablation-outscreen}
\end{figure}

\begin{figure*}
    \subfigure[Attention~\cite{senocak2018learning}]{
    \includegraphics[width=0.135\linewidth]{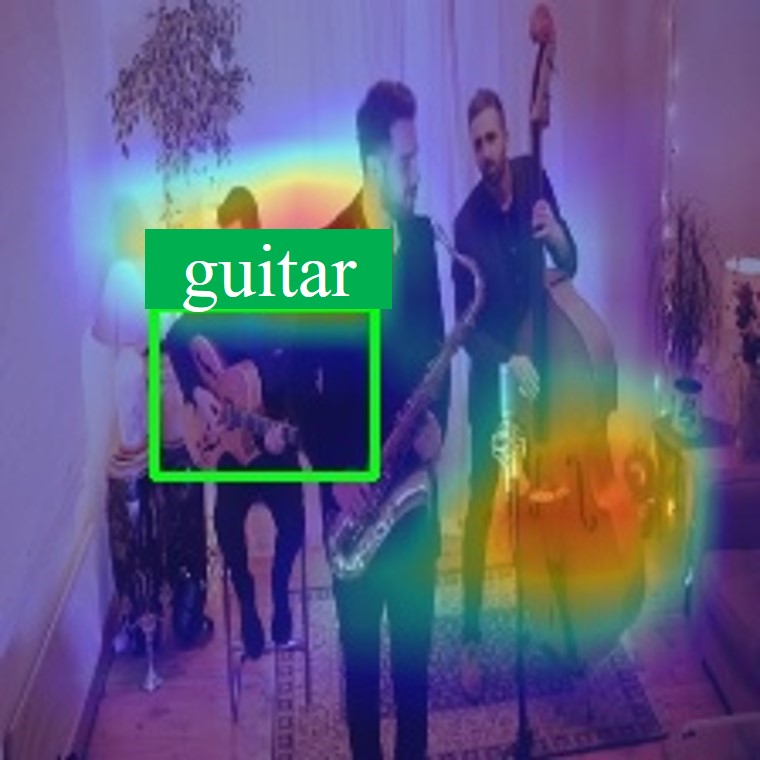}
    \includegraphics[width=0.135\linewidth]{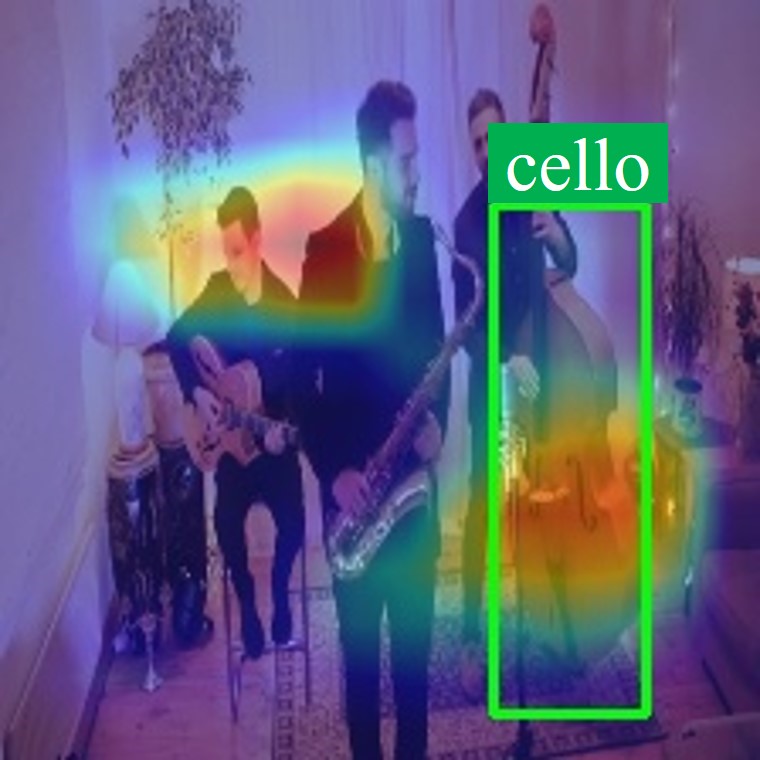}
    \includegraphics[width=0.135\linewidth]{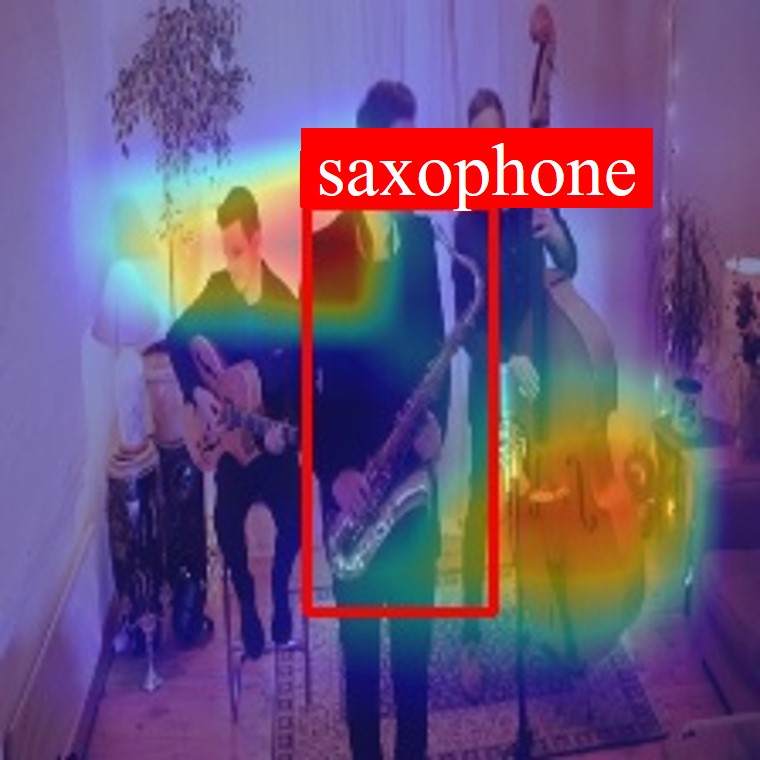}\hspace{1mm}
    \includegraphics[width=0.135\linewidth]{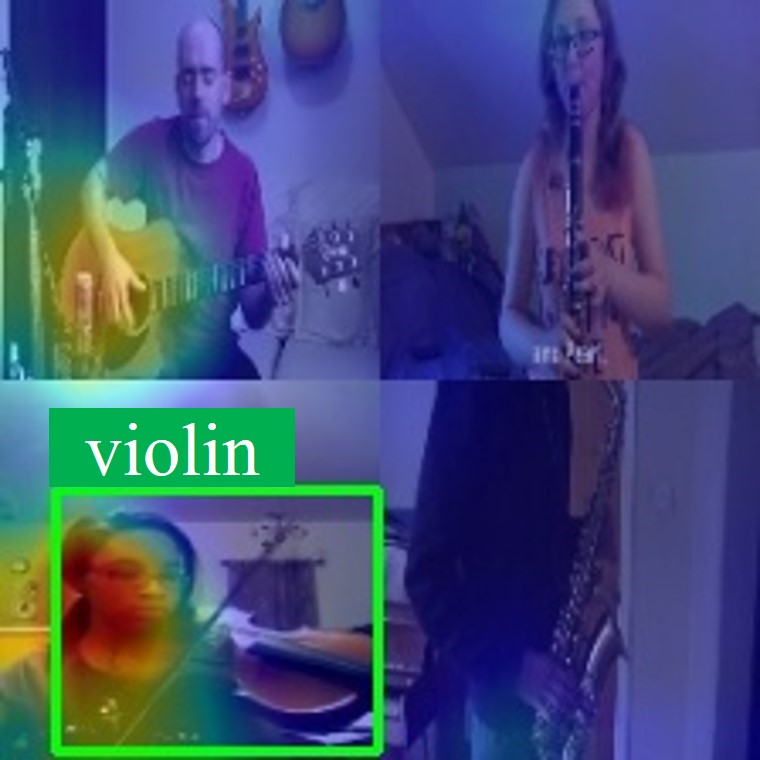}
    \includegraphics[width=0.135\linewidth]{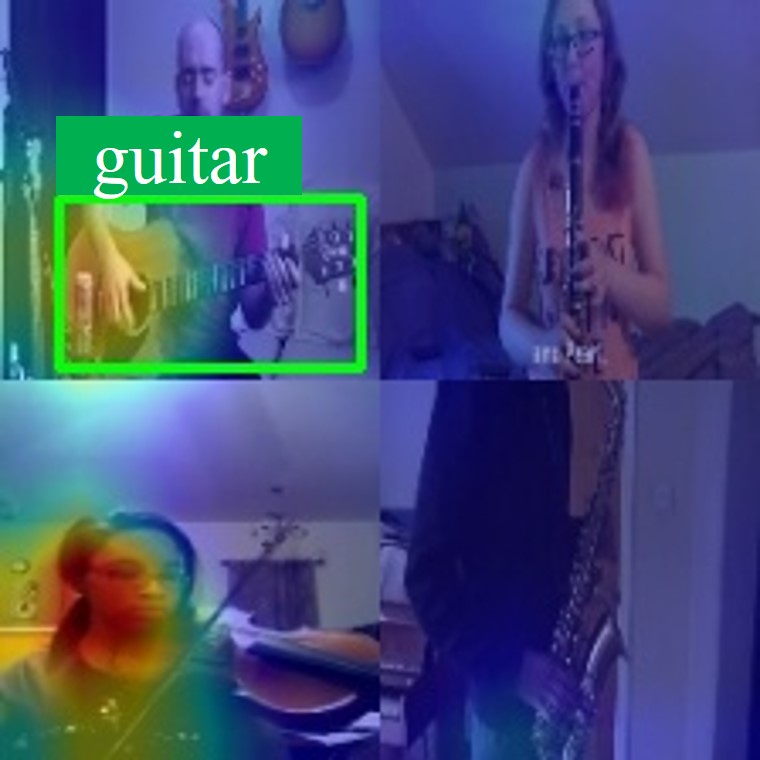}
    \includegraphics[width=0.135\linewidth]{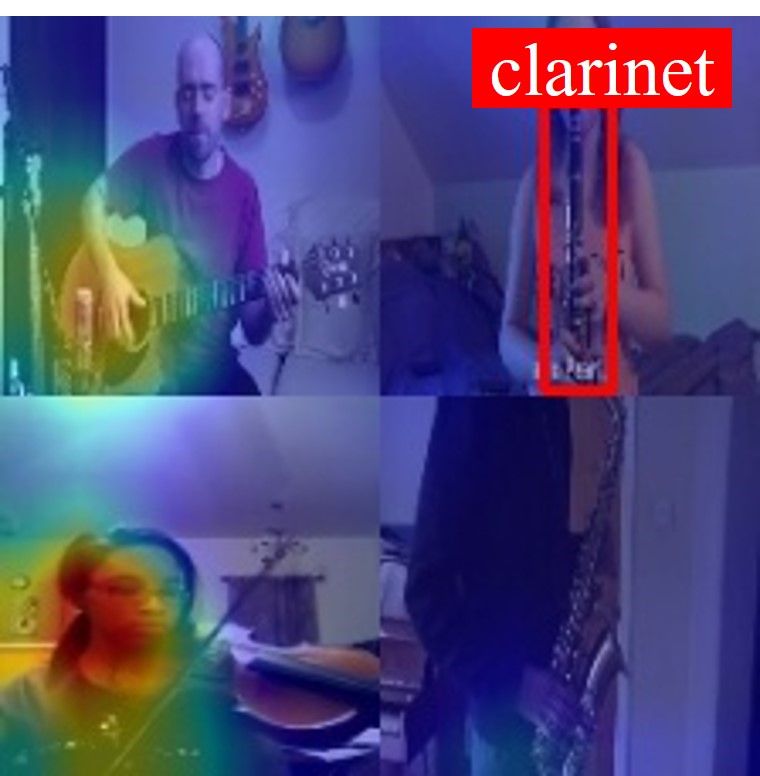}
    \includegraphics[width=0.135\linewidth]{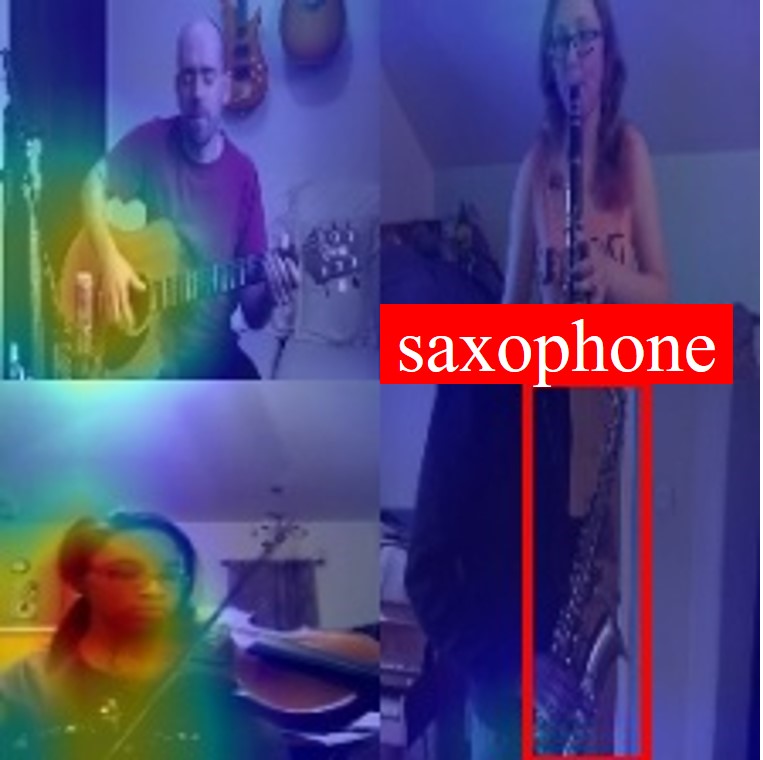}
    }
    \subfigure[Object-that-sound~\cite{arandjelovic2017objects}]{
    \includegraphics[width=0.135\linewidth]{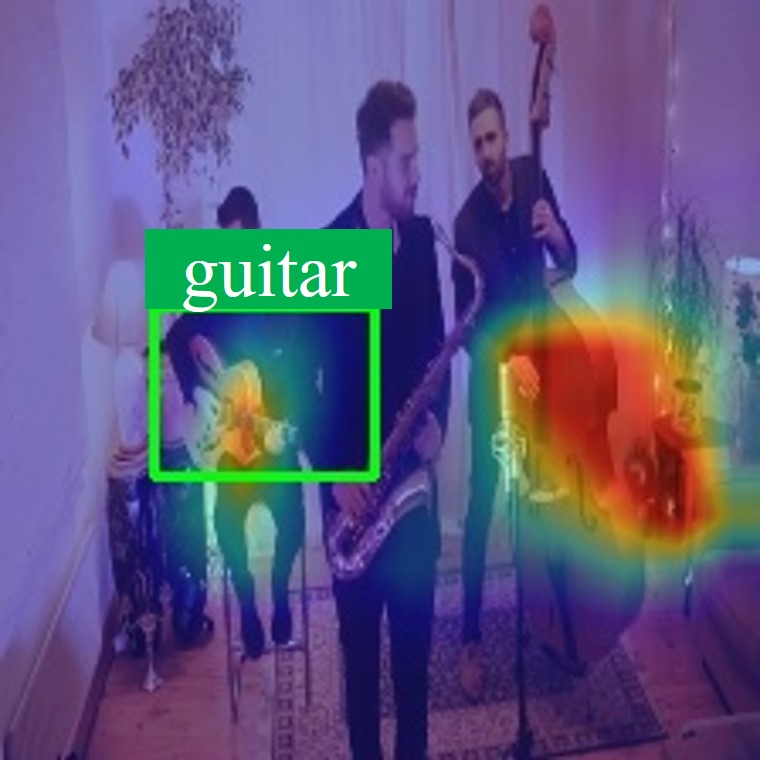}
    \includegraphics[width=0.135\linewidth]{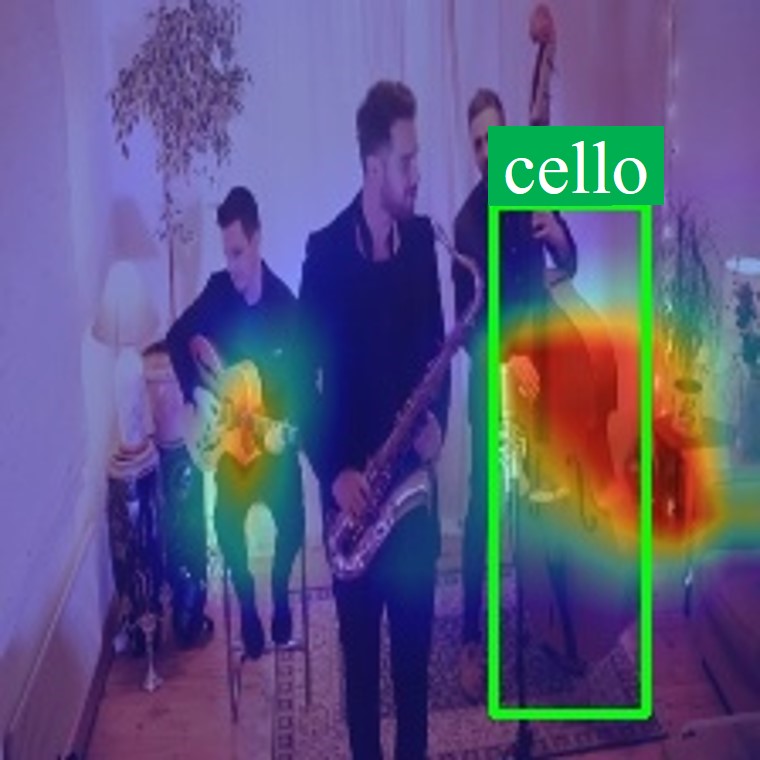}
    \includegraphics[width=0.135\linewidth]{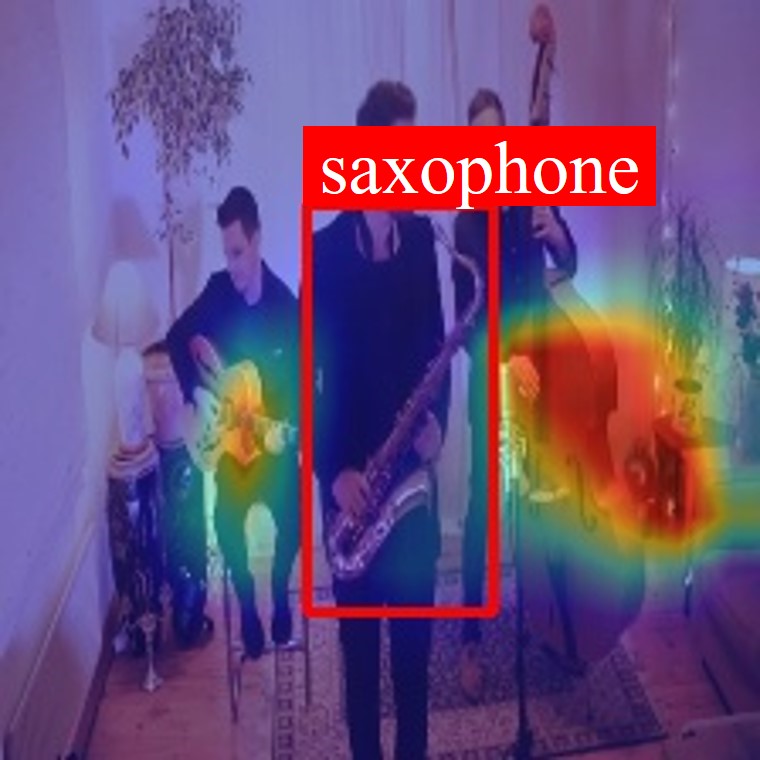}\hspace{1mm}
    \includegraphics[width=0.135\linewidth]{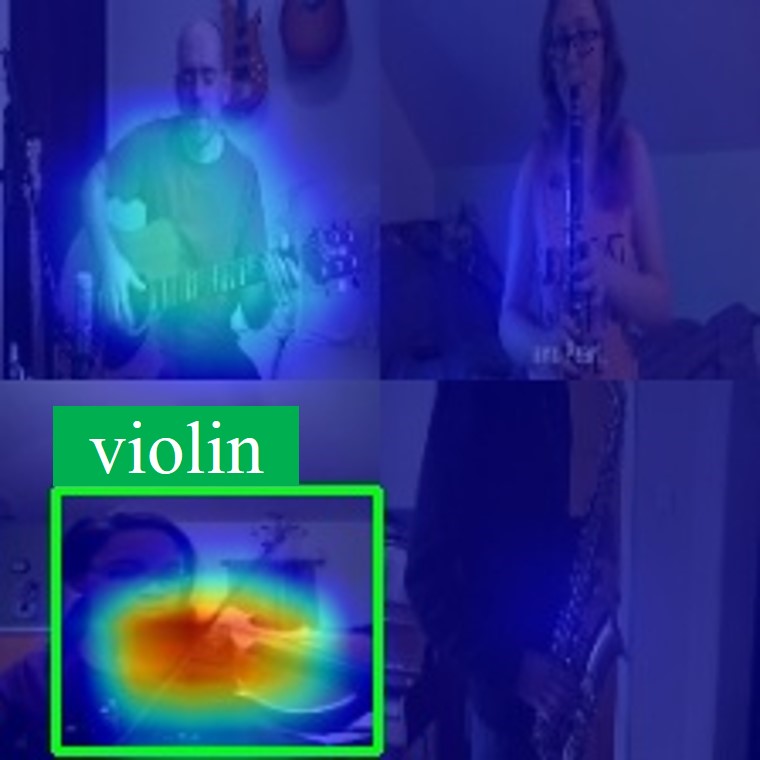}
    \includegraphics[width=0.135\linewidth]{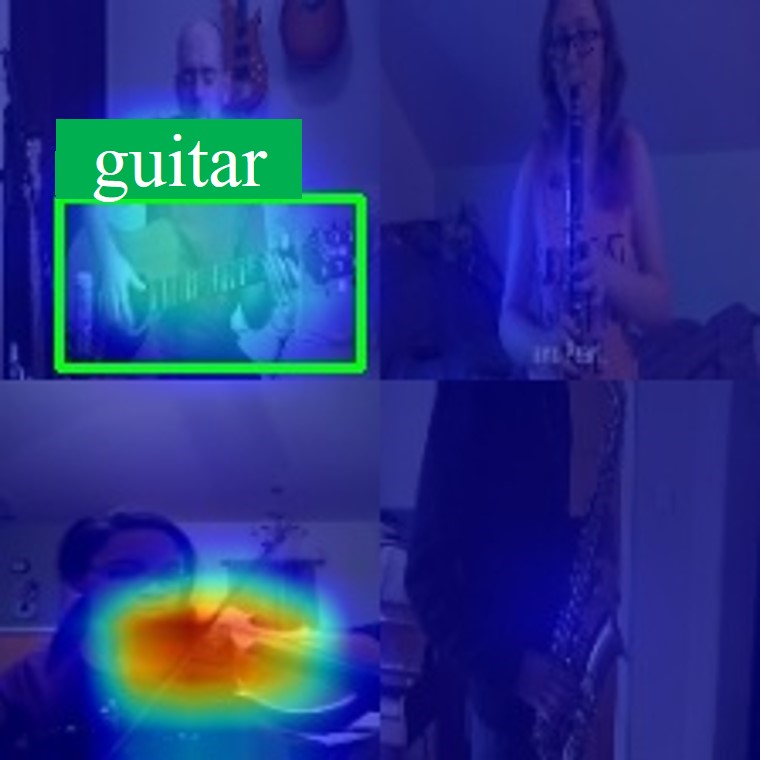}
    \includegraphics[width=0.135\linewidth]{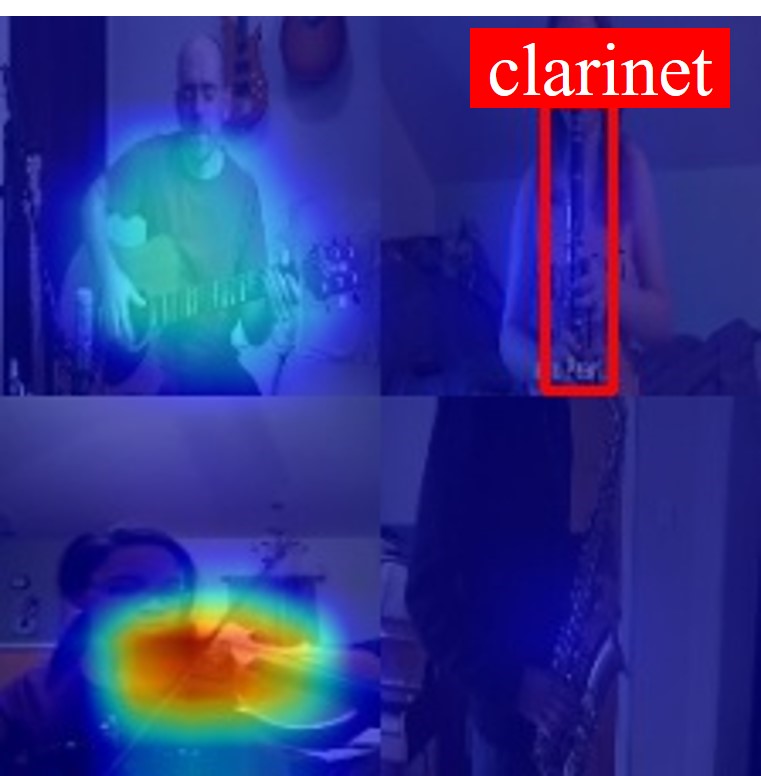}
    \includegraphics[width=0.135\linewidth]{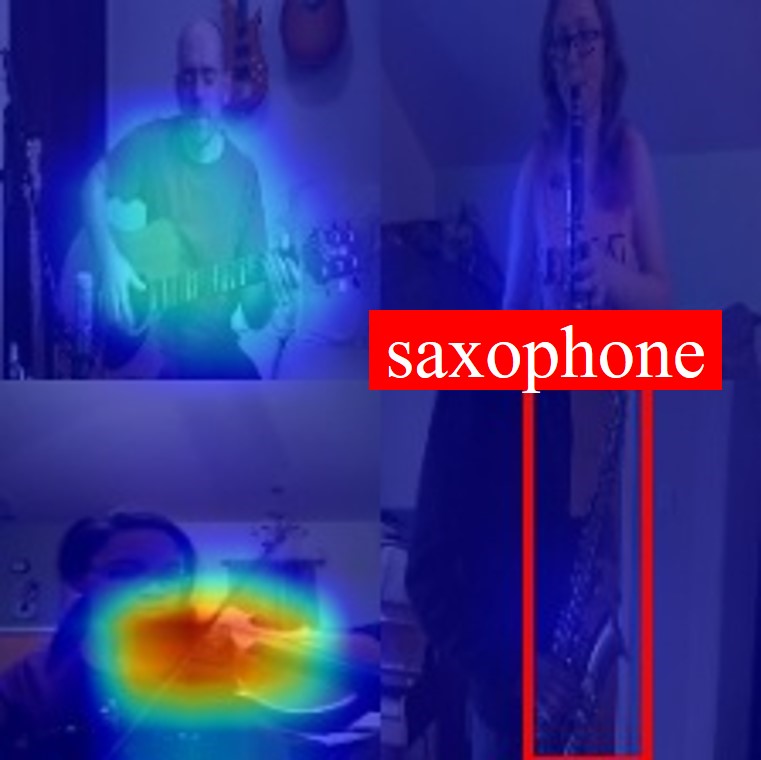}
    }
    \subfigure[DMC~\cite{hu2019deep}]{
    \includegraphics[width=0.135\linewidth]{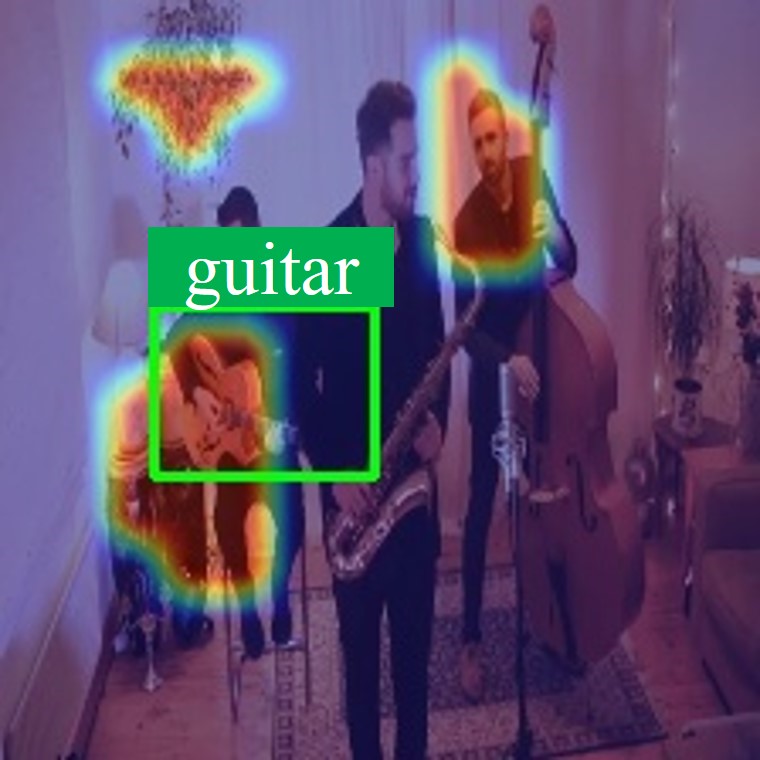}
    \includegraphics[width=0.135\linewidth]{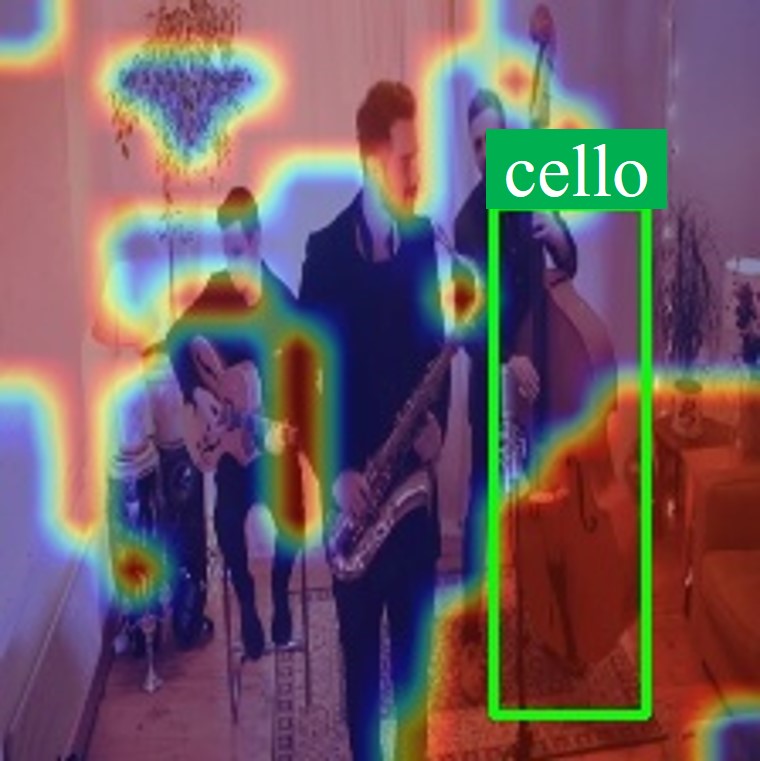}
    \includegraphics[width=0.135\linewidth]{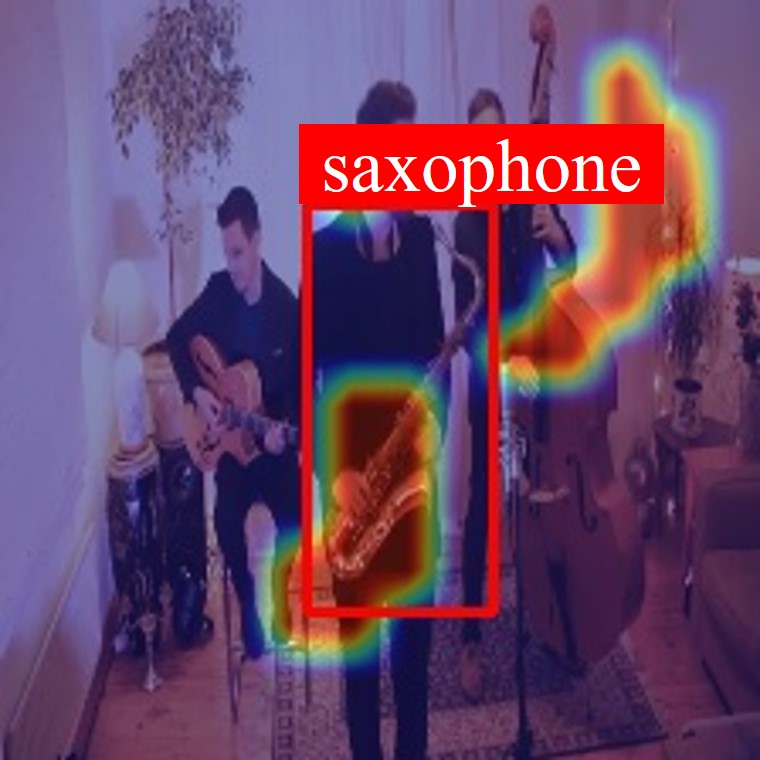}\hspace{1mm}
    \includegraphics[width=0.135\linewidth]{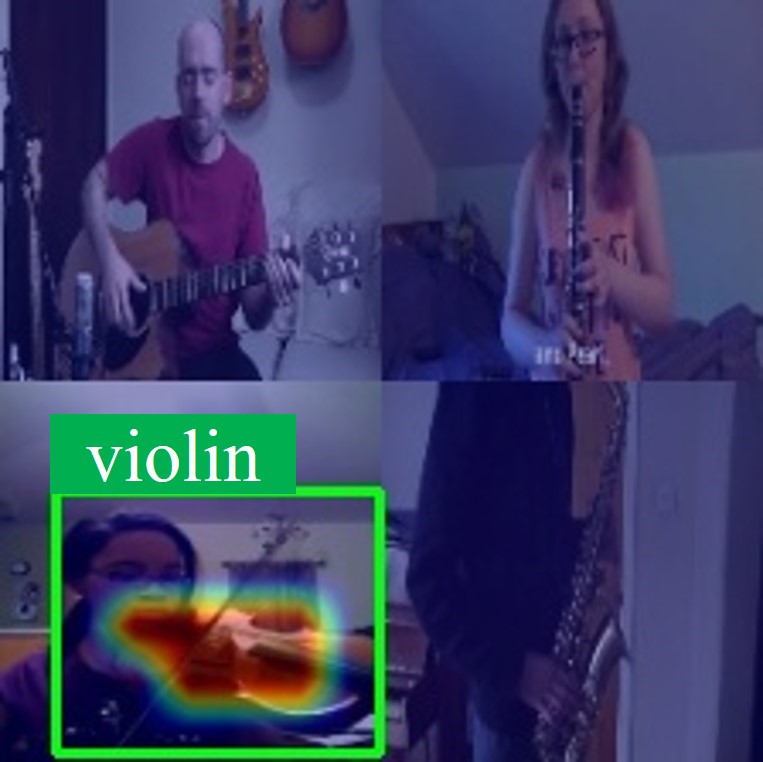}
    \includegraphics[width=0.135\linewidth]{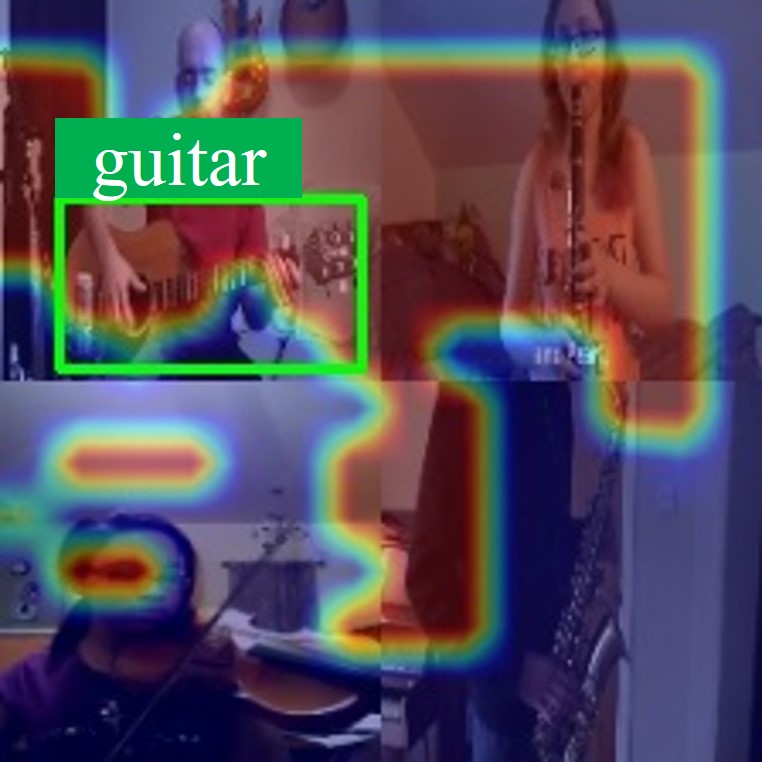}
    \includegraphics[width=0.135\linewidth]{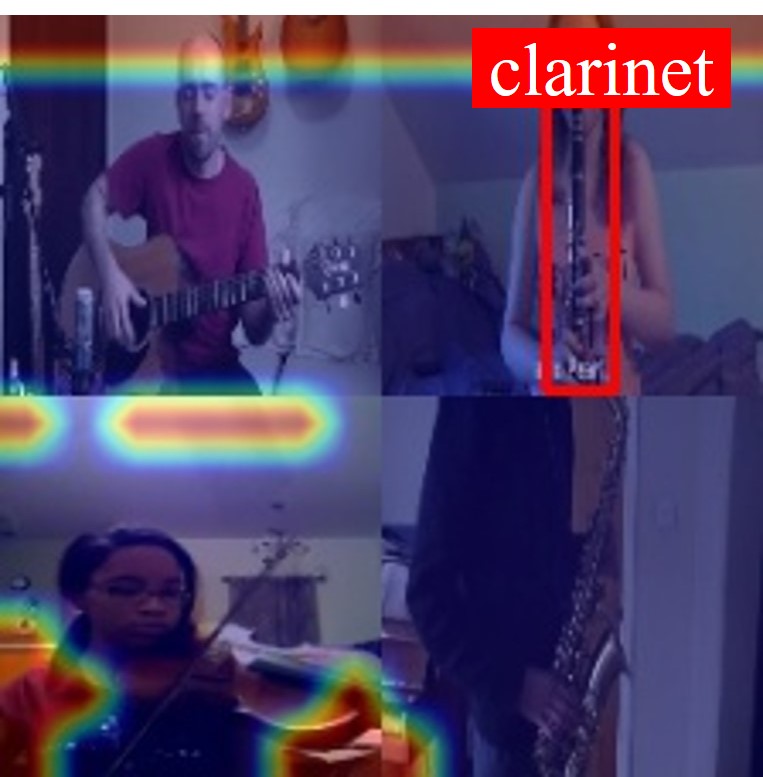}
    \includegraphics[width=0.135\linewidth]{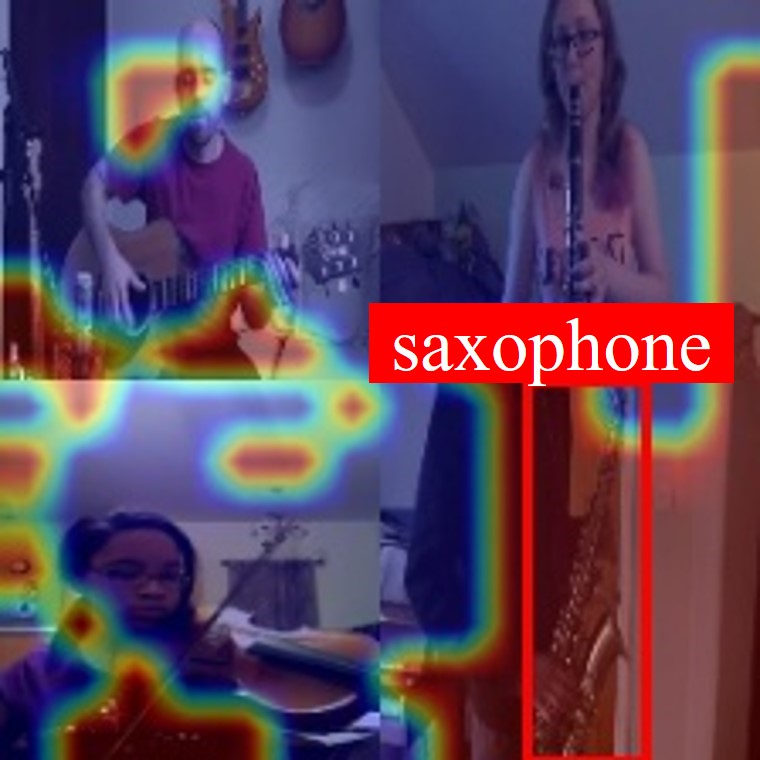}
    }
    \subfigure[Sound-of-pixel~\cite{zhao2018sop}]{
    \includegraphics[width=0.135\linewidth]{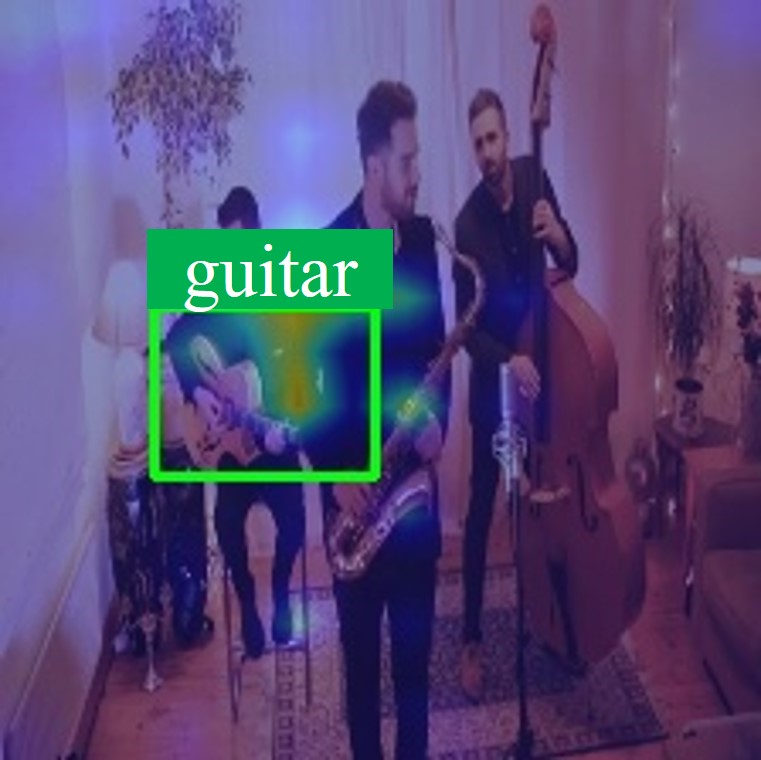}
    \includegraphics[width=0.135\linewidth]{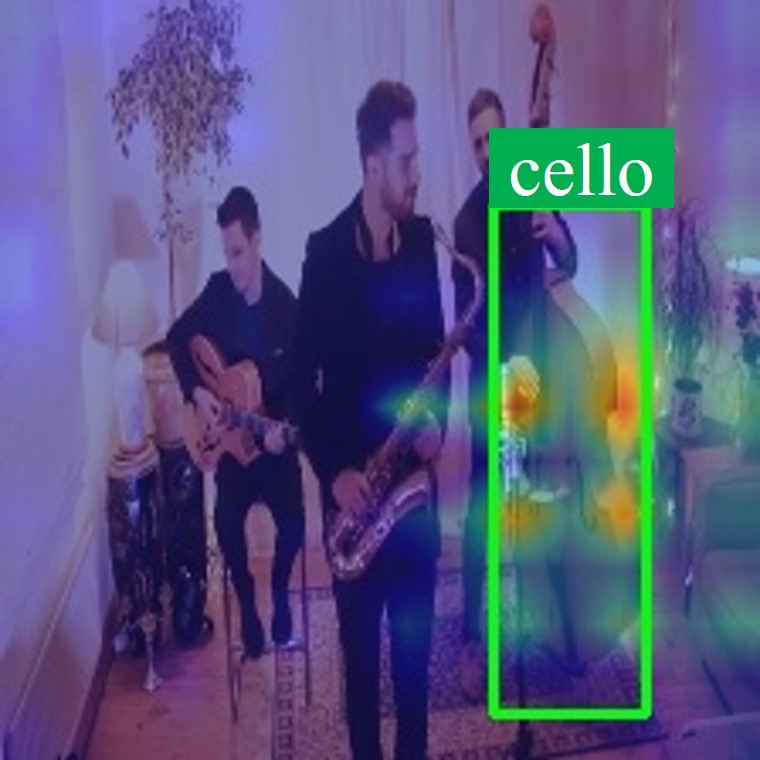}
    \includegraphics[width=0.135\linewidth]{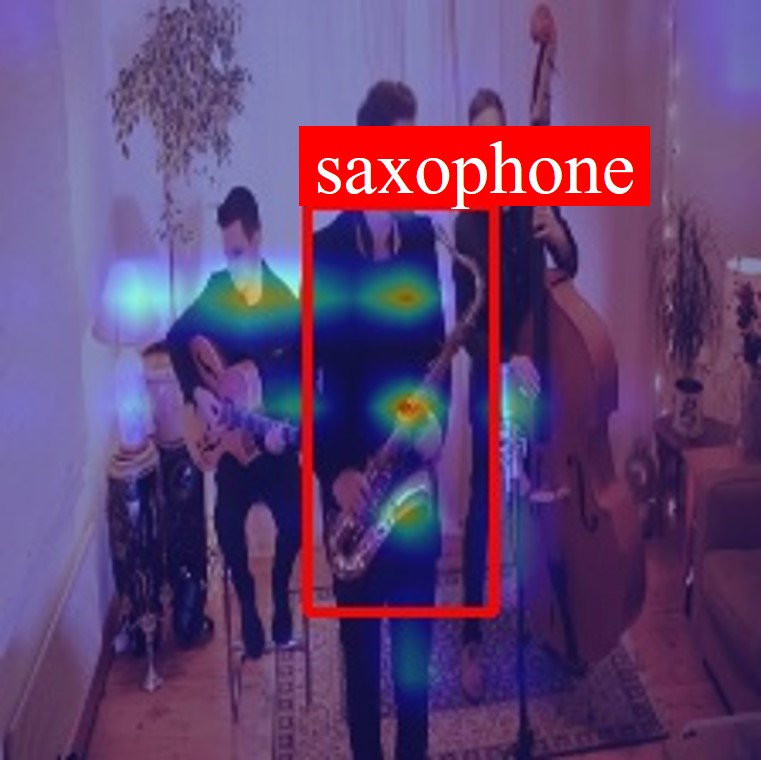}\hspace{1mm}
    \includegraphics[width=0.135\linewidth]{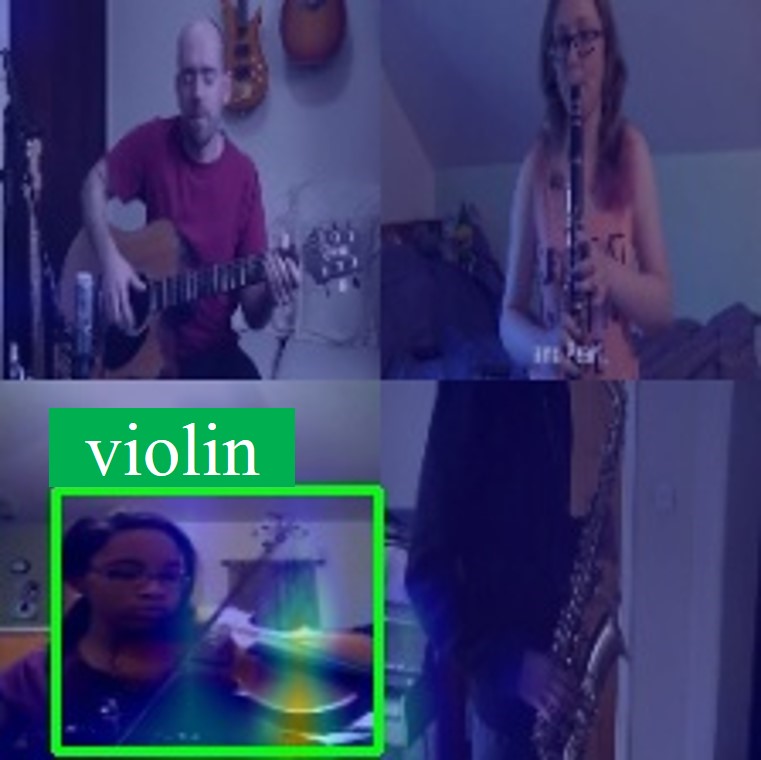}
    \includegraphics[width=0.135\linewidth]{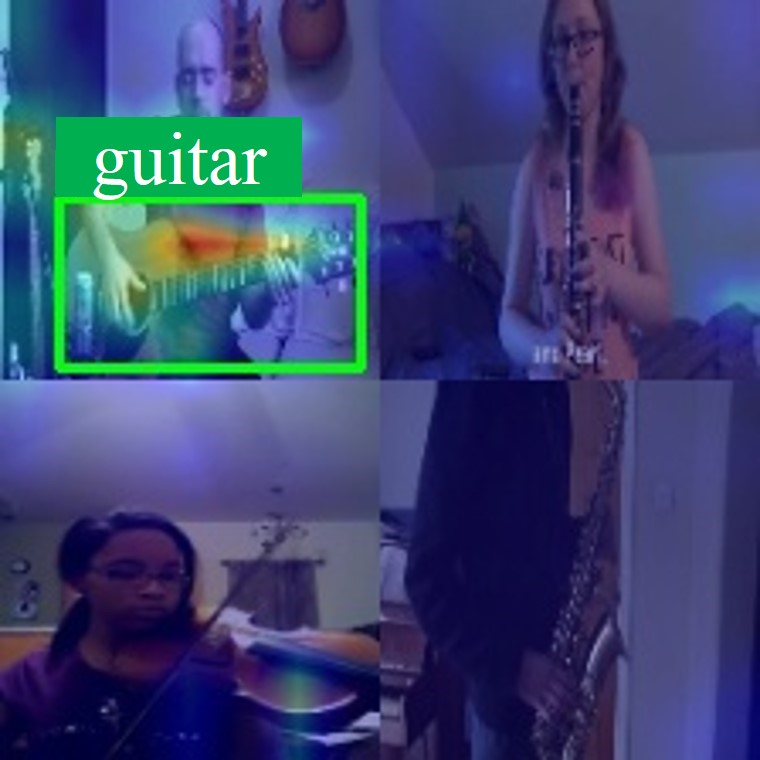}
    \includegraphics[width=0.135\linewidth]{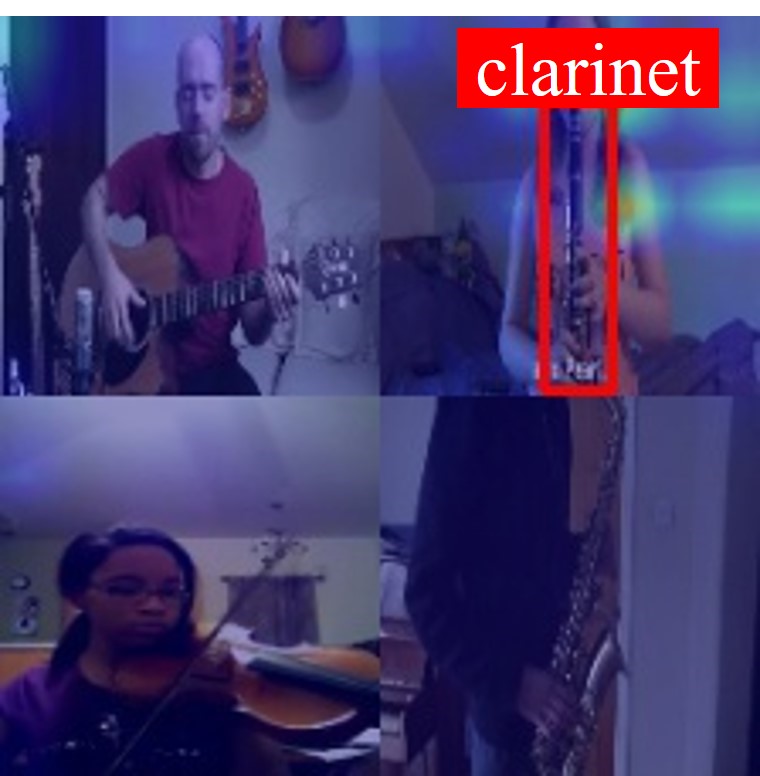}
    \includegraphics[width=0.135\linewidth]{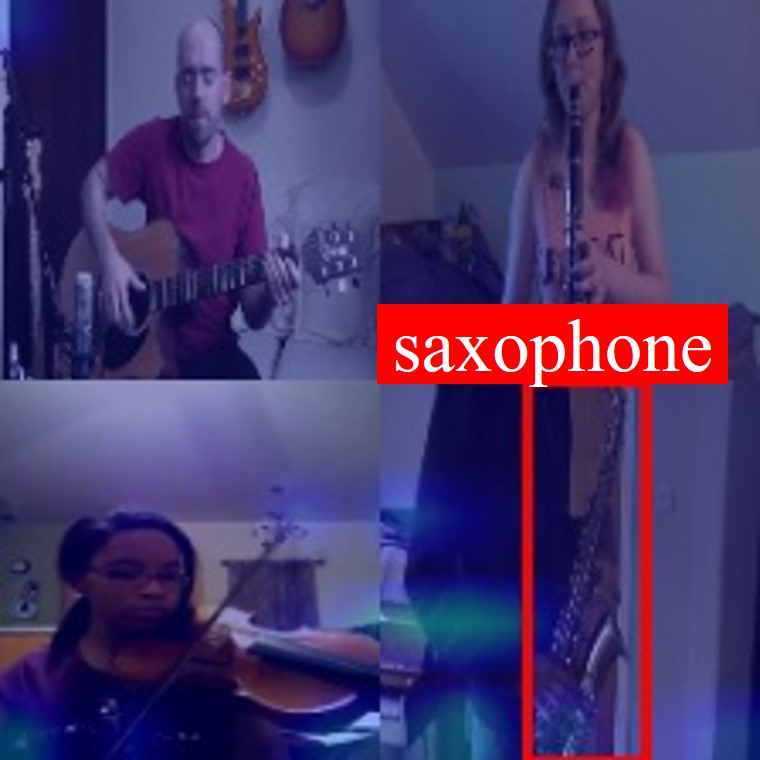}
    }
    \subfigure[Ours]{
    \includegraphics[width=0.135\linewidth]{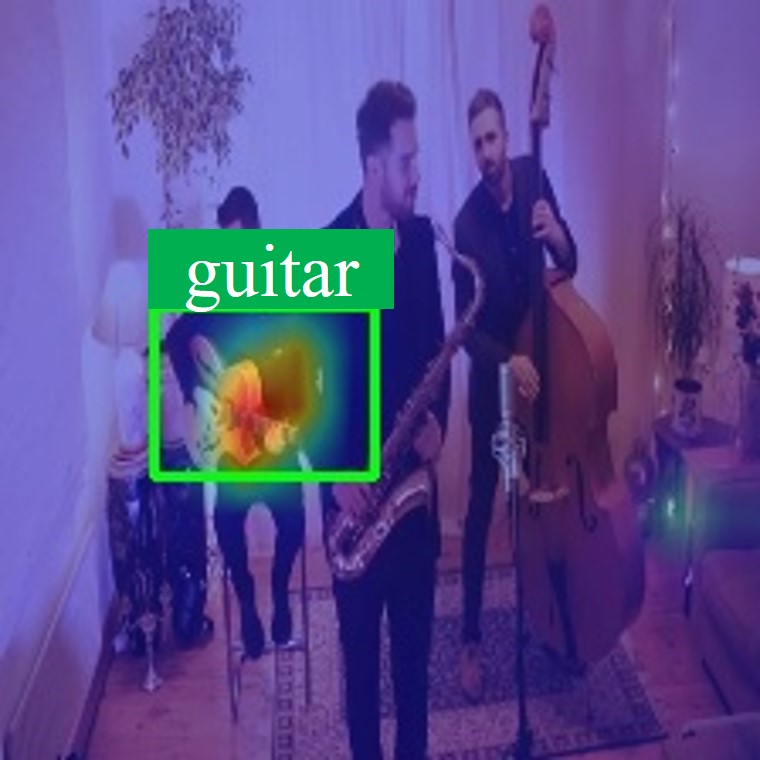}
    \includegraphics[width=0.135\linewidth]{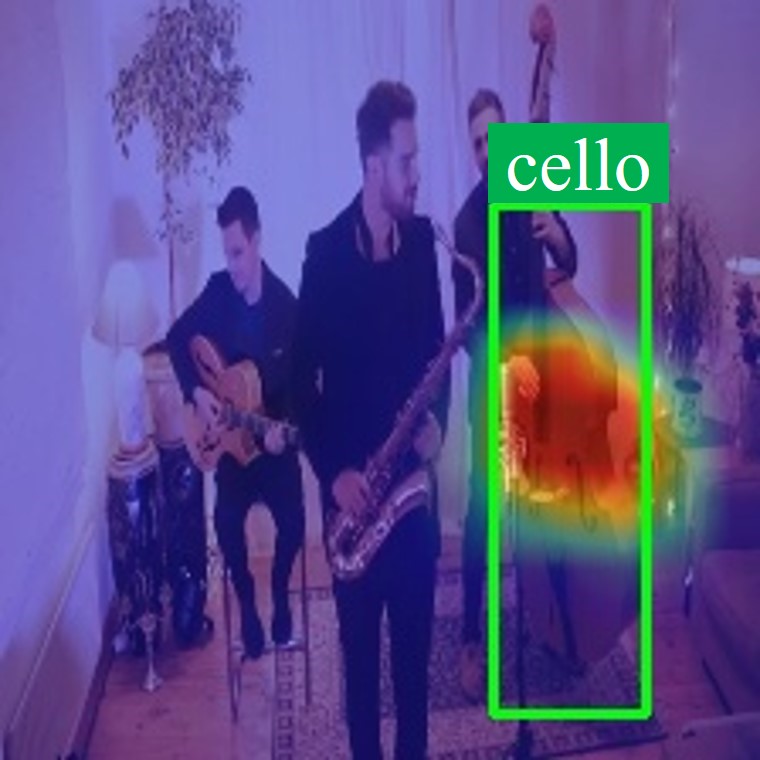}
    \includegraphics[width=0.135\linewidth]{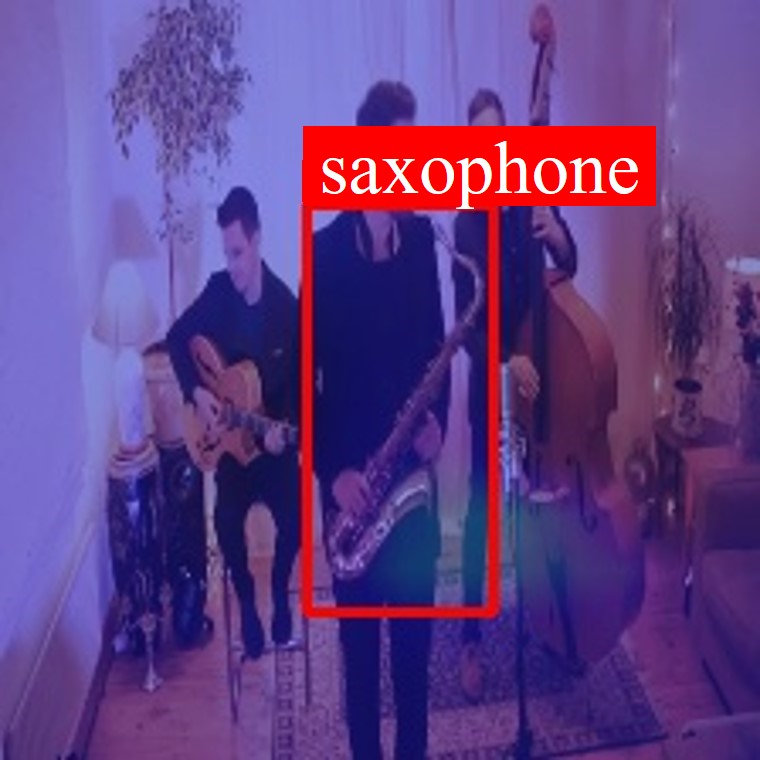}\hspace{1mm}
    \includegraphics[width=0.135\linewidth]{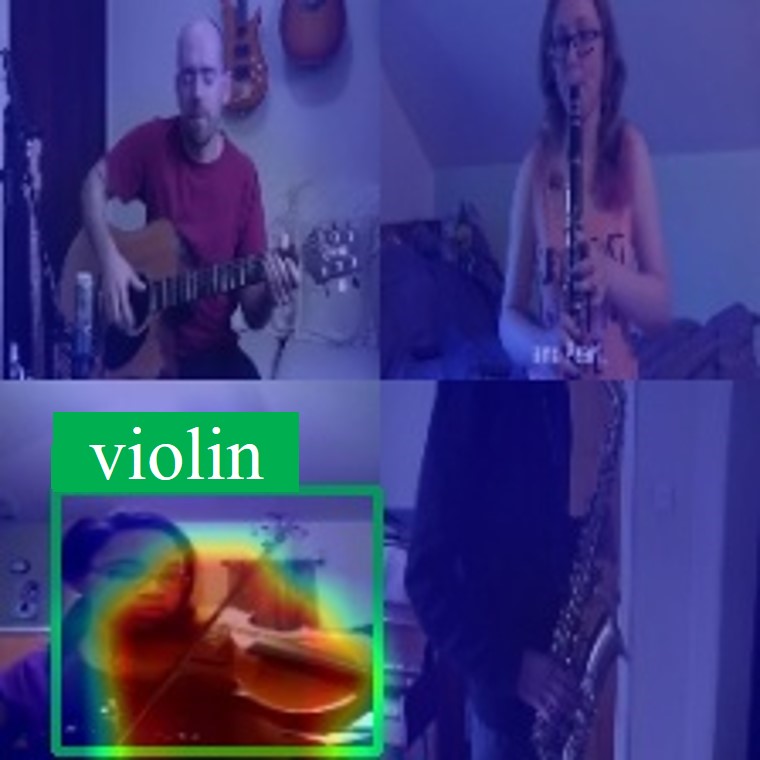}
    \includegraphics[width=0.135\linewidth]{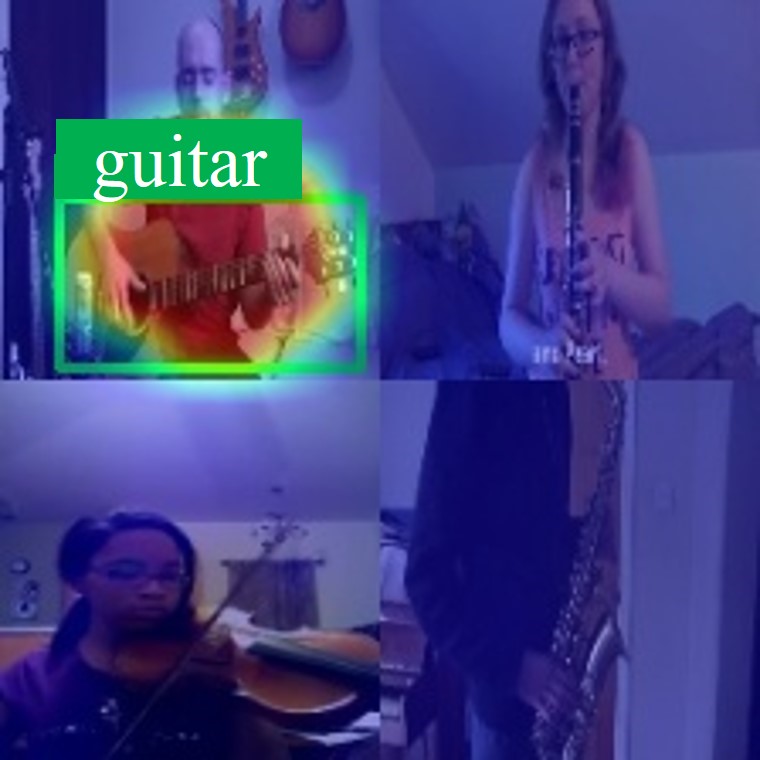}
    \includegraphics[width=0.135\linewidth]{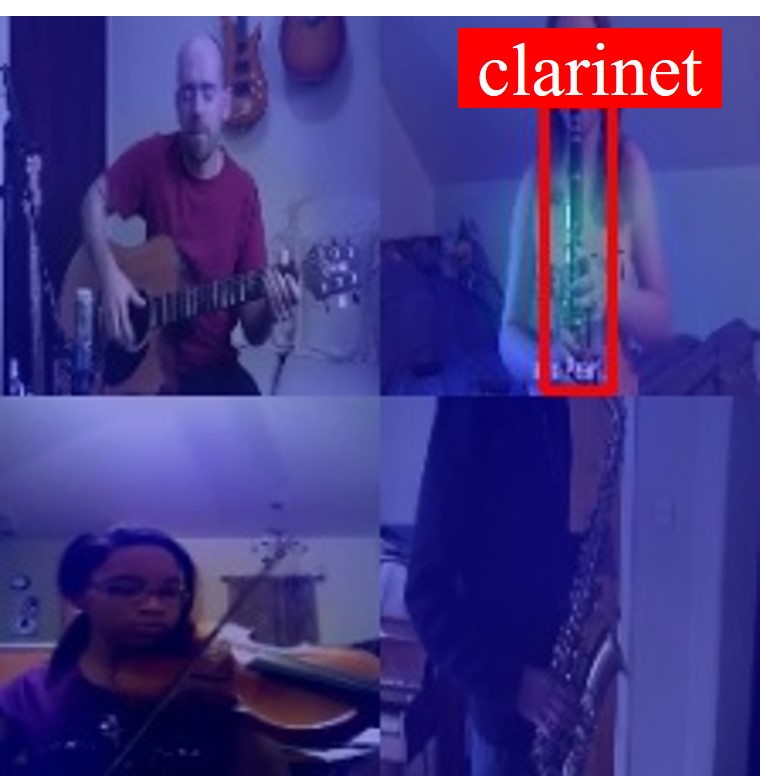}
    \includegraphics[width=0.135\linewidth]{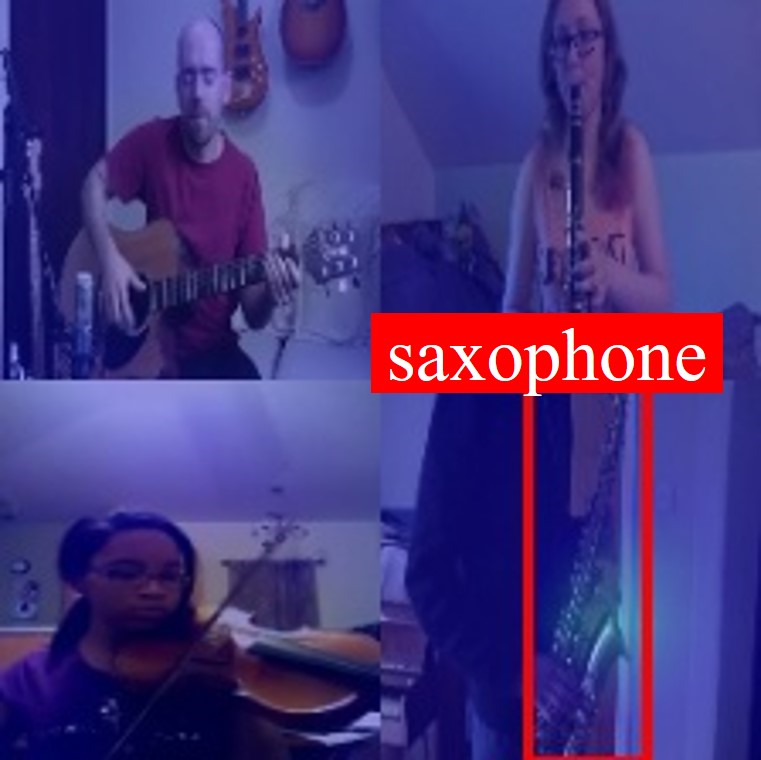}
    }
    \caption{\textbf{Sounding object localization results in music scenes.} We visualize some localization results of different methods on realistic and synthetic cocktail-party videos. The class-aware localization maps are expected to localize objects of different classes and filter out silent ones. The first 3 pictures and the last 4 pictures in each row respectively depict the location of instruments at the same moment in a realistic video and a synthetic video. The green box indicates target sounding object area, and the red box means this class of object is silent, and its activation value should be low.}
    \label{fig:sample}
\end{figure*}

\subsection{Multiple Sounding Objects Localization}

The realistic audio-visual scenario is usually composed of multiple sounding and silent objects, bringing great challenge to discriminative sounding objects localization. 
We evaluated different methods on both synthetic and real data to validate models more reliably. Table \ref{tbl:syn} illustrated that our model is significantly superior to all other comparative methods in CIoU. Three main reasons can explain this phenomenon.

First, the class information of sounding objects is taken into account through utilizing a category-based audiovisual alignment in our method, i.e., Eq.~\ref{eq:stage2_kl}. However, other methods~\cite{arandjelovic2017objects,senocak2018learning} just simply correlate the audiovisual features of the sounding area and cannot handle discriminatively localizing the sounding objects task. 

Second, our method benefits from the effective category-representation object dictionary learned in the first stage. Such an object dictionary can help our model locate potential objects in complex cocktail scenarios. The compared method, Sound-of-pixel~\cite{zhao2018sop} cannot handle this scenario with a mix-then-separate learning strategy. 

Third, the silent objects can be filtered by our model, while DMC~\cite{hu2019deep} has to depend on the known number of sounding objects. In addition, the high-value of~\cite{zhao2018sop} in NSA may be due to the channel activation being too low to detect objects, rather than successfully filtering out silent objects.

We also show visualized localization results in realistic scenarios in Fig.~\ref{fig:sample}, in addition to quantitative evaluation. 
Based on the shown results, the attention-based approach~\cite{senocak2018learning} and Object-the-sound~\cite{arandjelovic2017objects} can just simply localize the sounding area and is unable to discriminate the guitar or cello. DMC~\cite{hu2019deep} fails in such scenarios, mixing up different visual areas. Although sound-of-pixel~\cite{zhao2018sop} provides comparable results, it is not able to localize the sounding object exactly and successfully filter out the silent saxophone, possibly because it relies highly on the quality of the mixed sounds. In contrast to these methods, our model localizes sounding guitar and cello and shows the low response for the silent saxophone and sound-unrelated area. The synthetic data yield similar results. 

In addition, to further evaluate both localization and recognition abilities of our model, we introduce the sounding mAP metric, which is similar to the mAP metric in object detection~\cite{everingham2010pascal}, while only taking the bounding box and category label of sounding objects as the ground truth. Since other methods cannot discriminatively localize sounding objects, we compare our model with one baseline method:
\begin{itemize}
  \item [1)] \textbf{Ours} generates the bounding boxes for each category based on the localization results. The heatmap of each category is normalized, and the region larger than the threshold (0.5 in our experiment) is selected. Then, we extract boxes that can cover these regions with minimum size as the final bounding boxes. 
  \item [2)] \textbf{Random} is simple baseline. The bounding boxes are the same as \textbf{Ours} while their category is randomly assigned.
\end{itemize}

We evaluate these methods on the MUSIC-Duet and MUSIC-Synthetic, and the evaluation metric is $\text{Sounding mAP}_{30}$. For the MUSIC-Duet dataset, \textbf{Random} obtains 1.8 and \textbf{Ours} method achieves 11.5. In addition, \textbf{Random} and \textbf{Ours} respectively achieves 6.3 and 31.4 on the MUSIC-Synthetic dataset. Our method obtains reasonable performance, especially on the MUSIC-Synthetic dataset with more balanced instrument category distribution.

In the experiment, we also find that some sound sources are out-of-screen in our experiment. For example, the instrument disappears from the screen sometimes. These out-of-screen clips were not excluded during the training. Based on the results on different dataset, this phenomenon does not have a huge impact on the model performance. However, for the sounding object localization results of such an out-of-screen scenario, the model would fail. Fig.~\ref{fig:ablation-outscreen} shows a failure case of out-of-screen sounding objects.

\label{general}

\subsection{From Music to General Cases}

Besides the aforementioned music scenes, we expand our method to more general daily cases covering various sounding objects, e.g., humans, animals, vehicles. In this subsection, we analyze the single and multiple sounding object localization performance in general scenes on VGGSound dataset~\cite{vedaldi2020vggsound} as well as a series of complex scene videos collected from YouTube. Table~\ref{tab:vggsound} shows the single sound localization results on VGGSound test videos. Note that there are much more sound categories in VGGSound than in music scenes (98 vs 21), Object-that-sound~\cite{arandjelovic2017objects} based on simple audiovisual correspondence performs well among compared methods. Sound-of-pixel~\cite{zhao2018sop} has difficulty establishing channel-wise correlations, and DMC~\cite{hu2019deep} also suffer from dealing with too many concrete audiovisual components. However, our model is robust to this change and still outperforms others, proving that our alternative localization-classification learning scheme can be adapted to general scenarios. And class-aware multiple sounding object localization performance is shown in Table~\ref{tab:vggsyn}, since the large number of categories makes this task quite challenging, the performance in terms of CIoU and AUC is lower than music scenes. According to the results, our model is superior in all three metrics, but the improvement is not that significant as in music scenarios.

\begin{table}[]
    \centering
    \caption{Single sounding object localization results on VGGSound.}
    \begin{tabular}{l|cc}
    \hline
        Methods & IoU@0.5 & AUC \\
        \hline
        Sound-of-pixel~\cite{zhao2018sop} & 42.5 & 45.1 \\
        Object-that-sound~\cite{arandjelovic2017objects} & 48.4 & 46.1 \\
        Attention~\cite{senocak2018learning} & 39.3 & 40.5 \\
        DMC~\cite{hu2019deep} & 40.4 & 41.3 \\
        Ours & \textbf{49.5} & \textbf{46.7} \\
        \hline
    \end{tabular}
    \label{tab:vggsound}
\end{table}

\begin{table}[]
    \centering
    \caption{Multiple sounding object localization results on VGGSound-Synthetic and realistic DailyLife.}
    \subtable[Results on VGGSound-Synthetic.]{
    \begin{tabular}{l|ccc}
    \hline
        Methods & CIoU@0.3 & AUC & NSA \\
        \hline
        Sound-of-pixel~\cite{zhao2018sop} & 6.7 & 12.4 & 93.4 \\
        Object-that-sound~\cite{arandjelovic2017objects} & 8.1 & 14.3 & 27.3 \\
        Attention~\cite{senocak2018learning} & 7.3 & 13.2 & 64.5 \\
        DMC~\cite{hu2019deep} & 7.5 & 14.1 & - \\
        Ours & \textbf{13.4} & \textbf{17.9} & \textbf{94.1} \\
        \hline
    \end{tabular}
    \label{tab:vggsyn}}

  \qquad 
\subtable[Results on realistic DailyLife.]{
    \begin{tabular}{l|ccc}
    \hline
        Methods & CIoU@0.3 & AUC & NSA \\
        \hline
        Sound-of-pixel~\cite{zhao2018sop} & 12.2 & 15.5 & \textbf{91.4} \\
        Object-that-sound~\cite{arandjelovic2017objects} & 15.4 & 17.9 & 19.4 \\
        Attention~\cite{senocak2018learning} & 14.1 & 18.4 & 43.7 \\
        DMC~\cite{hu2019deep} & 13.6 & 17.1 & - \\
        Ours & \textbf{22.6} & \textbf{24.1} & 87.5 \\
        \hline
        Oracle-a & 21.3 & 23.6 & 88.7\\
        Oracle-v & 44.0 & 31.0 & 95.0 \\
        \hline
    \end{tabular}
    \label{tab:vggreal}}
\end{table}

\begin{table}[!]
    \centering
    \caption{Comparison between ours and oracle settings.}
    \begin{tabular}{l|cc|ccc}
    \hline
        Methods & NMI & mAP & CIoU@0.3 & AUC & NSA \\
        \hline
        Ours & 0.463 & 0.232 & 13.4 & 17.9 & 94.1 \\
        Oracle-a & 0.470 & 0.289 & 13.7 & 17.4 & 90.2 \\
        Oracle-v & \textbf{0.746} & \textbf{0.640} & \textbf{21.4} & \textbf{19.6} & \textbf{95.4} \\
        \hline
    \end{tabular}
    \label{tab:compare}
\end{table}

To delve into this phenomenon, we further analyze the ability of our model in visual object feature discrimination and audio event classification in detail. Table~\ref{tab:compare} shows the comparison between our method with two oracle versions, where Ours adopts pseudo categories from clustering to supervise classification. The two oracles respectively use annotated audio labels (Oracle-a) or object categories (Oracle-v) as the supervision. Note that we manually aggregate audio labels in VGGSound subset into the categories in COCO dataset~\cite{lin2014microsoft} to obtain corresponding object labels. Results show that replacing pseudo labels with ground-truth audio categories improves NMI and mAP, and using object labels as classification targets significantly facilitates performance. For the discriminative sounding object localization performance, Ours is comparable with Oracle-a but much lower than Oracle-v. The underlying reasons are two-fold: First, in general scenes, the audio and visual object categories are not aligned: the same object can produce very different sounds, e.g., humans can make sound of speech, clapping hands, etc. In other words, one visual object category could correspond to multiple audio classes. And in our experimental settings, we need to aggregate different various sounds to the specific objects. Therefore, the annotated audio labels are too fine-grained and not suitable for class-aware sounding object localization. Second, since there lacks one-to-one correspondence between audio and object, it tends to correlate audio with background scenes rather than the sounding object. For example, it is straightforward for the model to align singing with concert scene, and hiccup with eating scene. In this way, our audiovisual localization map would favour some discriminate components in the background visual scene instead of sounding objects, which corrupts class-aware sounding objects localization. 

\begin{figure}
\centering
    \includegraphics[width=0.22\linewidth]{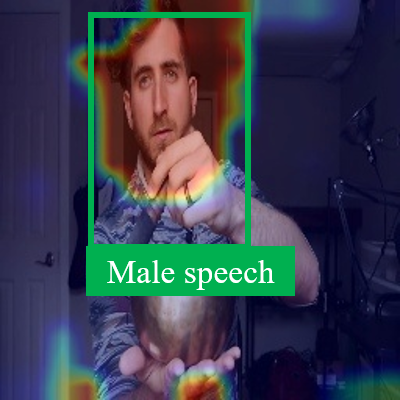}
    \includegraphics[width=0.22\linewidth]{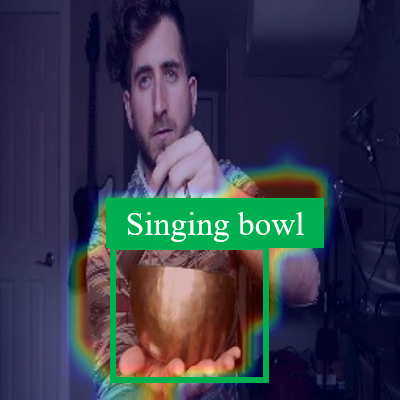}\hspace{1mm}
    \includegraphics[width=0.22\linewidth]{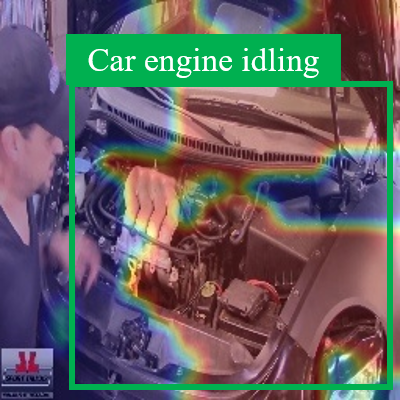}
    \includegraphics[width=0.22\linewidth]{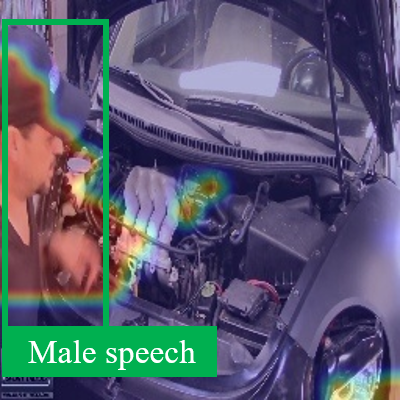}\\
    
    \subfigure[DailyLife dataset]{
    \includegraphics[width=0.22\linewidth]{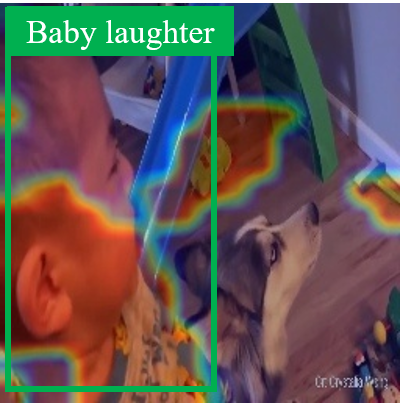}
    \includegraphics[width=0.22\linewidth]{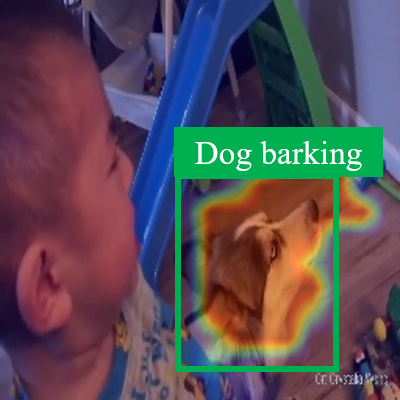}\hspace{1mm}
    \includegraphics[width=0.22\linewidth]{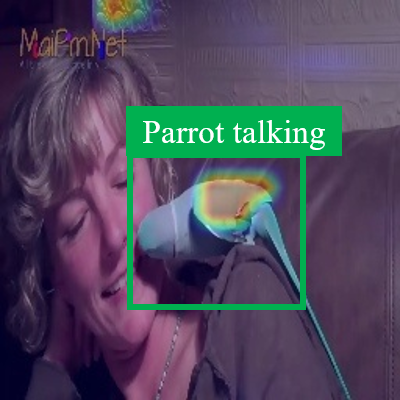}
    \includegraphics[width=0.22\linewidth]{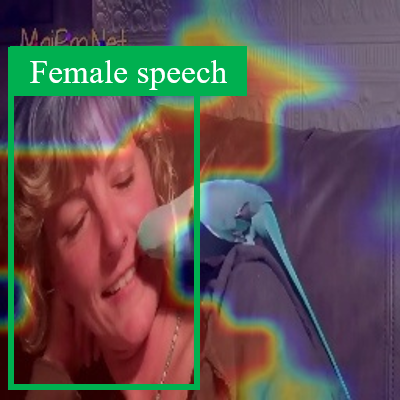}}\\
    
    \subfigure[Realistic MUSIC dataset]{
    \includegraphics[width=0.22\linewidth]{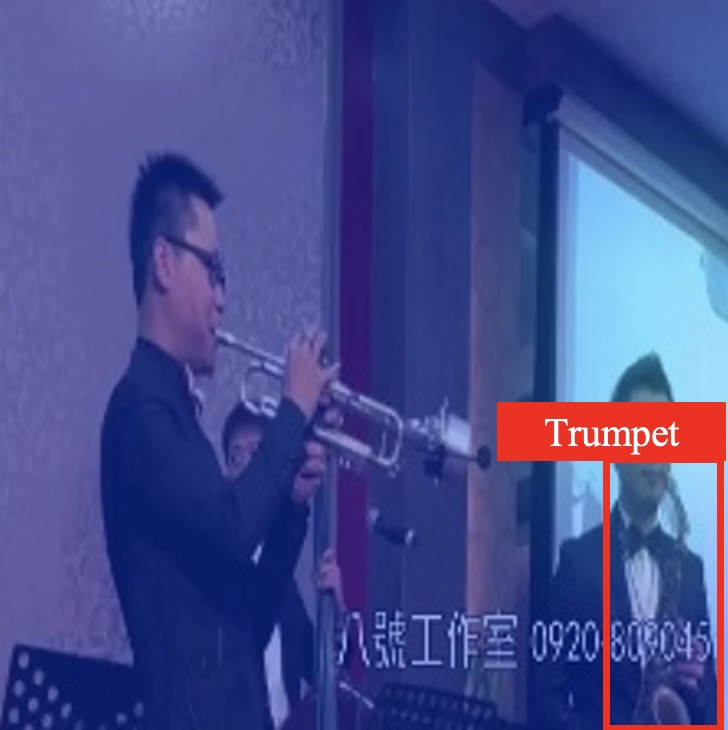}
    \includegraphics[width=0.22\linewidth]{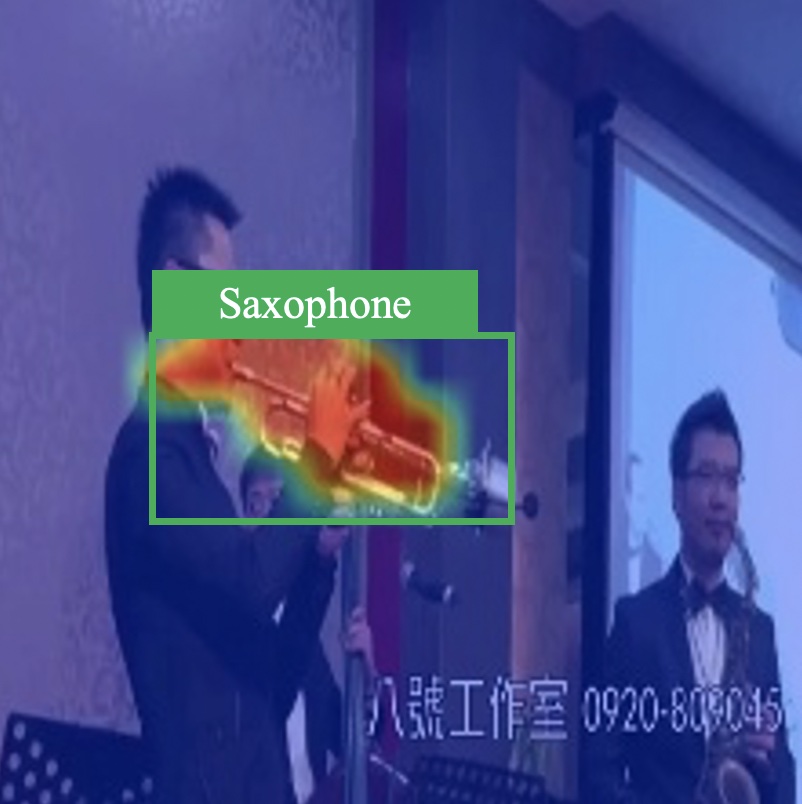}\hspace{1mm}
    \includegraphics[width=0.22\linewidth]{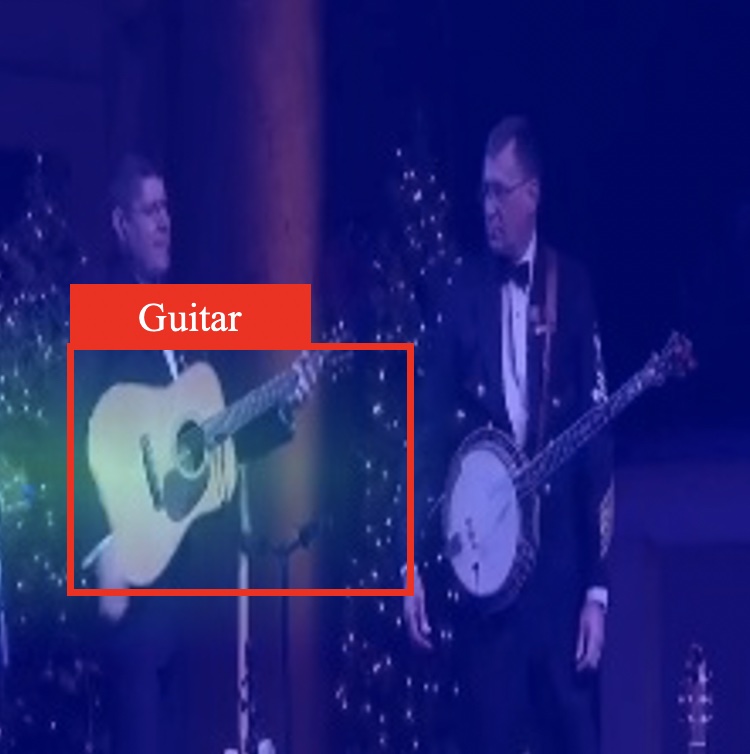}
    \includegraphics[width=0.22\linewidth]{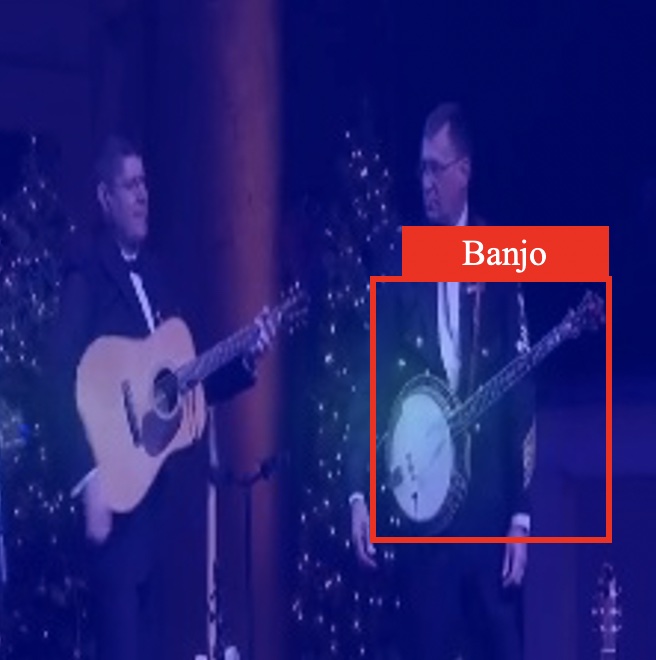}}\\
    \subfigure[VGGSound dataset]{
    \includegraphics[width=0.3\linewidth]{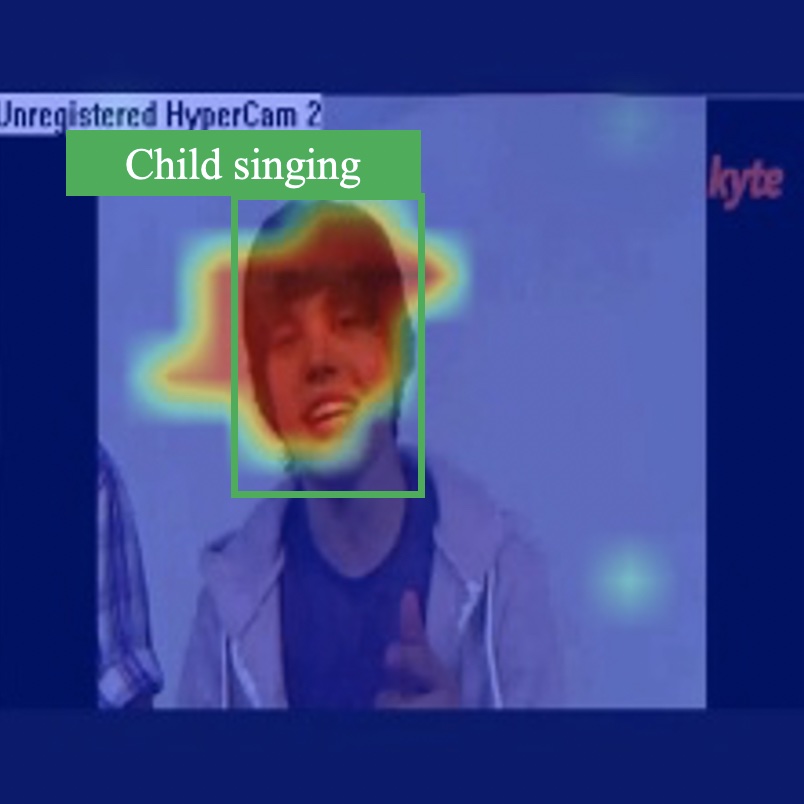}
    \includegraphics[width=0.3\linewidth]{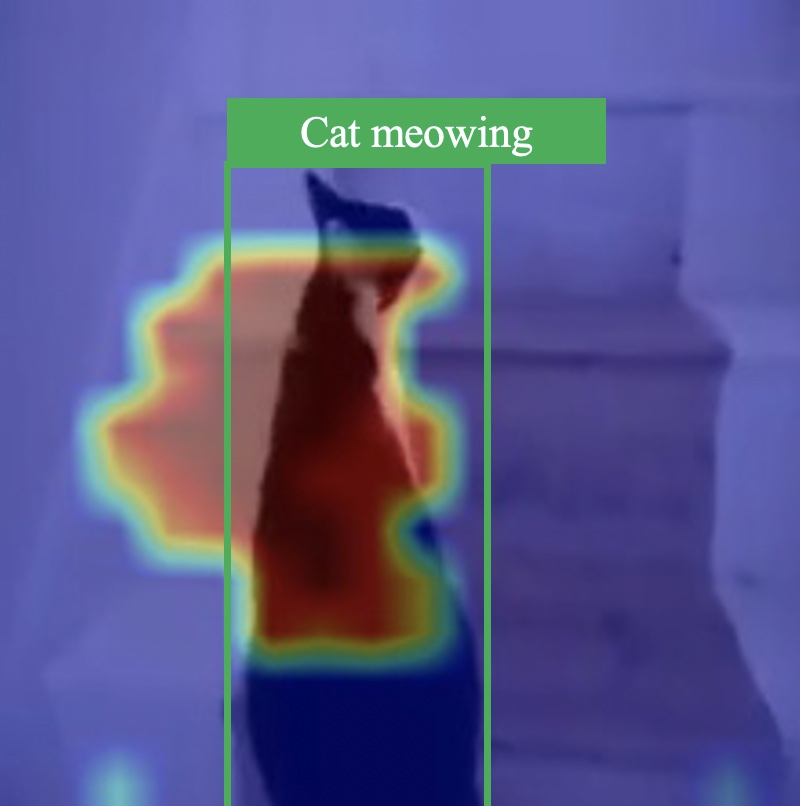}
    \includegraphics[width=0.3\linewidth]{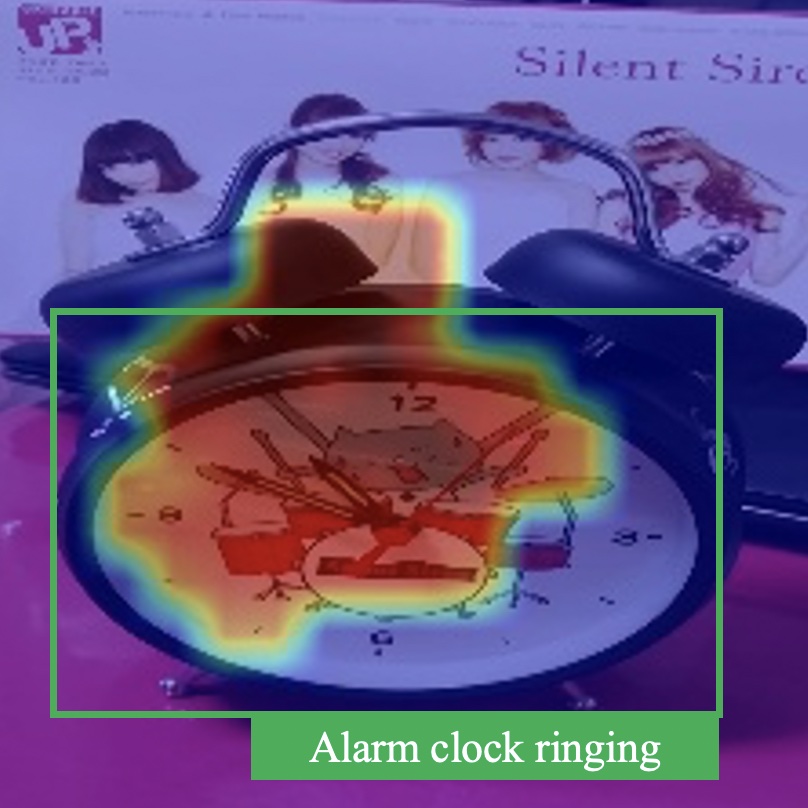}}
\caption{\textbf{Sounding object localization results on DailyLife, 
Realistic MUSIC and VGGSound dataset.} We visualize our localization results in realistic daily life and music scenarios. The green box indicates target sounding object area. Our model can distinguish different categories and respectively assign them to corresponding objects.}
\label{fig:vggreal}
\end{figure}

\subsection{Evaluation on Realistic Data}

In order to further verify the generalization ability of our model, we test it on the Realistic MUSIC \& DailyLife dataset. Table~\ref{tbl:syn} shows the class-aware sounding object localization performance on Realistic MUSIC dataset. The overall performance of models is not good enough compared to a clean dataset, e.g., MUSIC dataset, since the realistic music scenes are very noisy. However, our model still outperforms recent SOTA methods~\cite{senocak2018learning,arandjelovic2017objects,zhao2018sop,hu2019deep}. 

\begin{table*}
    \centering
        \caption{Ablation study for the second stage on the VGGSound-Synthetic, Realistic DailyLife, MUSIC-Synthetic and MUSIC-Duet dataset.}
        \begin{tabular}{ccc|ccc|ccc|ccc|ccc}
            \hline
            \multicolumn{3}{c|}{Dataset} & \multicolumn{3}{c|}{VGGSound-Synthetic} & \multicolumn{3}{c|}{DailyLife} & \multicolumn{3}{c|}{MUSIC-Synthetic} & \multicolumn{3}{c}{MUSIC-Duet} \\
            \hline
            $\mathcal{L}_{loc}$ & Prod & $\mathcal{L}_c$ & CIoU@0.3 & AUC & NSA & CIoU@0.3 & AUC & NSA & CIoU@0.3 & AUC & NSA & CIoU@0.3 & AUC & NSA\\
            \hline
            \XSolidBrush & \XSolidBrush & \CheckmarkBold & 0.0 & 6.6 & 92.3 & 7.9 & 12.3 & 83.1 & 0.0 & 7.2 & 91.0 & 20.8 & 15.9 & 78.0\\
            \CheckmarkBold & \XSolidBrush & \CheckmarkBold & 3.1 & 9.7 & 83.5 & 8.1 & 12.2 & 77.3 & 2.6 & 7.5 & 88.1 & 20.6 & 20.2 & 79.6\\
            \CheckmarkBold & \CheckmarkBold & \XSolidBrush & 11.5 & 15.4 & 90.3 & 20.1 & 20.4 & 82.2 & 18.0 & 17.4 & 92.9 & 22.4 & 19.2 & 85.0\\
            \CheckmarkBold & \CheckmarkBold & \CheckmarkBold & {13.4} & {17.9} & {94.1} & {22.6} & {24.1} & {87.5} & 32.3 & 23.5 & 98.5 & 30.2 & 22.1 & 83.1\\
            \hline
        \end{tabular}
        \label{2nd-ablation-vgg}   
\end{table*}

\begin{figure}
    \centering
    \includegraphics[width=0.22\linewidth]{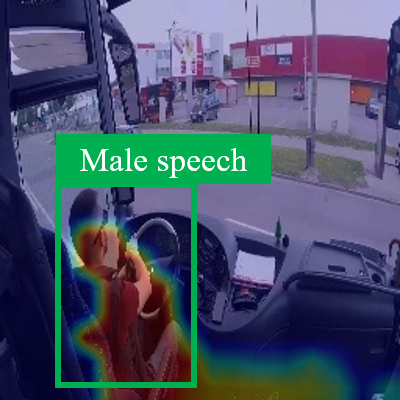}
    \includegraphics[width=0.22\linewidth]{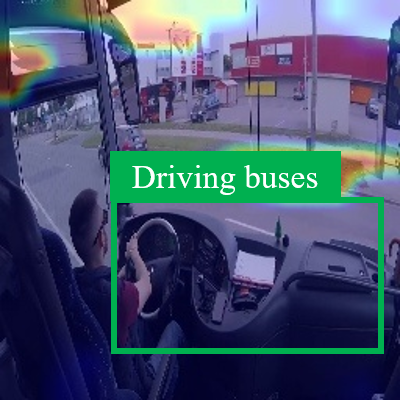}\hspace{1mm}
    \includegraphics[width=0.22\linewidth]{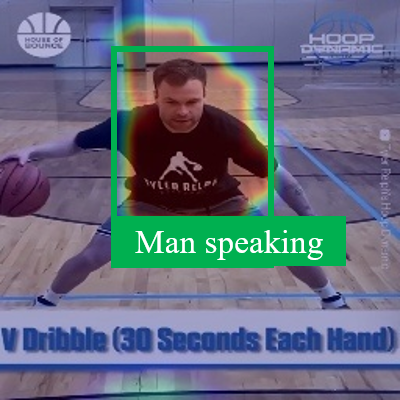}
    \includegraphics[width=0.22\linewidth]{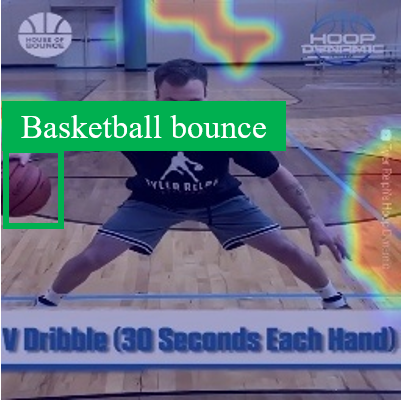}\\
    \vspace{2mm}
    \includegraphics[width=0.22\linewidth]{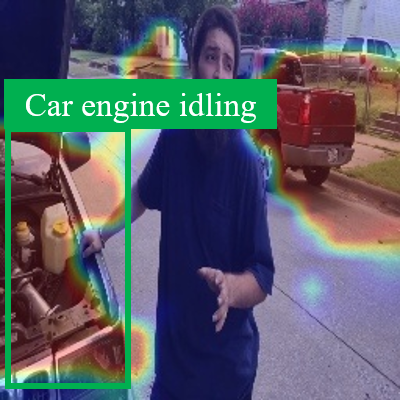}
    \includegraphics[width=0.22\linewidth]{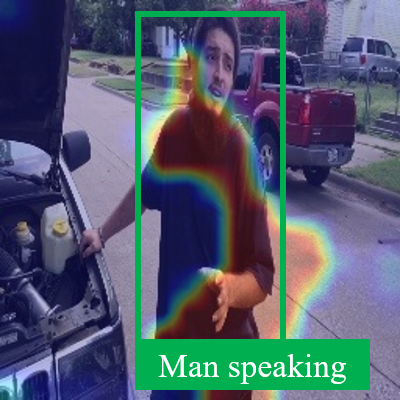}\hspace{1mm}
    \includegraphics[width=0.22\linewidth]{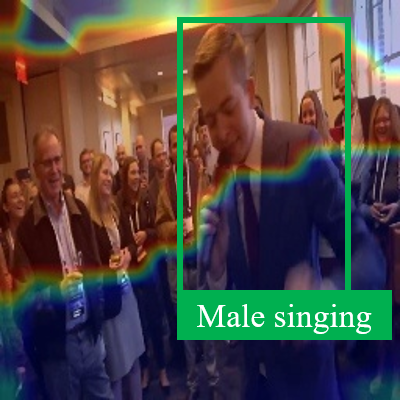}
    \includegraphics[width=0.22\linewidth]{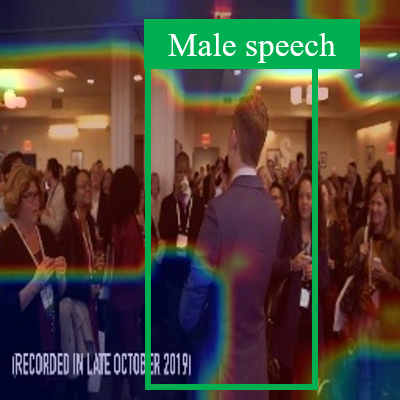}
    \caption{\textbf{Failure cases on DailyLife dataset.} We visualize some failure cases in realistic daily life scenarios. The first row shows that our method fails to localize sounding object but focus on some representative background areas. The second row indicates that our model cannot identify the specific sounding instance in some complex scenes.}
    \label{fig:vggfail}
\end{figure}

Table~\ref{tab:vggreal} shows the class-aware sounding object localization performance on Realistic DailyLife dataset. Overall, the performance of different methods is consistent with that on VGGSound-Synthetic dataset, which demonstrates that the model trained on synthesized cocktail-party videos can generalize to realistic complex scenes. Our method trained with audiovisual consistency outperforms recent SOTA methods~\cite{senocak2018learning,arandjelovic2017objects,zhao2018sop,hu2019deep} as well as the supervised Oracle-a, which indicates the efficacy and generalization ability of the audiovisual correspondence supervision. Though there remains a gap between Ours with Oracle-v supervised by object labels, it would be promising to leverage recent progress in self-supervised representation learning to bridge this gap. Besides the quantitative results, we also visualize some realistic cases in Fig.~\ref{fig:vggreal}.
Like music scenes, our model can distinguish different audio categories in mixed sounds and assign them to corresponding visual objects, e.g., we can correlate laughter sound with the baby face and assign barking to the dog. However, there are also some failure cases in two kinds of scenes as shown in Fig.~\ref{fig:vggfail}. First, for audio categories whose sound source usually exists in typical background scenes, e.g., driving buses with street scene, basketball bounce with basketball court, the model tends to figure out discriminative parts in the background scene as sounding objects. Second, in the scene containing multiple instances of the same category, since we only input single static image with mixed sound, our method fails to determine specific sounding instance, e.g., distinguishing starting car and silent one, finding speaker or singer among a crowd of people. We will delve into these problems in future studies.

\subsection{Ablation Study}
In this section, we perform ablation studies on the hyper-parameters and learning schemes of our method.

\textbf{Number of clusters.} In default experiment settings on VGGSound dataset, we do not consider the audio or object category number in the subset and directly set the number of clusters to 100. In the ablation study, we explore different numbers, e.g., 100, 200, 5000 and also try using the number of audio categories 98 for the first-stage feature clustering. In evaluation, like music scenes, we adaptively aggregate multiple clusters to one specific category for class-aware localization. Table~\ref{cluster-ablation-vgg} shows the results on VGGSound, VGGsound-Synthetic, and DailyLife dataset. The results show that the performance of single source scenarios is not hugely affected by the number of clusters since the first stage does not need to infer based on the clusters, and setting the clustering number to a comparatively large number does not harm the multi-source localization performance much. However, when the number of clusters is much larger than the number of categories, e.g., 2000 and 5000, the model performance has a noticeable drop since it is a considerable burden to generate a category-representation object dictionary with high-quality under such a situation, which need further exploration in the future work.

\begin{figure}
    \centering
    \includegraphics[width=0.49\linewidth]{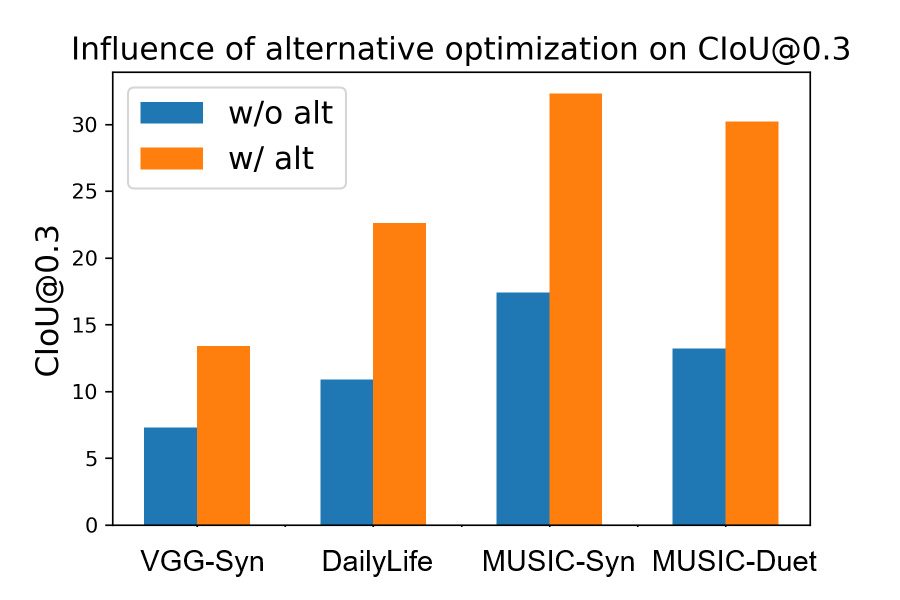}
    \includegraphics[width=0.49\linewidth]{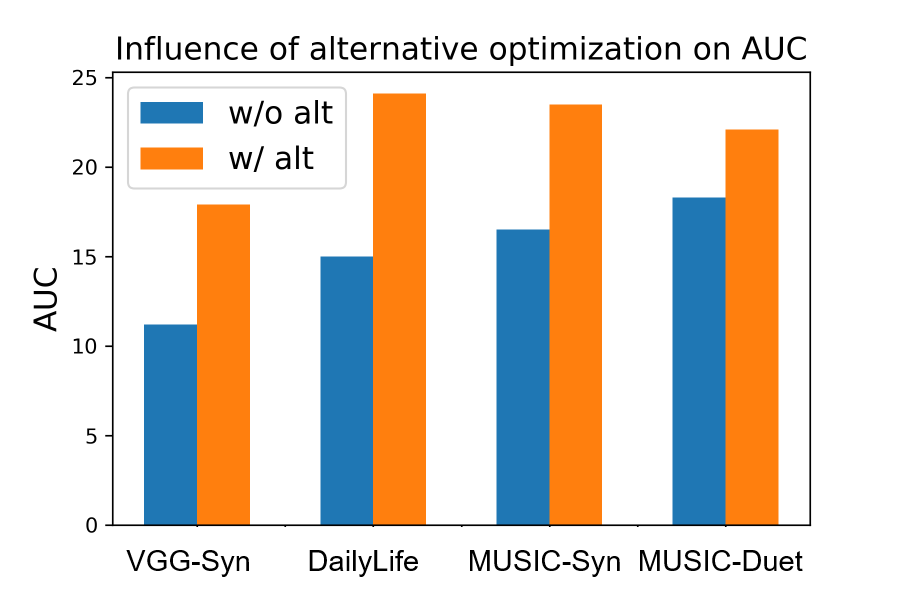}
    \caption{Influence of alternative localization-classification learning on the VGGSound-Synthetic, Realistic DailyLife, MUSIC-Synthetic and MUSIC-Duet dataset.}
    \label{fig:alt-ablation-vgg}
\end{figure}

\begin{figure}
    \centering
    \includegraphics[width=0.49\linewidth]{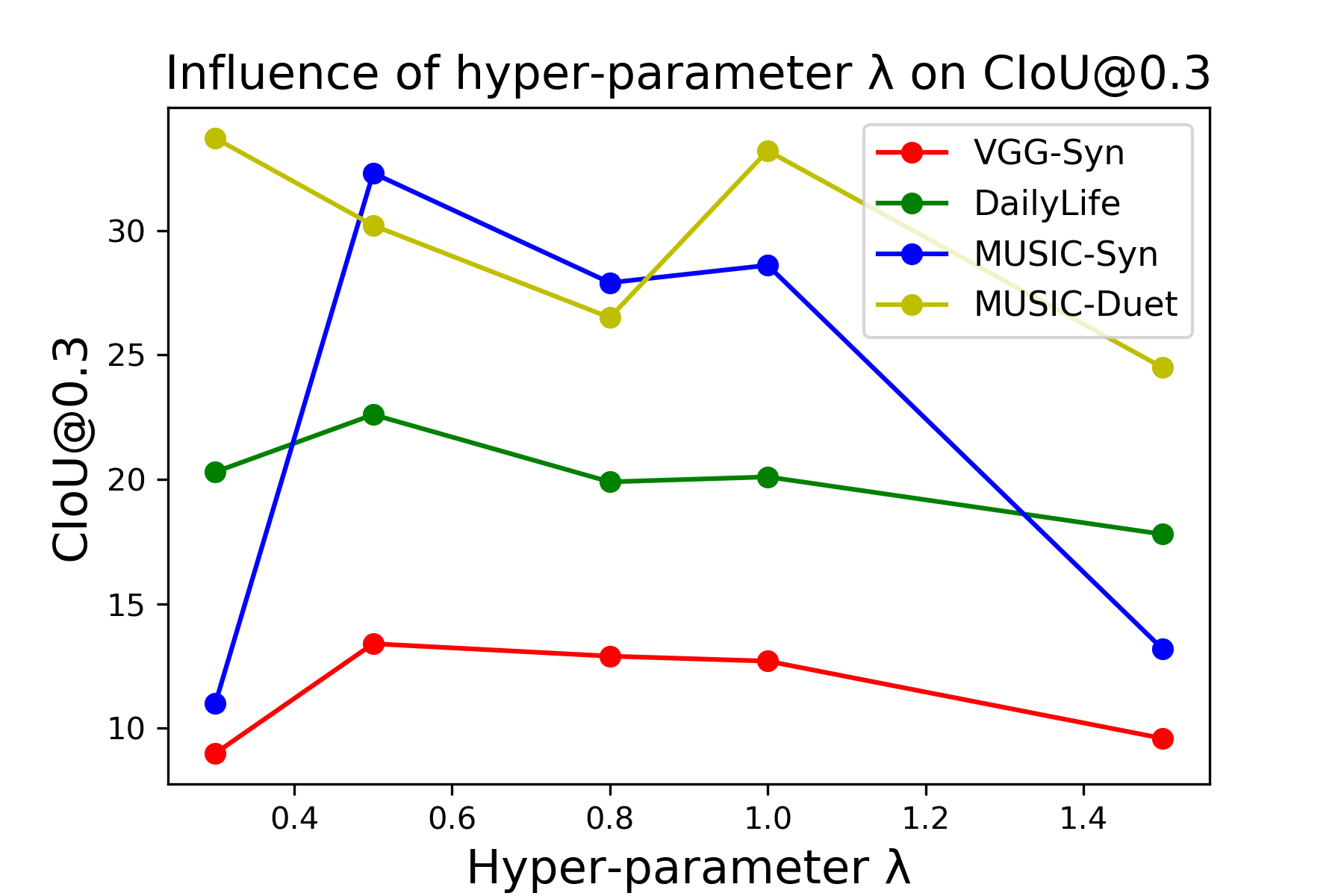}
    \includegraphics[width=0.49\linewidth]{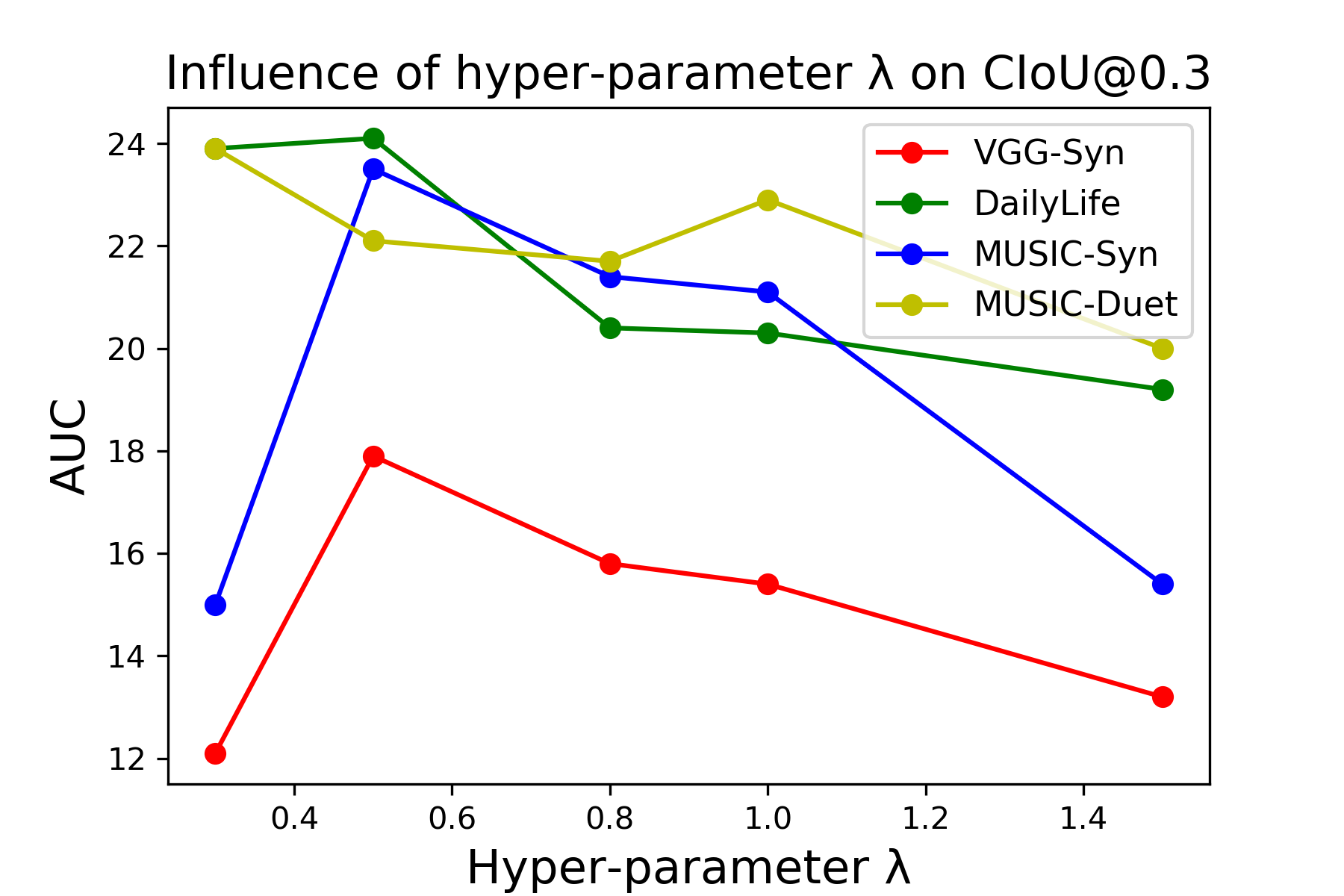}
    \caption{Influence of the hyper-parameter $\lambda$ on the VGGSound-Synthetic, Realistic DailyLife, MUSIC-Synthetic and MUSIC-Duet dataset.}
    \label{fig:loss-ablation-vgg}
\end{figure}

\begin{table*}
    \centering
    \caption{Ablation study on the number of clusters in general cases.}
    \begin{tabular}{c|ccc|ccc|ccc}
        \hline
        \multicolumn{1}{c|}{Dataset} & \multicolumn{3}{c|}{VGGSound} & \multicolumn{3}{c|}{VGGSound-Synthetic} & \multicolumn{3}{c}{DailyLife} \\ 
        \hline
        Cluster & NMI & IoU@0.5 & AUC & CIoU@0.3 & AUC & NSA & CIoU@0.3 & AUC & NSA\\
        \hline
        98 & 0.465 & 50.1 & 47.3 & 13.2 & 17.1 & 95.3 & 21.2 & 24.4 & 89.4 \\
        100 & 0.463 & 49.5 & 46.7 & 13.4 & 17.9 & 94.1 & 22.6 & 24.1 & 87.5 \\
        200 & 0.459 & 47.4 & 45.3 & 12.2 & 16.8 & 96.5 & 22.1 & 23.2 & 85.6 \\
        500 & 0.458 &44.7 &43.6  & 8.2&13.2  &98.8 & 21.6 &17.4  &89.1  \\
        2000 & 0.428 &44.9 &43.5  &1.4 &9.5  &99.8 &11.3  &14.6  &90.2 \\
        5000 & 0.417 &44.7 &42.6  &0.8  &4.6 &99.7  &3.6  &10.3 &89.7\\
        \hline
    \end{tabular}
    \label{cluster-ablation-vgg}
\end{table*}

\begin{table*}
  \caption{Influence of the number of missing categories and noise rate on the MUSIC-solo, MUSIC-Synthetic and MUSIC-Duet dataset.}
  \label{loss-ablation-miss}
  \centering{
\begin{tabular}{c|cc|ccc|ccc}
\hline
Dataset          & \multicolumn{2}{c|}{MUSIC-solo} & \multicolumn{3}{c|}{MUSIC-Synthetic} & \multicolumn{3}{c}{MUSIC-Duet} \\ \hline
Number of missing categories & IoU@0.5          & AUC          & CIoU@0.3      & AUC       & NSA      & CIoU@0.3    & AUC     & NSA    \\ \hline
0                & 51.4             & 43.6         & 32.3          & 23.5      & 98.5     & 30.2        & 22.1    & 83.1   \\
1                & 49.1             & 40.4         & 21.8          & 19.8      & 92.2     & 12.4        & 13.5    & 98.3   \\
2                & 45.8             & 43.7         & 10.4          & 14.6      & 92.1     & 6.6         & 12.8    & 95.3   \\
3                & 46.7             & 42.5         & 8.4           & 8.5       & 96.1     & 3.7         & 9.1     & 98.1   \\ \hline \hline
Noise rate & IoU@0.5          & AUC          & CIoU@0.3      & AUC       & NSA      & CIoU@0.3    & AUC     & NSA    \\ \hline
0\%          & 51.4             & 43.6         & 32.3          & 23.5      & 98.5     & 30.2        & 22.1    & 83.1   \\
5\%           & 30.8             & 38.6         & 29.2          & 20.7      & 97.1     & 15.6        & 16.6    & 95.8   \\
10\%          & 29.7             & 37.8         & 23.3          & 20.2      & 91.6     & 11.4        & 14.9    & 96.9   \\
15\%          & 21.4             & 33.1         & 19.3          & 20.1      & 90.4     & 6.3         & 12.5    & 95.1   \\ \hline
\end{tabular} 
  }
\end{table*}

\textbf{Missing categories in the first stage.} To measure the robustness of the model, we conduct experiments in the presence of missing categories in the first stage, i.e., the single-source dataset does not include all the categories of objects, and the object dictionary is not complete. The top part of Table~\ref{loss-ablation-miss} shows the results on the MUSIC-solo, MUSIC-Synthetic, and MUSIC-Duet dataset. There are no missing categories during evaluation. The experimental results show that in the case of simple single sound scenarios, the missing categories do not have much effect, since no object dictionary is needed for single sounding object localization. However, in multiple source scenarios, an incomplete object dictionary affects the localization results a lot, especially in the case of a large number of missing categories. The effect of missing categories is particularly significant on the MUSIC-duet dataset. This is because the unbalanced instrument category distribution of duet videos brings great difficulty to training and bias to testing, e.g., more than 80\% duet videos contain the sound of the guitar. When the guitar is in the missing categories, it will have a huge impact on the results. 

\textbf{Noise videos in the first stage.} In practice, obtaining clean single source videos is difficult and it oftem contains noise of videos with multiple sound sources. Hence, we introduce multiple sound sources videos (taken from the MUSIC-duet dataset) and construct the dataset with different percentages of noise for the training of the first stage. The bottom part of Table~\ref{loss-ablation-miss} shows the localization results. Our model shows considerable robustness on the harder MUSIC-Synthetic dataset. However, the learned object dictionary has a huge impact on the results of multiple source localization on the MUSIC-duet dataset. This may be because the noise in the first stage is selected from the MUSIC-duet dataset, which mixes up the training of two stages, making it difficult to distinguish different objects on the MUSIC-duet dataset.

\textbf{Alternative optimization.} To validate the effects of alternative localization-classification optimization in the first stage, we compare the results with and without alternative optimization on VGGSound, Realistic DailyLife, MUSIC-Synthetic, and MUSIC-Duet dataset. As shown in Fig.~\ref{fig:alt-ablation-vgg}, the localization results with alternative learning are significantly better. This is because the categorization task provides considerable reference for generalized object localization~\cite{oquab2015object,zhou2016learning}, and cluster-classification training paradigm improves feature discrimination~\cite{caron2018deep,alwassel2020self}.

\textbf{Training settings for the second stage.} We also present some ablation studies on the second-stage training procedure and learning objectives. $\mathcal{L}_{loc}$ denotes localization loss in Eq.~\ref{eq:simple_local}, $\mathcal{L}_c$ the audiovisual consistency loss in Eq.~\ref{eq:stage2_kl} , and Prod denotes the silent area suppress operation in Eq.~\ref{eq:stage2_filter}. As shown in Table~\ref{2nd-ablation-vgg}, the Prod operation is crucial, especially for the synthetic scenes. This is because our manually synthesized audiovisual pair contains four different visual scenes in a single frame, where two objects are making sound with the other two silent. Without Prod operation, all visual objects would contribute high response in localization maps generated in Eq.~\ref{eq:stage2_loc}, which corrupts categorical audiovisual matching. Besides, the $L_c$ objective boosts synthetic and realistic data localization, proving that cross-modal modelling is facilitated by fine-grained consistency between two modalities.

\textbf{Loss function weight $\lambda$.} As shown in Fig.~\ref{fig:loss-ablation-vgg}, we find that our model is basically robust to the hyper-parameter $\lambda$ in the range of $[0.5, 1.0]$. However, when it is too small or too large, the performance drops. We empirically find that $\mathcal{L}_{loc}$ is easier to converge, and when $\lambda$ is too large, the model tends to overfit in terms of localization. When $\lambda$ is too small, it might fail to infer the correct sounding area to filter out background disturbance.

\section{Unsupervised object detection}
In this work, we utilize the correspondence between audio and vision to perform class-aware sounding objects localization. In the single sound source scenario, the learned localization heatmap and the cluster label have the potential to migrate to other vision tasks, since the localization heatmap can reflect the visual information of object and the cluster label offers category information. In this section, we aim to explore the application of our audiovisual network in object detection task.

\begin{table}
  \caption{\textbf{Unsupervised object results on the subset of ImageNet dataset.} The evaluation metric is $\text{AP}_{50}$.} 
  \label{table:detection}
  \centering{
\begin{tabular}{c|cccc|c}
\hline
Method    & Random & SS   & Proposal & Heatmap & Mix  \\ \hline
flute     & 4.5    & 0.0  & 2.1      & 9.1     & 10.3 \\
guitar    & 12.1   & 6.8  & 21.3     & 27.2    & 28.1 \\
accordion & 11.4   & 6.5  & 9.6      & 15.8    & 16.1 \\
xylophone & 5.1    & 6.6  & 10.2     & 18.9    & 18.7 \\
saxophone & 9.8    & 2.2  & 6.1      & 10.2    & 10.3 \\
cello     & 2.8    & 10.3 & 15.3     & 2.4     & 15.4 \\
violin    & 0.9    & 1.0  & 10.6     & 10.6    & 10.5 \\
trumpet   & 0.7    & 0.2  & 2.5      & 5.1     & 5.6  \\ \hline
$\text{mAP}_{50}$       & 5.9    & 4.2  & 9.7      & 12.4    & 14.4 \\ \hline
\end{tabular}
  }
\end{table}

\begin{figure}
    \centering
    \includegraphics[width=\linewidth]{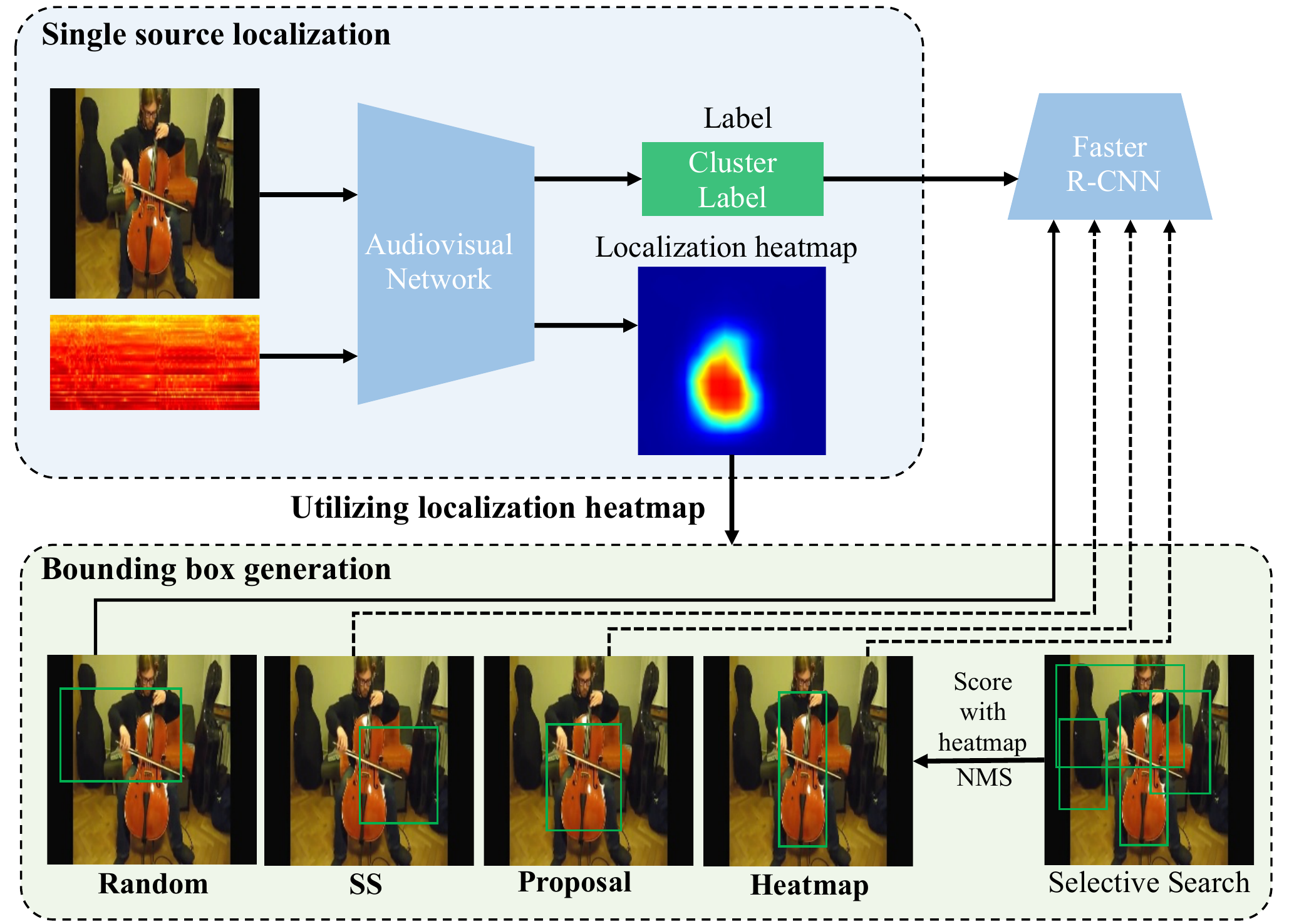}
    \caption{\textbf{Details of the object detection framework.} We conduct single source localization task on the MUSIC-solo dataset. The obtained localization heatmap is then used to generate bounding box. The generated bounding box and the cluster label are utilized to train a Faster-RCNN model. The solid line represents the selected bounding box construction method.}
    \label{fig:det_fram}
\end{figure}

\begin{figure}
\centering
    \subfigure[\textbf{Heatmap} bounding box]{
    \includegraphics[width=0.43\linewidth]{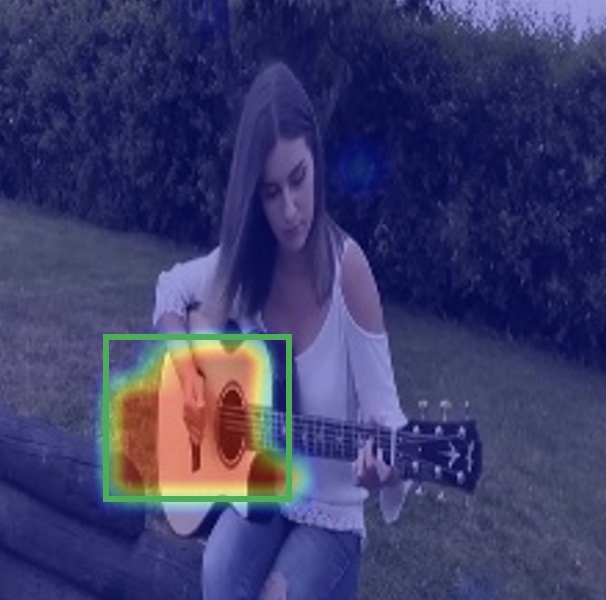}
    \includegraphics[width=0.43\linewidth]{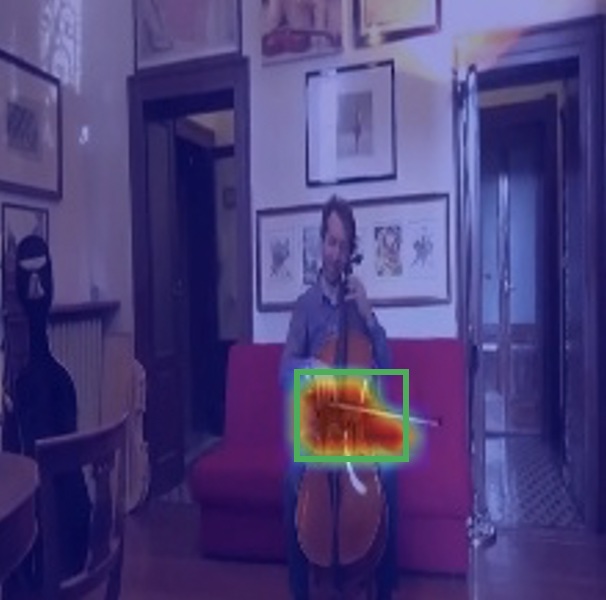}
    \label{fig:heatmap}
    }
    
    \includegraphics[width=0.43\linewidth]{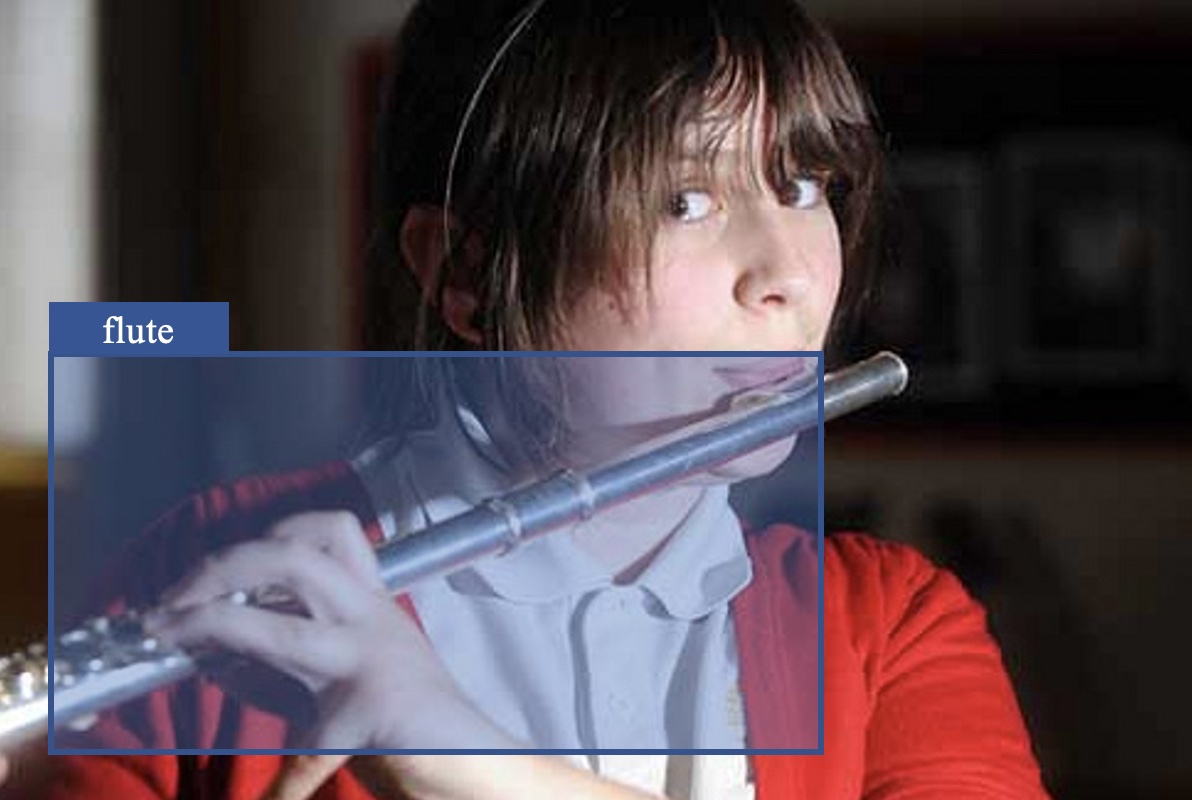}
    \includegraphics[width=0.43\linewidth]{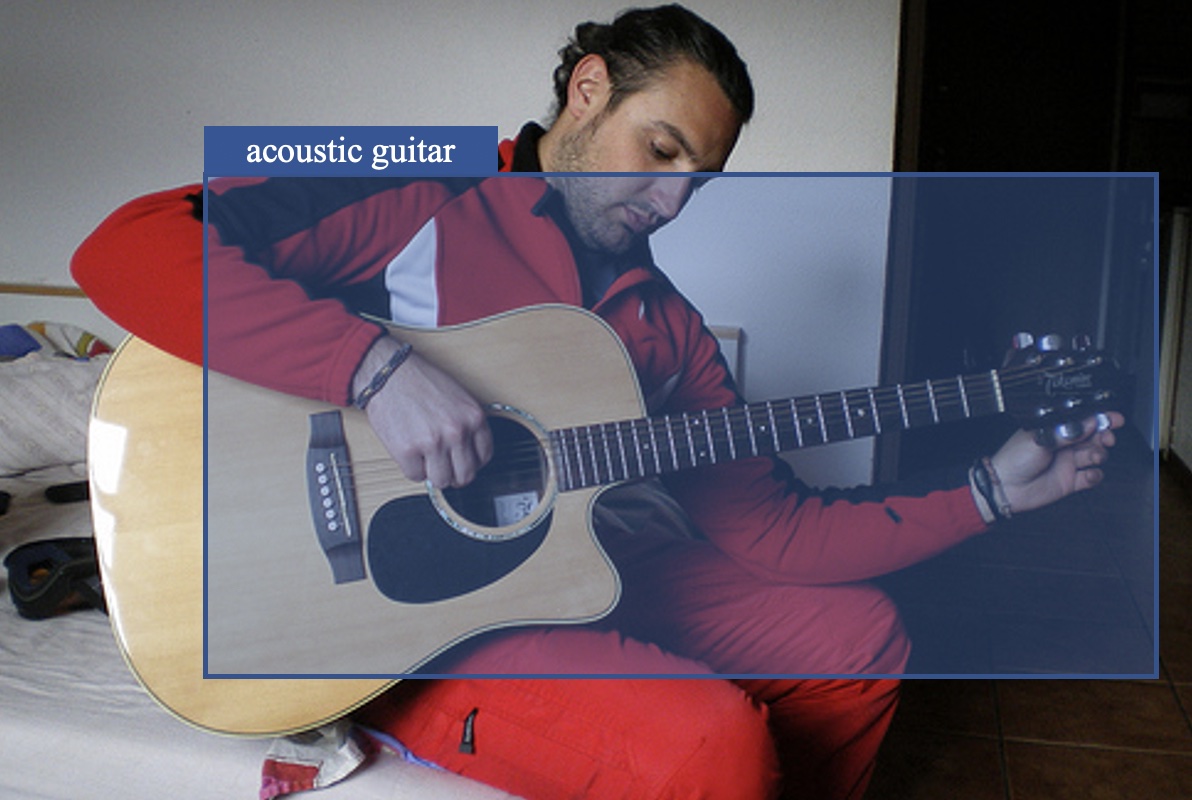}
    \subfigure[Object detection results]{
    \includegraphics[width=0.43\linewidth]{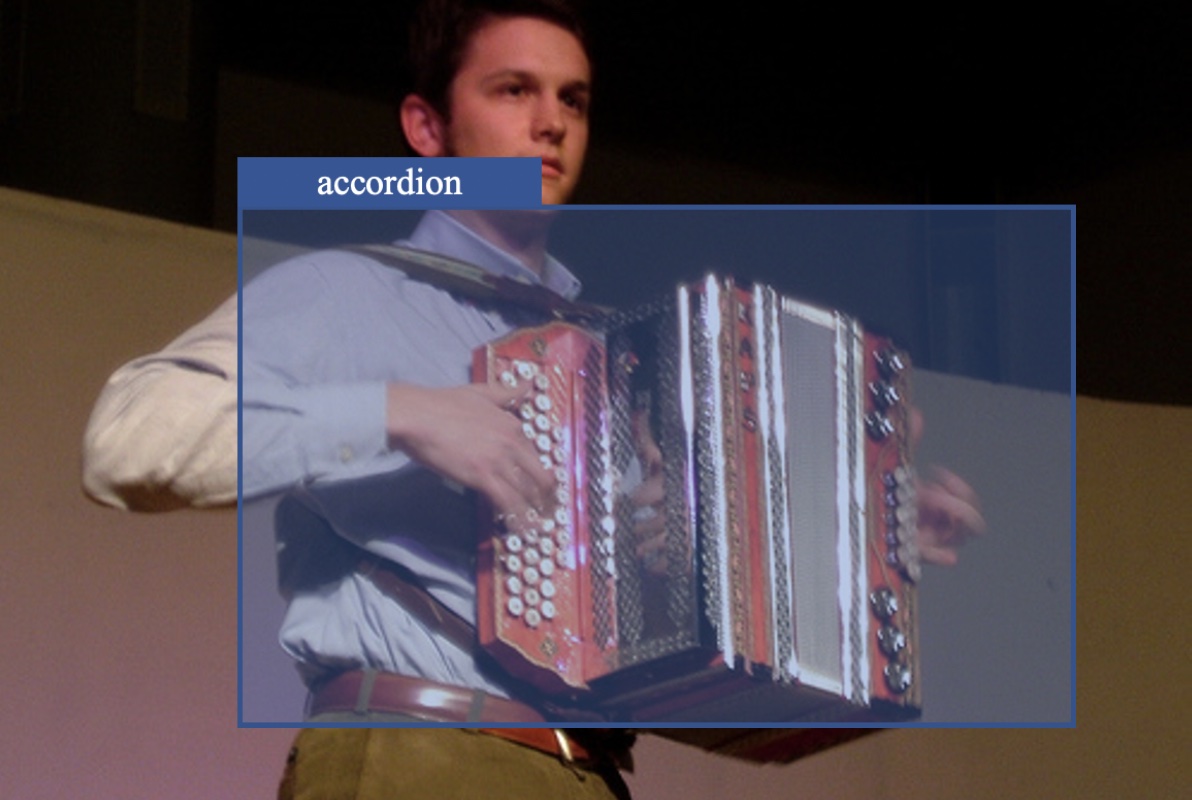}
    \includegraphics[width=0.43\linewidth]{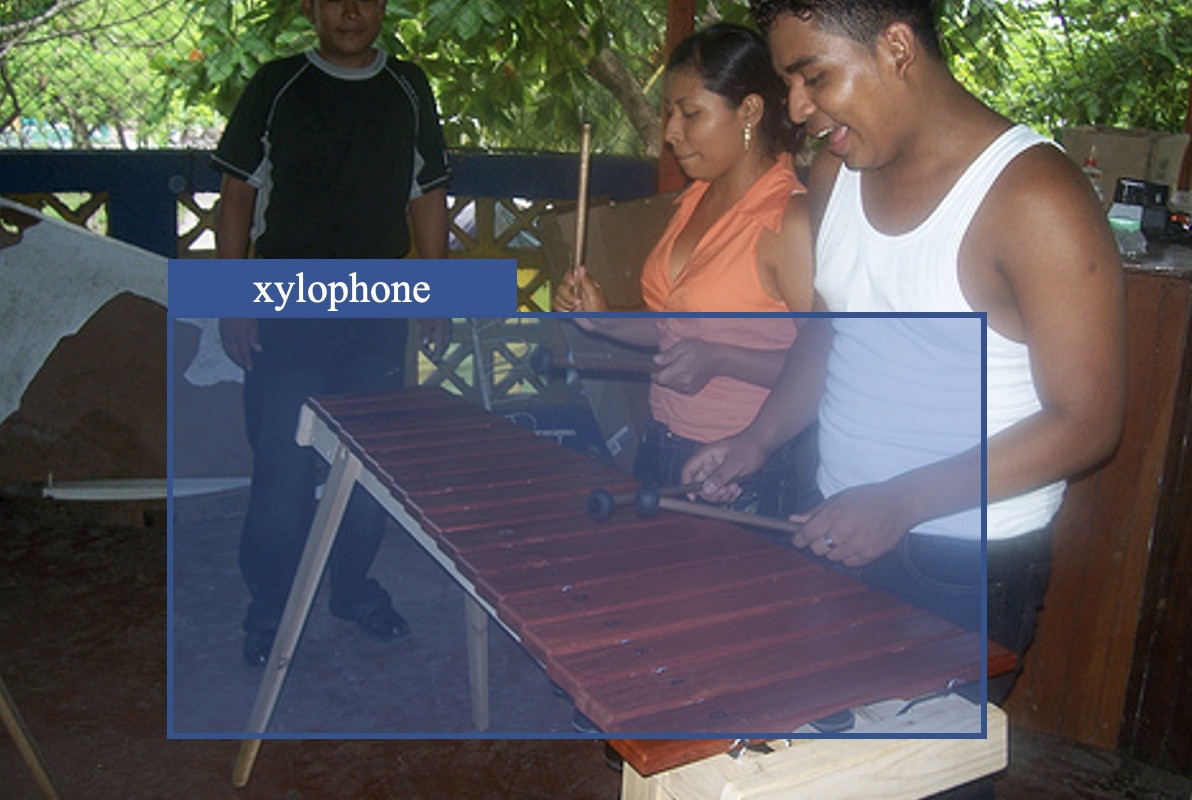}
    \label{fig:detection_re}
    }
    
    \subfigure[Failure cases]{
    \includegraphics[width=0.43\linewidth]{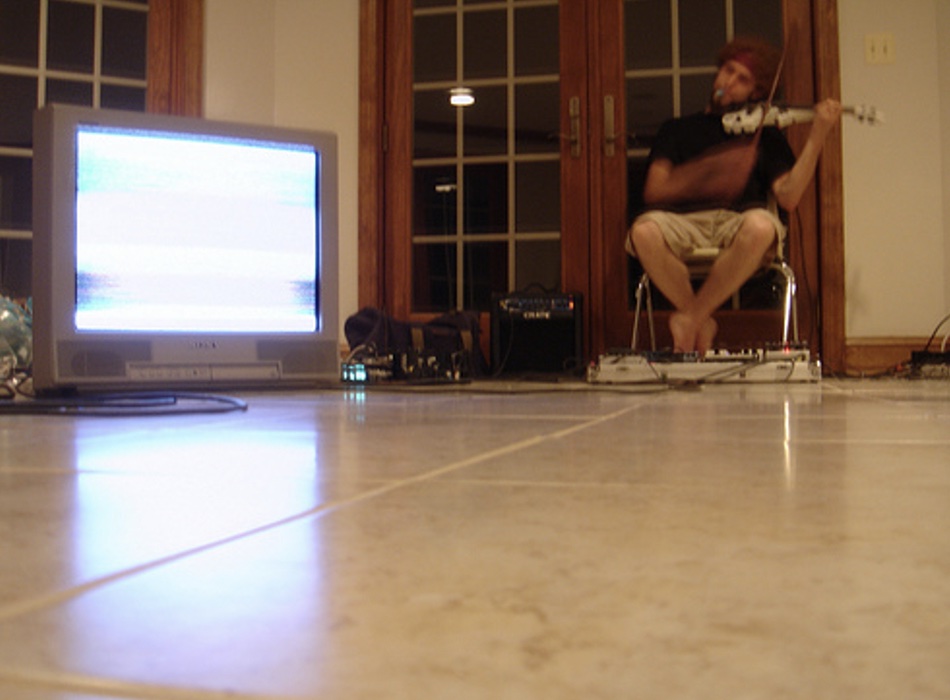}
    \includegraphics[width=0.43\linewidth]{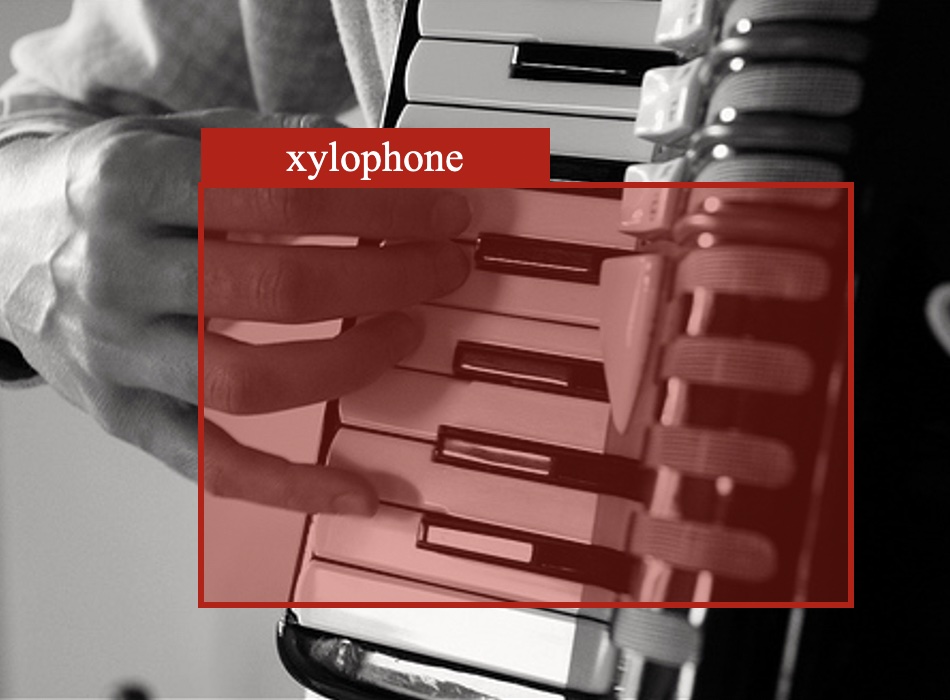}
    \label{fig:fail_cases}}
\caption{\textbf{Heatmap bounding box and object detection results on the subset of ImageNet dataset.} The bounding boxes generated by the \textbf{Heatmap} method are visualized. In addition, we also visualize our object detection results and some failure cases. Our method fails to detect objects that are not obvious in the figure. It is also difficult to recognize categories with similar visual information (both xylophone and saxophone keyboards have striped patterns).}
\label{fig:detection}
\end{figure}

Fig~\ref{fig:det_fram} shows the details of the object detection framework. Concretely, we first perform single source localization task on the MUSIC-solo dataset. The generated localization heatmap is then used to generate bounding box. The bounding box and the cluster label are utilized for the training of a Faster R-CNN~\cite{ren2015faster} model. Finally, the trained Faster R-CNN model is evaluated on the validation subset of the ImageNet dataset~\cite{deng2009imagenet}. To comparison, we conduct five methods to generate bounding box:
\begin{itemize}
  \item [1)]
  \textbf{Random} is simple baseline. We generate box with random size and location as the bounding box.
  \item [2)]
  \textbf{Selective Search (SS)}: This method follows the Selective Search method~\cite{uijlings2013selective}, and the proposal with the highest score is selected as the bounding box.
  \item [3)] 
  \textbf{Proposal} adopts the method that combines the localization heatmap and region proposal. First, we use Selective Search method to obtain object proposals. Then each proposal is scored based on the localization heatmap. After non-maximum suppression, the final proposal is used as the bounding box.
  \item [4)] 
  \textbf{Heatmap} adopts the bounding box constructed based on the localization heatmap. We normalize the heatmap and then select the region larger than the threshold (0.5 in our experiment). A box that can cover this region with the minimum size is extracted as the final bounding box.
  \item [5)]
  \textbf{Mix} method manually selects the better annotation results between \textbf{Heatmap} and \textbf{Proposal} as the final bounding box.
  
\end{itemize}

Based on the experiment results shown in Table~\ref{table:detection}, our method is able to achieve reasonable performance on the object detection task. \textbf{Proposal} combines the localization heatmap and is better than baselines, \textbf{Random} and \textbf{SS}. \textbf{Heatmap} is superior than \textbf{Proposal} because \textbf{Heatmap} fully focusing on sounding objects. \textbf{Proposal} failed on some categories, probably due to the size of the object being small and then the proposal tends to include the player as well (especially in the case of flute), leading to bounding boxes with low quality. But in the case of instruments with large sizes, such as the cello, the heatmap often can only capture parts of the instrument while the proposal can provide a more accurate bounding box. Therefore, after our manual selection, \textbf{Mix} achieves the best results. However, the overall AP for object detection is comparatively low for the reason that the localization heatmap sometimes cannot fully cover the object and the generated bounding boxes are not accurate, which is shown in Fig~\ref{fig:heatmap}. We also offer some object detection results as well as failure cases in Fig~\ref{fig:detection_re} and Fig~\ref{fig:fail_cases}. Although our method fails in some cases, such as when objects do not stand out in the figure or when objects of different categories have a similar appearance, these are common difficulties in the object detection task. Our method is able to distinguish different objects and generate bounding boxes effectively.
\section{Conclusion}
In this work, we introduce class-aware sounding objects localization without resorting to object category annotations and design an effective method with several dataset. We propose a step-by-step learning framework evolving from single source scenes to cocktail-party scenarios to learn robust visual object representation with category information through single object localization and then expand to multiple objects cases. We leverage category-level audiovisual consistency for fine-grained audio and sounding object distribution matching, which helps to refine learned audiovisual knowledge. Experiments show that our method can achieve decent performance in music scenes as well as harder general cases where the same object could produce different sounds. Multiple ablation experiments are conducted to validate the robustness of our framework. Besides, we conclude two additional aspects for further discussion.\\
\textbf{Generation of category-representation object dictionary.} In this work, we use simple enumeration to construct cluster-to-category correspondence and achieves considerable results when the number of clusters is not too large. But it would be a huge burden to generate the category-representation object dictionary with high quality under the situation that the number of clusters or categories is quite large. It is essential to explore a more effective dictionary generation method in future studies. \\
\textbf{Partition of single and multiple source videos.} Though our model does not resort to category labels to learn audiovisual knowledge, it still requires the rough partition of single and multiple source data. According to the results of the ablation study, the model performance is affected when multiple source noises are mixed in the first stage. In the future study, we will explore a better plan to learn category and visual representation information from a mixture of simple and complex audiovisual scenes.\\
\textbf{Sounding instance determination.} Our method achieves class-aware multiple sounding objects localization in different cases, but it cannot determine which specific instance is producing sound where there exist multiple instances of the same category in the scene. It might be promising to integrate motion and temporal information with multiple frames to tackle this problem.


%



\ifCLASSOPTIONcompsoc
  \section*{Acknowledgments}
\else
  \section*{Acknowledgment}
\fi

The work is supported by Intelligent Social Governance Platform, Major Innovation \& Planning Interdisciplinary Platform for the “Double-First Class” Initiative, Renmin University of China, Beijing Outstanding Young Scientist Program (NO. BJJWZYJH012019100020098), Public Computing Cloud, Renmin University of China, National Natural Science Foundation of China (NO. 62106272), 2021 Tencent AI Lab Rhino-Bird Focused Research Program  (No. R202141) and the cooperation project with Big Data Lab, Baidu, and in part by the Research Funds of Renmin University of China (NO. 21XNLG17) and National Natural Science Foundation of China under grant U21B2013.

\ifCLASSOPTIONcaptionsoff
  \newpage
\fi



\bibliographystyle{IEEEtran}
\bibliography{location.bib}
%

%
\begin{IEEEbiography}[{\includegraphics[width=1in,height=1.25in]{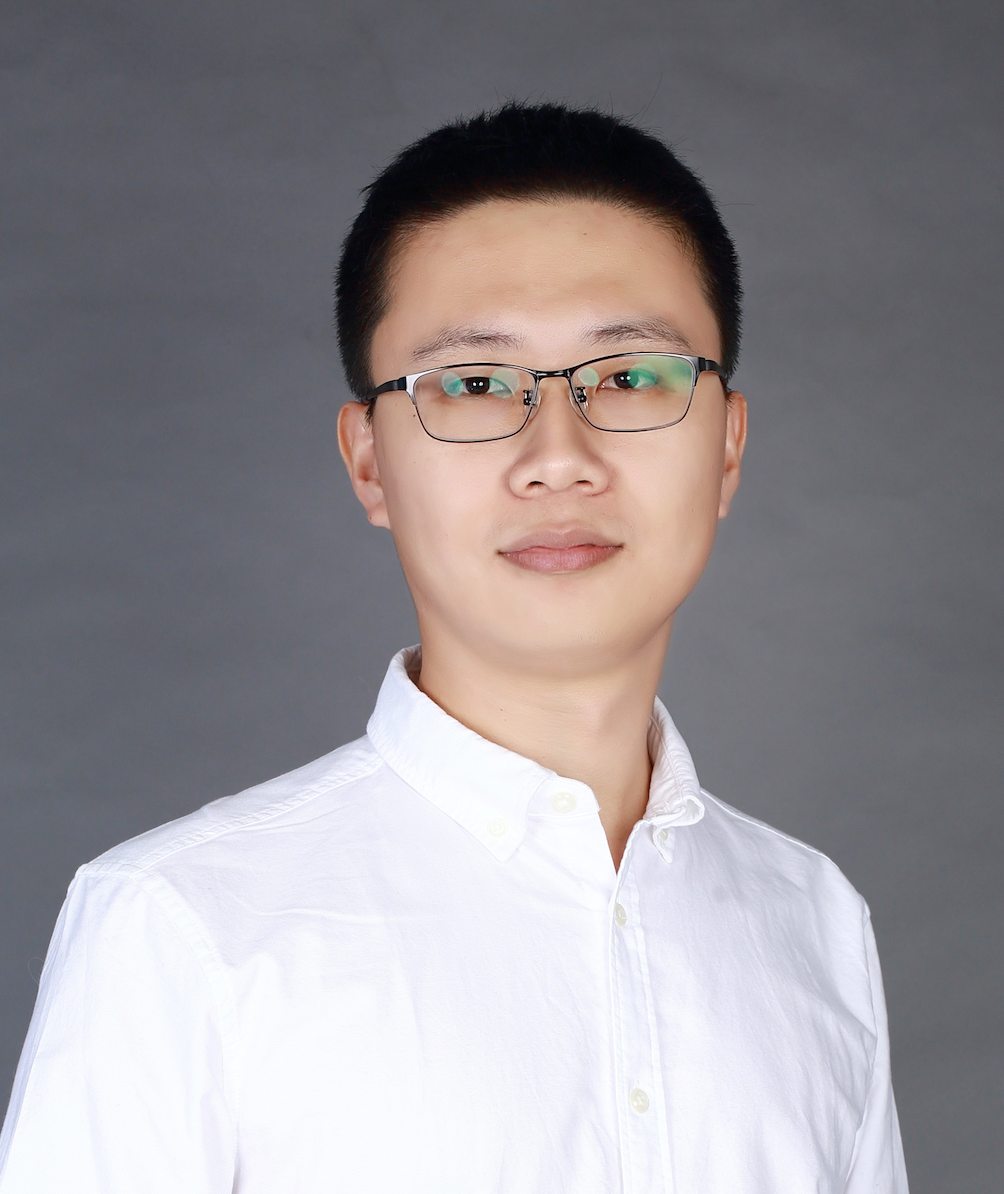}}]{Di Hu} is an assistant professor at Gaoling School of Artificial Intelligence, Renmin University of China. His research interests include multimodal perception and learning. He has published more than 20 peer-reviewed top conference and journal papers, including NeurIPS, CVPR, ICCV, ECCV etc. He served as a PC member of several top-tier conferences. Di is the recipient of the Outstanding Doctoral Dissertation Award by the Chinese Association for Artificial Intelligence, also the recipient of ACM XI’AN Doctoral Dissertation Award.
\end{IEEEbiography}

\begin{IEEEbiography}[{\includegraphics[width=1in,height=1.25in]{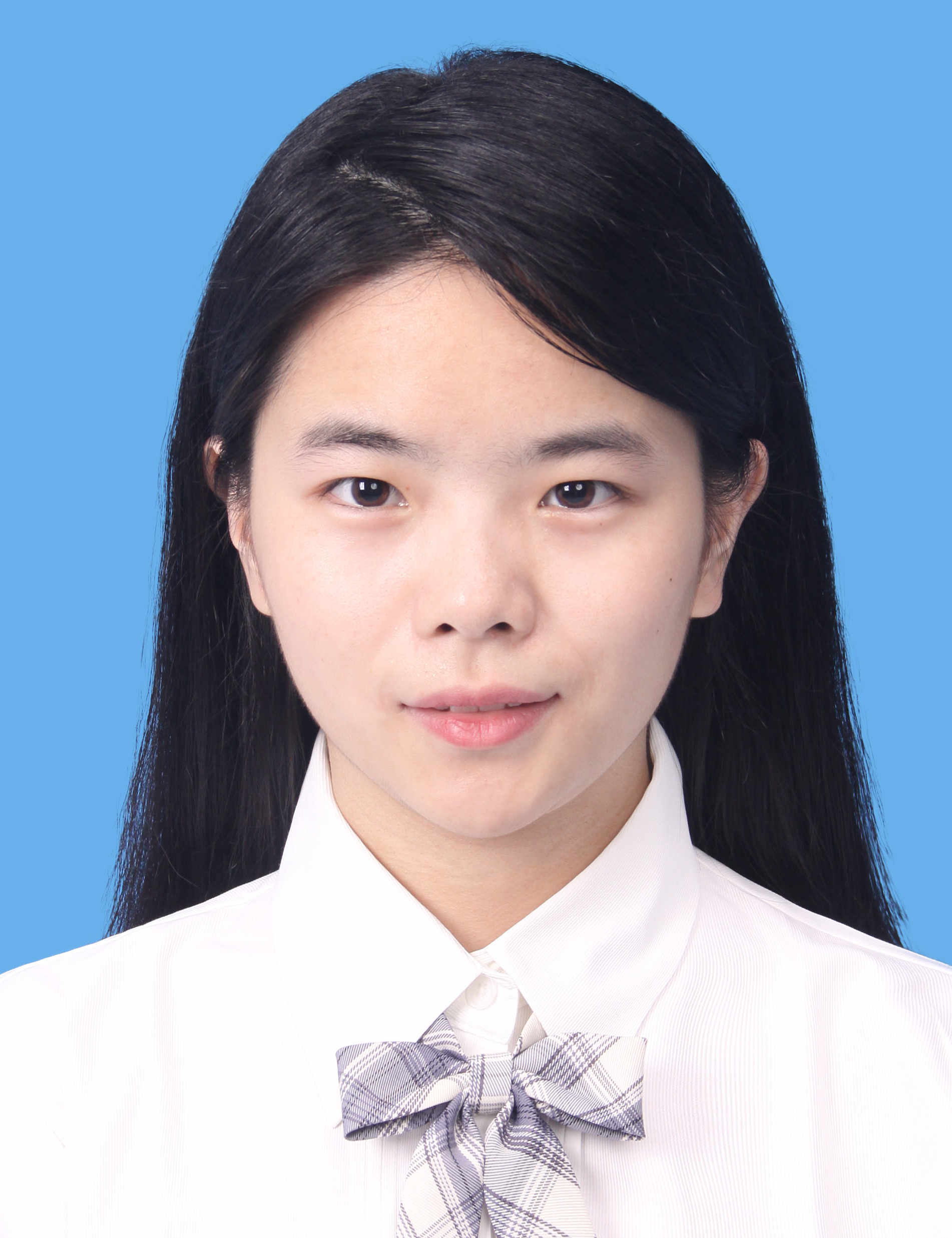}}]{Yake Wei}
received her B.E. degree of computer science from University of Electronic Science and Technology of China, China, in 2021. She is currently a Ph.D student in Gaoling School of Artificial Intelligence, Renmin University of China. Department. Her current research interests include multimodal perception and learning.
\end{IEEEbiography}

\begin{IEEEbiography}[{\includegraphics[width=1in,height=1.25in]{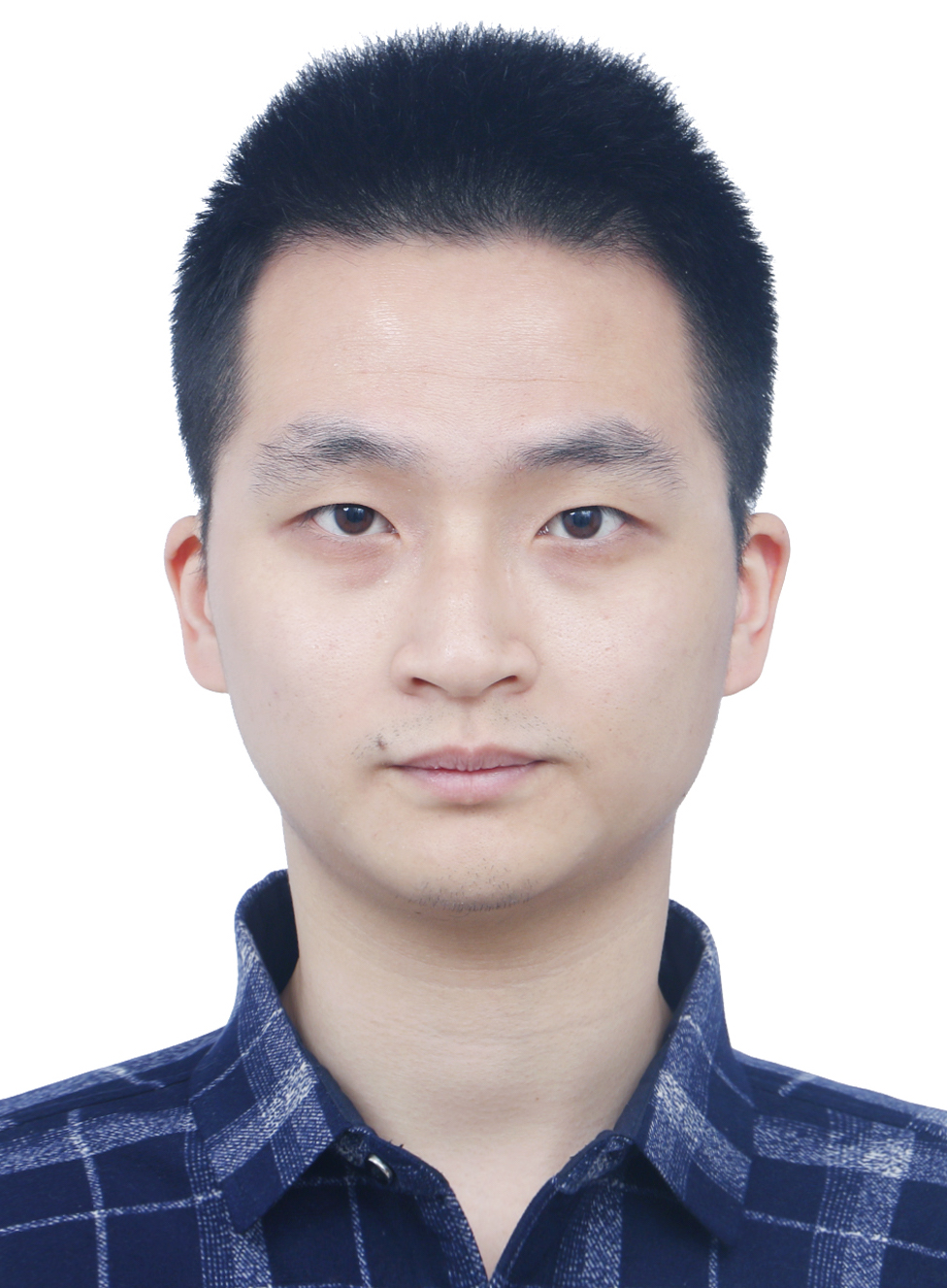}}]{Rui Qian}
received his B.E. degree of Information Engineering from Shanghai Jiao Tong University, China, in 2021. He is currently a Ph.D. student in Information Engineering in the Chinese University of Hong Kong. His current research interests include computer vision and machine learning.
\end{IEEEbiography}

\begin{IEEEbiography}[{\includegraphics[width=1in,height=1.25in]{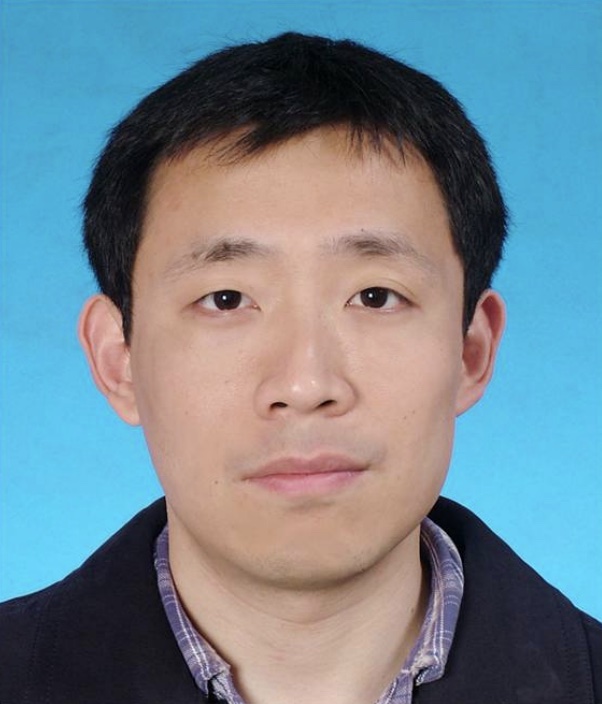}}]{Weiyao Lin}
received the B.E. and M.E. degrees from Shanghai Jiao Tong University, China, in 2003 and 2005, and the Ph.D degree from the University of Washington, Seattle, USA, in 2010. He is currently a professor at Department of Electronic Engineering, Shanghai Jiao Tong University, China. He has authored or coauthored 100+ technical papers on top journals/conferences including TPAMI, IJCV, TIP, CVPR, and ICCV. He is a senior member of IEEE and holds 20+ patents. His research interests include video/image analysis, computer vision, and video/image processing applications.
\end{IEEEbiography}

\begin{IEEEbiography}[{\includegraphics[width=1in,height=1.25in]{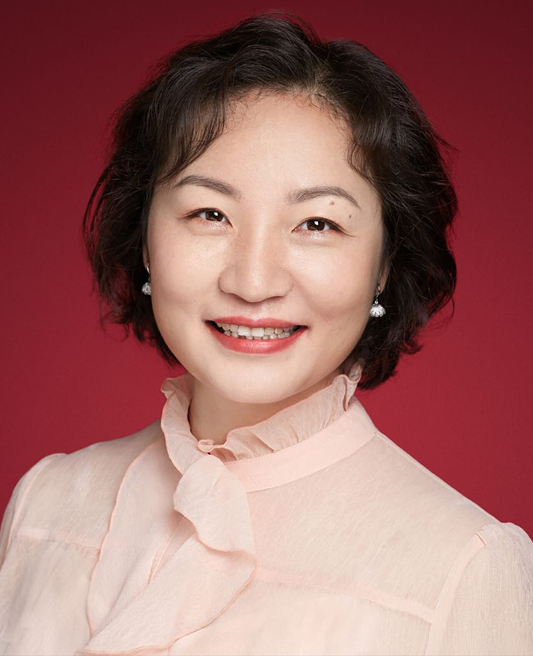}}]{Ruihua Song}
is now a tenured associate professor of Gaoling School of Artificial Intelligence, Renmin University of China. She received her B.E. and M.E. from Tsinghua University in 2000 and 2003. She received her PhD from Shanghai Jiao Tong University in 2010. Dr. Song published more than 80 research papers and hold more than 25 patents. Her recent research interests include AI based creation, multi-modal understanding of natural language, and multi-modal dialogue systems.
\end{IEEEbiography}

\begin{IEEEbiography}[{\includegraphics[width=1in,height=1.25in]{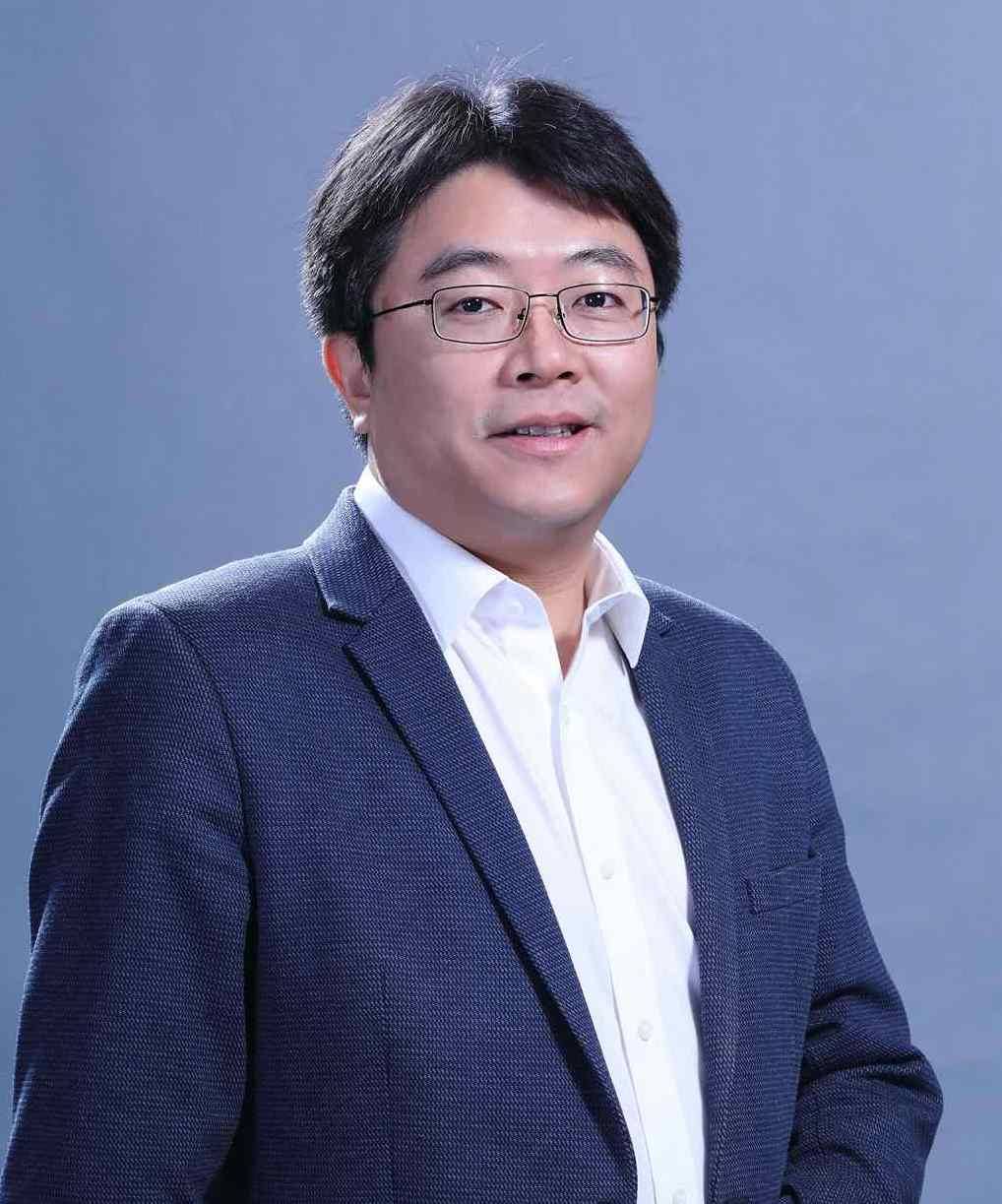}}]{Ji-Rong Wen}
is a full professor at Gaoling School of Artificial Intelligence, Renmin University of China. He worked at Microsoft Research Asia for fourteen years and many of his research results have been integrated into important Microsoft products (e.g. Bing). He serves as an associate editor of ACM Transactions on Information Systems (TOIS). He is a Program Chair of SIGIR 2020. His main research inter- ests include web data management, information retrieval, data mining and machine learning.
\end{IEEEbiography}







\end{document}